\documentclass[sn-mathphys,Numbered]{sn-jnl}

\usepackage[version=4]{mhchem}
\usepackage{graphicx}%
\usepackage{multirow}%
\usepackage{amsmath,amssymb,amsfonts}%
\usepackage{amsthm}%
\usepackage{mathrsfs}%
\usepackage[title]{appendix}%
\usepackage{xcolor}%
\usepackage{textcomp}%
\usepackage{float}
\usepackage{placeins}
\usepackage{booktabs}%
\usepackage{algorithm}%
\usepackage{algorithmicx}%
\usepackage{algpseudocode}%
\usepackage{subcaption}
\usepackage{tabularx}    
\usepackage{array}
\usepackage{upgreek}
\usepackage{float}
\usepackage{manyfoot}
\usepackage{listings}

\theoremstyle{plain}
\theoremstyle{plain}%

\theoremstyle{plain}%

\begin{document}

\title[Article Title]{Physics Encoded Blocks in Residual Neural Network Architectures for Digital Twin Models}


\author*[1]{\fnm{Muhammad Saad} \sur{Zia}}\email{msz6@leicester.ac.uk}
\author[1]{\fnm{Corentin} \sur{Houpert}}\email{ch586@leicester.ac.uk}
\author[1]{\fnm{Ashiq} \sur{Anjum}}\email{a.anjum@leicester.ac.uk}

\author[1]{\fnm{Lu} \sur{Liu}}\email{l.liu@leicester.ac.uk}

\author[2]{\fnm{Anthony} \sur{Conway}}\email{anthony.conway@bt.com}
\author[2]{\fnm{Anasol} \sur{Peña-Rios}}\email{anasol.penarios@bt.com}

\affil[1]{\orgdiv{School of Computing and Mathematical Sciences}, \orgname{University of Leicester}, \orgaddress{\city{Leicester}, \postcode{LE1 7RH},  \country{United Kingdom}}}

\affil[2]{\orgname{BT Research Labs}, \city{Ipswich}, \postcode{IP5 3RE}, \country{United Kingdom}}


\abstract{Physics Informed Machine Learning has emerged as a popular approach for modeling and simulation in digital twins, enabling the generation of accurate models of processes and behaviors in real-world systems. However, existing methods either rely on simple loss regularizations that offer limited physics integration or employ highly specialized architectures that are difficult to generalize across diverse physical systems. This paper presents a generic approach based on a novel physics-encoded residual neural network (PERNN) architecture that seamlessly combines data-driven and physics-based analytical models to overcome these limitations. Our method integrates differentiable physics blocks—implementing mathematical operators from physics-based models—with feed-forward learning blocks, while intermediate residual blocks ensure stable gradient flow during training. Consequently, the model naturally adheres to the underlying physical principles even when prior physics knowledge is incomplete, thereby improving generalizability with low data requirements and reduced model complexity. We investigate our approach in two application domains. The first is a steering model for autonomous vehicles in a simulation environment, and the second is a digital twin for climate modeling using an ordinary differential equation (ODE)-based model of Net Ecosystem Exchange (NEE) to enable gap-filling in flux tower data. In both cases, our method outperforms conventional neural network approaches as well as state-of-the-art Physics Informed Machine Learning methods.}

\maketitle

\section{Introduction}\label{sec:introduction}

\textit{Digital twins} are virtual counterparts of any real-world system or entity synchronized with the system or entity through real-time data. A central concept in digital twins is the modelling and simulation of processes and behaviour associated with the real-world entity. These models allow users to investigate behaviour in the alternative \textit{`what-if'} scenarios and can be divided into two fundamental categories: \textit{physics-based models} and \textit{data-driven models}.  

While physics-based models offer better accuracy and reliability of predictions, they are often computationally expensive and inflexible to be used in near-real-time simulations - especially in the context of Digital Twins \cite{Willcox2021, Wang2022}. On the other hand, data-driven models provide flexible and adaptive solutions, especially with a limited understanding of underlying physics. However, they popularly suffer in terms of reliability and interpretation. \cite{Rasheed2020}. The benefits of combining physics with data-driven models to target these limitations are found in several studies, especially in the context of developing simulation models in situations with limited availability of prior physics knowledge and real-world data \cite{Rasheed2020, Thelen2022}. 

\begin{figure}[htbp]
    \centering
    \includegraphics[width=1\linewidth]{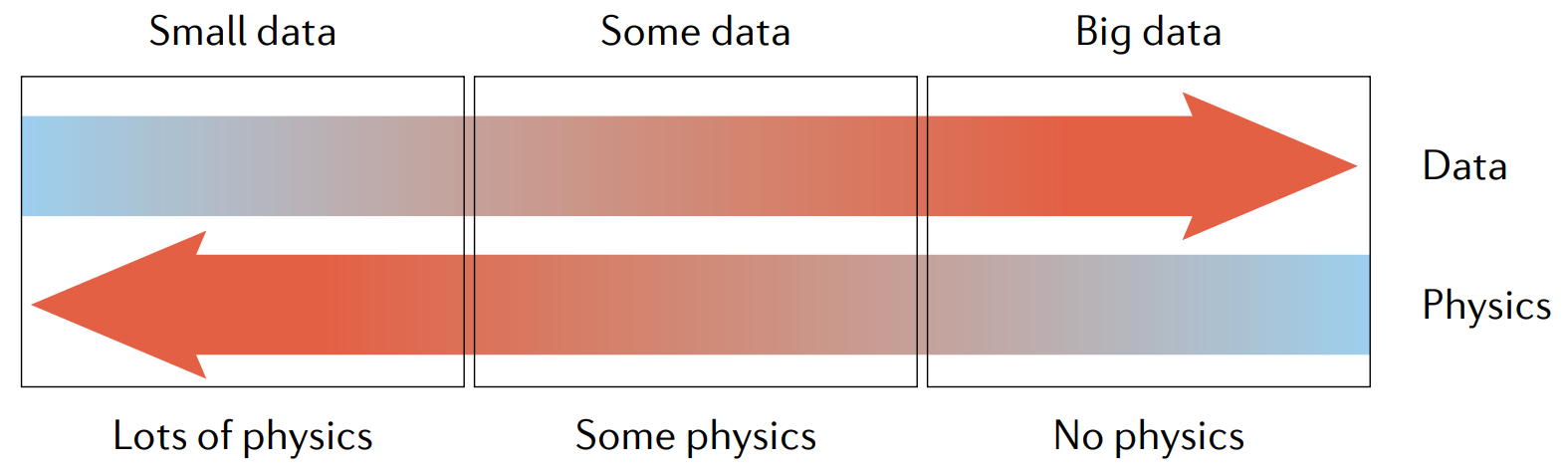}
    \caption{Three possible categories of physical problems and associated available data and knowledge of physics (Figure from \cite{Lu2021}).}
    \label{fig:physics_v_data}
\end{figure}

Modelling of physical systems can be categorized into three regimes based on the availability trade-off between data and known physics, as illustrated in Figure \ref{fig:physics_v_data}. As data increases, the balance shifts from leveraging prior physics knowledge to data-driven extraction of physical relationships. In the ubiquitous middle ground, some physics and data are known, with potentially missing parameters and initial/boundary conditions, and this is the most common scenario in models for digital twins in general.  \cite{Lu2021}. 

\begin{figure}[htbp]
    \centering
    \includegraphics[width=0.75\linewidth]{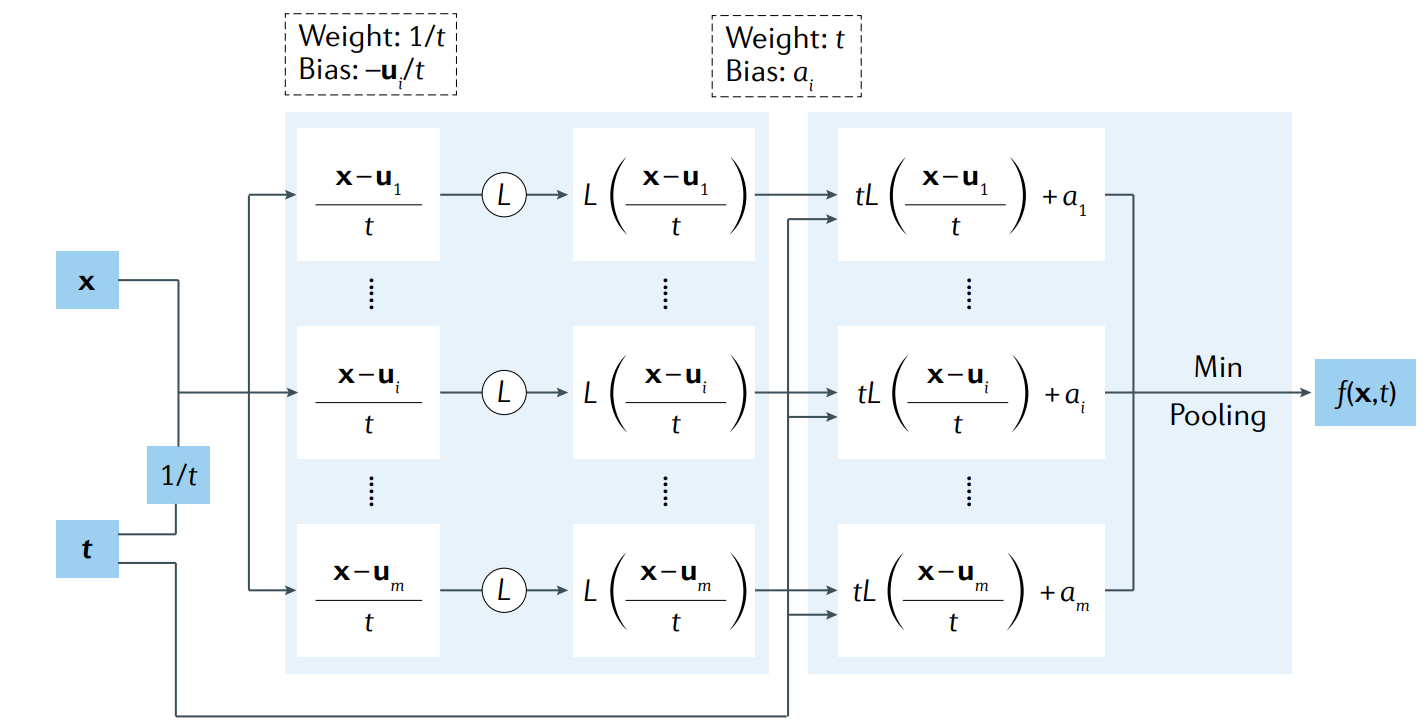}
    \caption{A neural network with the Lax–Oleinik formula baked into the architecture. Here, $f$ represents the solution of the Hamilton–Jacobi partial differential equations, while $x$ and $t$ correspond to the spatial and temporal variables, respectively. The activation function $L$ is convex and Lipschitz. The parameters $a_i \in \mathbb{R}$ and $u_i \in \mathbb{R^n}$ are specific to the neural network. Image courtesy of J. Darbon and T. Meng, Brown University and cited here from \cite{Lu2021}.}
    \label{fig:hj_pde_pinn}
\end{figure}

Physics-informed machine learning is a recent domain that offers a streamlined approach to incorporate both data and the fundamental laws of physics as prior knowledge, even when dealing with models with incomplete physical information. This integration, or \textit{`bias'}, is achieved by regularizing the loss function of the learning algorithms to adhere to the constraints defined by the respective physics models and equations. More specialized approaches involve specifically designed neural networks, which ensure that the predictions generated adhere to the governing physical principles. One of these approaches recently gaining traction in the research community is known as `differentiable physics'. This approach allows differentiable physics models to be directly baked into neural network architectures. An example is the recent work on new connections between neural network architectures and viscosity solutions to Hamilton–Jacobi PDEs \cite{Darbon2021} as shown in Figure \ref{fig:hj_pde_pinn}. 

While existing work mainly focuses on isolated physical phenomena \cite{Li2019, Toussaint2018, SanchezGonzalez2018}, and lacks the generalization of the approaches to wider problems, we propose Physics Encoded Residual Neural Network (PERNN), a general framework to capture complex behaviour of real-world systems. Our approach aims to target scenarios where data availability from the target systems is limited to specific operating scenarios and known physics only captures specific regions of the behaviour. Our approach defines a hybrid neural network which comprises feature extraction blocks to compute unknown variables (learning blocks) and combines them with differentiable physics operator layers (physics blocks) to model the complete behaviour of the target system. To facilitate the convergence of the model, we introduce intermediate `residual' layers between the physics and the learning blocks. We also make our code available publicly on GitHub.\footnote{\url{https://github.com/saadzia10/Physics-Encoded-ResNet.git}}
 
In this way, domain knowledge can inform the underlying neural network architecture while the gradient descent optimization using data can fill in gaps \cite{Seo2022, Stewart2019}. The goal is to leverage interpretable mechanistic modelling to improve generalization and avoid relying solely on empirical correlations \cite{Greydanus2019, Raissi2019}. This approach could benefit several complex systems, including robotic platforms \cite{Cranmer2020, Mrowca2018}, autonomous vehicles \cite{Ajay2018}, and aircraft swarms \cite{Mohan2021}. 

In this paper, we present two concrete applications to evaluate our method. The first application focuses on modeling dynamic steering for autonomous grounded vehicles. In this case, we develop a hybrid neural network architecture that integrates differentiable geometrical and kinematic operators from the well-established pure pursuit algorithm with learning blocks that discover intermediate unknown variables in the driving environment. The model is trained on observation data from an expert driving system in a simulation environment to learn the expert's behavior or \textit{policy} using supervised learning. This forms a Behaviour Cloning problem, which involves transforming expert demonstrations into i.i.d. state-action pairs and learning to \textit{imitate} the expert actions given each recorded state. In the context of digital twins, this approach is particularly useful because it models the exact behavior of a real-world system using observational data.

The second application demonstrates our approach in a real-world digital twin setting for climate modeling. Net Ecosystem Exchange (NEE), a key metric representing the net CO$_2$ exchange between ecosystems and the atmosphere, is typically measured by flux towers that often suffer from data gaps due to instrumentation failures and other issues. To address this challenge, we model NEE using an ordinary differential equation (ODE) derived from an Arrhenius-type formulation of ecosystem respiration. Our Physics Encoded Residual Neural Network (PERNN) for NEE explicitly incorporates the NEE ODE into the network via a dedicated physics block, while learning and residual blocks predict critical intermediate variables such as temperature sensitivity (\(E_0\)), base respiration (\(r_{b\_night}\)), and the rate of change of ambient air temperature (\(\frac{dT}{dt}\)). This integration enables accurate gap-filling and forecasting of NEE, leading to improved digital twin fidelity for climate monitoring and forecasting.

This paper presents the following key contributions:
\begin{itemize}
    \item It is the first (to our understanding) generic framework for embedding physics models into neural networks for end-to-end training on observation data.
    \item It is a significantly more interpretable hybrid neural network comprising learning blocks that predict human-understandable variables and physics blocks that predict the actual target variables.
    \item It allows the implicit discovery of unknown environmental variables that are not measured or observed.
    \item It outperforms conventional neural networks regarding key driving metrics on unseen road scenarios in simulation environments.
    \item It uses significantly fewer observation data, with fewer parameters, than conventional neural networks while outperforming key driving metrics.
    \item It achieves superior gap-filling performance in digital twin applications for climate modeling by accurately forecasting Net Ecosystem Exchange (NEE) from flux tower data, outperforming state-of-the-art methods such as Random Forest and XGBoost in key error metrics.
    \item It provides a robust digital twin framework for environmental monitoring by integrating an ODE-based physics model of ecosystem respiration directly into the neural network architecture.
\end{itemize}

\section{Related Work}\label{sec:related_work}

Physics Informed Machine Learning (PIML) is a recent approach that combines prior physics knowledge with conventional machine learning approaches to form hybrid learning algorithms of processes. Approaches like Physics-Informed Neural Networks (PINNs) encode analytical equations as loss terms to constrain deep learning models. Making a learning algorithm physics-informed refers to applying appropriate observational, inductive, or learning biases that direct the learning process towards identifying physically consistent solutions \cite{Lu2021}.  

Several methods have been proposed to incorporate observational bias in the form of input data generated from variable fidelity physical systems as a weak form of prior knowledge \cite{Lu2021,Kashefi2021, Li2021}. However, abundant data must be available, especially for large deep-learning models, making scaling these approaches to other areas difficult. Another approach to PIML involves incorporating a learning bias into the learning model by penalizing the cost function of deep learning algorithms regarding adherence to a physical constraint, such as satisfying a partial differential equation (PDE) associated with the modelled process. In this way, the learning model attempts to fit both the observational data and the constraining equation or rule, for example, conservation of momentum and mass \cite{Raissi2019, Lagaris1998, Zhu2019, Geneva2020, Wu2020}. This, however, only adds a soft constraint to the learning model as the physics no longer plays a direct, explicit role after the training of the model is complete. A more direct approach to PIML is incorporating an inductive bias into the learning model through crafted architectures of neural networks (NNs) to embed prior knowledge and constraints related to the given task, such as symmetry and translation invariance. For differential equations in particular, several approaches have been investigated to modify NN architectures to satisfy boundary conditions or encode a priori parts of PDE solutions \cite{Lagaris1998,Beidokhti2009,Wang2020,Cai2020,Darbon2021}. However, these models can struggle with extremely complex emergent behaviours arising from unknown dynamics outside the scope of hand-coded physics \cite{Miles2020}.

Differentiable physics-based PIML methods are based on incorporating inductive biases into the neural network architectures. This has allowed the integration of physical simulators directly into deep learning architectures through automatic differentiation \cite{Degrave2019, Holl2020}. But most of these applications have focused on modelling isolated phenomena rather than full-system behaviours and provide particular architectures suited to only the respective problem spaces \cite{Li2019, Toussaint2018, SanchezGonzalez2018}. A generic framework for integrating physics models into learning-based data-driven models remains an open challenge. Furthermore, applying these methods in digital twin scenarios where the observational data and prior knowledge are limited and only partially known is restrictive and requires thorough exploration. The method defined in this paper herewith effectively addresses these issues.

\section{Knowledge Blocks with Residual Learning}\label{sec:approach}

To address the problem of learning the behaviour of a target system with limited observation data and partially known physics, we propose a novel approach to integrate physics models into traditional neural networks to form modular architectures with `knowledge blocks'. These blocks are of two categories:
\begin{enumerate}
    \item Physics blocks: representations of known physics equations as computational graphs comprising non-trainable parameters and static operators to reflect the mathematical operations of the equations.
    \item Learning blocks: conventional neural network layers comprising trainable parameters (weights).
\end{enumerate}

The physics blocks provide an a priori understanding of the system's fundamental behaviour. The learning blocks can learn unknown, intermediate variables in the input space required by the physics blocks to predict the target variable(s). Collectively, the connected blocks form a neural network that can be trained end to end on observation data generated by the target system. For the scope of this work, we assume that the physics block encoding the known physics equation(s) of the target system can be represented as a differentiable physics model. This is necessary to define the physics equation(s) as a computational graph that can be integrated with the conventional learnable layers of a neural network.

\begin{figure}[htbp]
    \centering
    \includegraphics[width=1\linewidth]{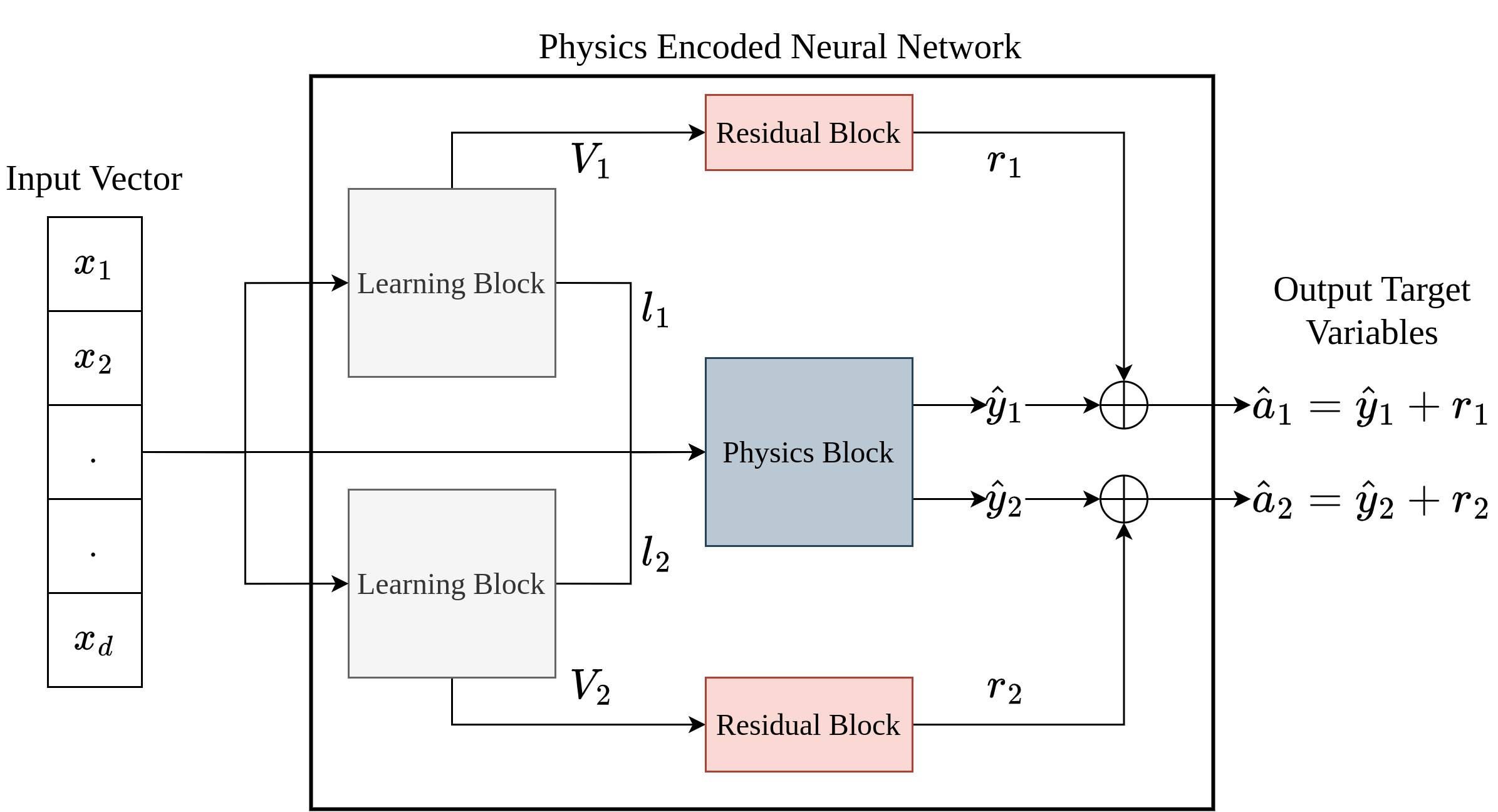}
    \caption{Proposed Physics-Encoded Neural Network Architecture with Residual Blocks. The example shows a problem case with two target intermediate variables and two target values to be predicted. Intermediate, unknown variables $l_1$ and $l_2$ are predicted (discovered) by the learning blocks and passed to the physics block as inputs alongside the original input vector $X$. Intermediate feature vectors $V_1$ and $V_2$ are passed from the learning blocks to the residual blocks to predict residual target values $r_1$ and $r_2$. The residuals eventually add up with the physics-predicted target values $y_1$ and $y_2$ to output the final target values $a_1$ and $a_2$.}
    \label{fig:pinn_general}
\end{figure}

To explain the proposed architecture, we define a Behaviour Cloning problem where expert demonstrations from the target system are used to learn a policy to model observed states (inputs) to actions (target values) taken. For a set of expert demonstrations $T$ from the target system, we assume i.i.d state-action pairs as:
\[T = \{(\textbf{x}_1, a_1), (\textbf{x}_2, a_2) ... (\textbf{x}_n, a_n) \}\]
We define a learning block as a set of fully connected layers $\mathcal{L}(\textbf{x};\theta)$ that takes observed state vector $\textbf{x}$ and outputs an intermediate state vector $\textbf{l}$. The learning block has a set of trainable weights $\phi$.
\[\textbf{l} = \mathcal{L}(\textbf{x};\phi)\]
This intermediate state could include unknown variables from the environment or parts of the behaviour that are not directly defined by prior physics and thereby require empirical modelling. We also define our physics block as a computational graph $\mathcal{P}$ that predicts an action $a \in \mathcal{A}$, from the available set of actions $\mathcal{A}$ usually taken by the target system, using the observed state $x$ and the intermediate state $\textbf{l}$:
\[\hat{a} = \mathcal{P}(\textbf{x}, \textbf{l})\]
We can then define our policy function $\pi$ to predict optimal action given the current state:
\[\hat{a} = \pi(\textbf{x}, \mathcal{L}, \mathcal{P}) \text{  where } \hat{a} \in \mathcal{A}\]

\paragraph{Residual Learning in Knowledge Blocks}

While the physics blocks provide key a priori knowledge to the model, they pose a significant problem in converging the loss function in the gradient descent algorithm. If the fixed relationships defined by these layers conflict with the learning blocks' modelled patterns,  it can result in conflicting gradients. This can make the training process highly unstable, thereby affecting convergence. Furthermore, static layers, such as those representing fixed physics equations, do not have learnable parameters or tweakable weights. When these static layers are integrated into a neural network, they can disrupt the flow of gradients to the learnable layers, often exacerbating the known problems of vanishing and exploding gradients. 

To overcome this problem, we propose to add `residual blocks' (similar to the concept introduced in the original ResNet paper \cite{He2016}) that branch out from the learning blocks to predict residual target value to be added to the physics block output. This introduces alternative pathways for gradients to flow to the learning blocks, mitigating the effect of the static layers within the physics blocks. Figure \ref{fig:pinn_general} shows an example case with two target intermediate variables and two target values to be predicted by the model. Intermediate, unknown variables $l_1$ and $l_2$ are predicted (discovered) by the learning blocks and passed to the physics block as inputs alongside the original input vector $X$. Intermediate feature vectors $V_1$ and $V_2$ are passed from the learning blocks to the residual blocks to predict residual target values $r_1$ and $r_2$. The residuals eventually add up with the physics-predicted target values $y_1$ and $y_2$ to output the final target values $a_1$ and $a_2$.

Consequently, we can update the following in our earlier example of a Behavioral Cloning problem. An intermediate layer output vector $V$ from the learning block $\mathcal{L}$ will be passed to a separate residual block $\mathcal{R}$, which also comprises trainable parameters $\phi_r$. 
\[r = \mathcal{R}(V;\phi_r)\]
The residual target value $r$ will then be added to the output from the physics blocks to predict the action $\hat{a}$.
\[\hat{a} = \mathcal{P}(\textbf{x}, \textbf{l}) + \textbf{r}\]
Thereby, the final policy function to the Behaviour Cloning problem will become:
\[\hat{a} = \pi(\textbf{x}, \mathcal{L}, \mathcal{P}, \mathcal{R}) \text{  where } \hat{a} \in \mathcal{A}\]

The gradient descent algorithm optimises any suitable loss function $L(a, \hat{a})$ that defines the difference between $a$ and $\hat{a}$. Since the policy function is differentiable and comprises the affine connection of $\mathcal{L}$ and $\mathcal{P}$, the gradients from $\nabla{L}$ will flow through the chain of static operators from the physics block $\mathcal{P}$ to optimize the weights in the learning block $\mathcal{L}$, thereby modulating the learning of intermediate unknown variables using known physical constraints of the system. This essentially gives the final model the capacity to learn unknown, empirical relationships in data while maintaining the fundamental structure of the physics of the behaviour being modelled.

\section{Experimental Study}

We validate the effectiveness of our approach in two distinct experimental setups. The first study involves modeling steering dynamics for autonomous vehicle simulation. In this scenario, we investigate our method within a simulated digital twin problem where the available prior physics is relatively simple and only partially defines the system's dynamics. The second study addresses a more complex, real-world digital twin problem for modeling CO$_2$ emissions and gap-filling of Net Ecosystem Exchange (NEE) in flux data. Here, we examine the ability of our approach to integrate an ordinary differential equation that partially characterizes the behavior of emissions with data-driven learning of the remaining dynamics from flux measurements, thereby enabling a forecasting model for gap-filling in emissions data—an important application in the climate digital twin domain

\subsection{Case Study 1: Steering Dynamics for Autonomous Vehicle Simulation }

We now apply this framework to a more complex digital twin problem where the available data and prior physics may not be substantially available to completely model the system dynamics independently. We investigate our approach in an autonomous driving task of steering a vehicle in a simulation as it manoeuvres around a race track. Using data generated from an existing driving agent within the simulation (treated as a reference agent), we train a feed-forward neural network architecture that combines fully connected layers that capture perceptual understanding with relevant mathematical operators from the physics-based steering model known as Pure Pursuit. Figure \ref{fig:pinn} illustrates the proposed architecture and is explained in detail in section \ref{sec:pinn}.

We use the TORCS open-source driving simulator as our experimental environment due to the tracks' diversity and rich non-visual sensory and perceptual information \cite{Espie2005}. The input parameter space offered to the model consists of a vehicle's state variables and perception information from the surrounding environment. Table 1 summarizes the subset of the available variables from the simulation environment used as input parameters and actuators/actions for our model.

\begin{figure}[htbp]
    \centering
    \includegraphics[width=1\linewidth]{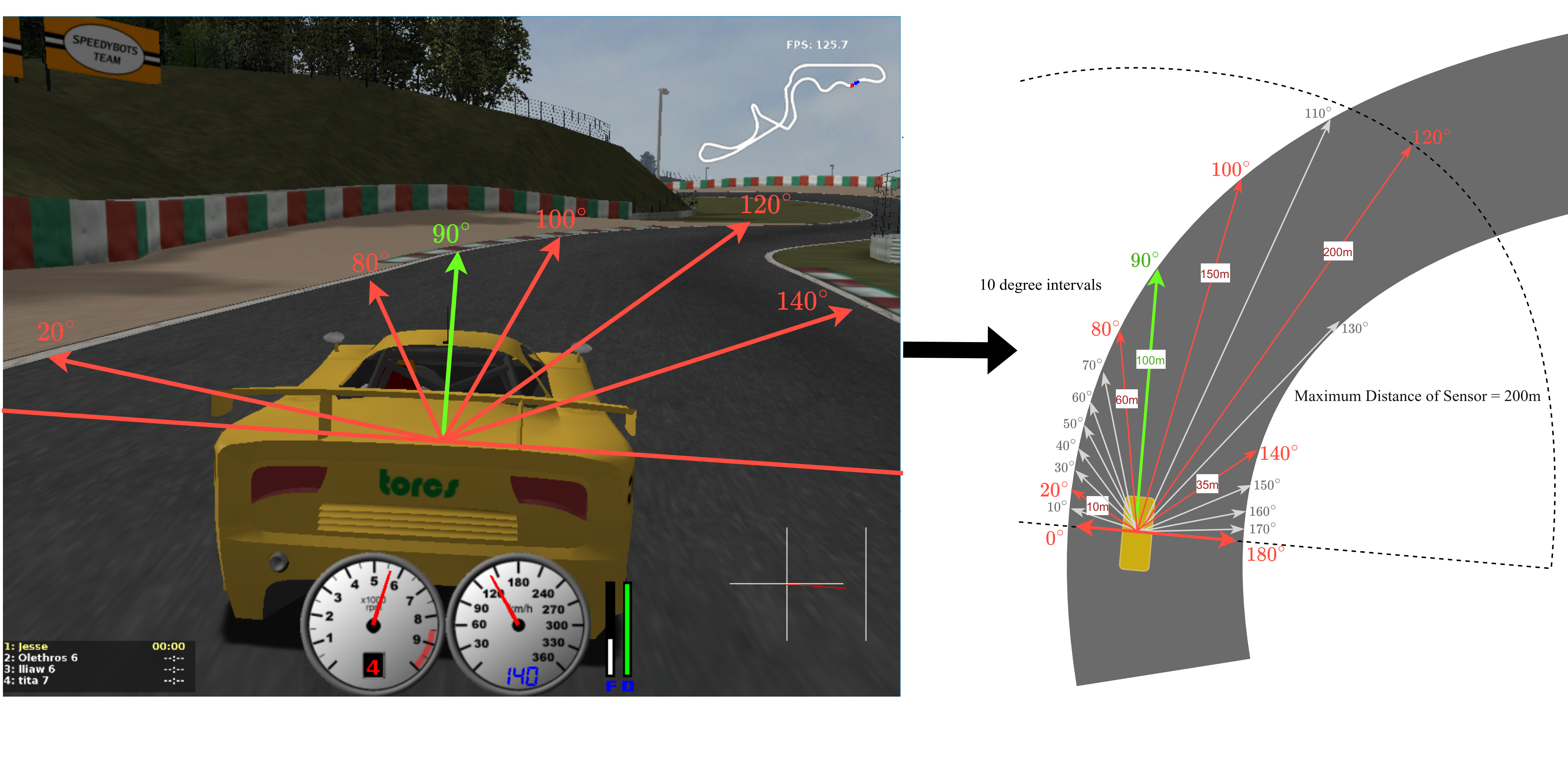}
    \caption{Illustration of the 19 range finder sensors ($R$) comprising distances to track edges, spanning clockwise from 0 to 180 degrees to the car axis at 10-degree intervals. The image from the TORCS simulation on the left shows a sample of sensors at different angles depicted with red arrows, which is further elaborated on in the figure on the right side. The maximum range of the sensors is 200m.  A similar sensor setup defines distances to surrounding traffic vehicles ($O$).}
    \label{fig:radar_sensor}
\end{figure}

The vector of 19 range finder sensors (\(R\)) comprises the distance between the track edge and the car centre within a range of 200 meters. Each value represents clockwise distances from 0 to 180 degrees to the car axis at 10-degree intervals. This is similar to how an automotive RADAR sensor is used to localize objects in the long-range vicinity of the autonomous vehicle. Figure \ref{fig:radar_sensor} illustrates this sensor setup. The same methodology defines the distances to surrounding traffic vehicles to define vector $O$, as Table \ref{tab:variables_parameters} explains.

\begin{table*}[htbp]
  \caption{System and Actuator Variables in the Driving Environment}
  \label{tab:variables_parameters}
  \centering
  \begin{subtable}[t]{\textwidth}
    \caption{State Variables}
    \label{subtab:state_variables}
    \centering
    \begin{tabularx}{\textwidth}{>{\raggedright\arraybackslash}X >{\centering\arraybackslash}X >{\raggedright\arraybackslash}X}
      \toprule
      \textbf{Variable} & \textbf{Range (unit)} & \textbf{Description} \\
      \midrule
      $\alpha_{axis}$ & $[-\pi, +\pi]$ (rad) & Angle between the car direction and the track axis. \\
      $D_{center}$    & $(-\infty, +\infty)$ (m) & Normalized distance from the car to the track axis (0 on axis, -1 at the right edge, +1 at the left edge). \\
      $z$             & $(-\infty, +\infty)$ (m) & Vertical distance of the car's mass centre from the track surface. \\
      $V_{x,y,z}$     & $(-\infty, +\infty)$ (km/h) & Speed vector in $\mathbb{R}^3$. \\
      $R$             & $[0,200]$ (m) & 19 range finder sensors: each returns the distance from the car to the track edge, sampling every 10° from -90° to +90° relative to the car's axis. \\
      $O$             & $[0,200]$ (m) & 19 range finder sensors for traffic vehicles with the same angular sampling. \\
      \bottomrule
    \end{tabularx}
  \end{subtable}
  
  \vspace{0.5cm}
  
  \begin{subtable}[t]{\textwidth}
    \caption{Actuator Variables}
    \label{subtab:actuator_variables}
    \centering
    \begin{tabularx}{\textwidth}{>{\raggedright\arraybackslash}X >{\centering\arraybackslash}X >{\raggedright\arraybackslash}X}
      \toprule
      \textbf{Variable} & \textbf{Range (unit)} & \textbf{Description} \\
      \midrule
      $\delta$ & $[-1, +1]$ & Steering value: -1 and +1 represent the right and left bounds, corresponding to 0.366519 rad. \\
      $a$      & $[0, 1]$  & Virtual acceleration pedal (0 = no gas, 1 = full gas). \\
      $b$      & $[0, 1]$  & Virtual braking pedal (0 = no brake, 1 = full brake). \\
      \bottomrule
    \end{tabularx}
  \end{subtable}
  
\end{table*}

\subsubsection{Physics-based Pure Pursuit Steering Model \label{section:pure_pursuit}}

The pure pursuit algorithm is a path-tracking algorithm that steers a vehicle along a desired path, given a set of reference points to follow. The algorithm calculates the steering angle required to follow the path based on the vehicle's current position and velocity \cite{Darbon2021}. It operates based on the concept of "pursuit," where a hypothetical point, known as the look-ahead point, is established ahead of the vehicle on the desired path. The objective is to steer the vehicle towards this point, thereby ensuring path tracking. The algorithm computes the optimal steering angle required to reach the look-ahead point by analyzing the geometric relationship between the vehicle and the path.

\begin{figure}[htbp]
    \centering
    \includegraphics[width=1\linewidth]{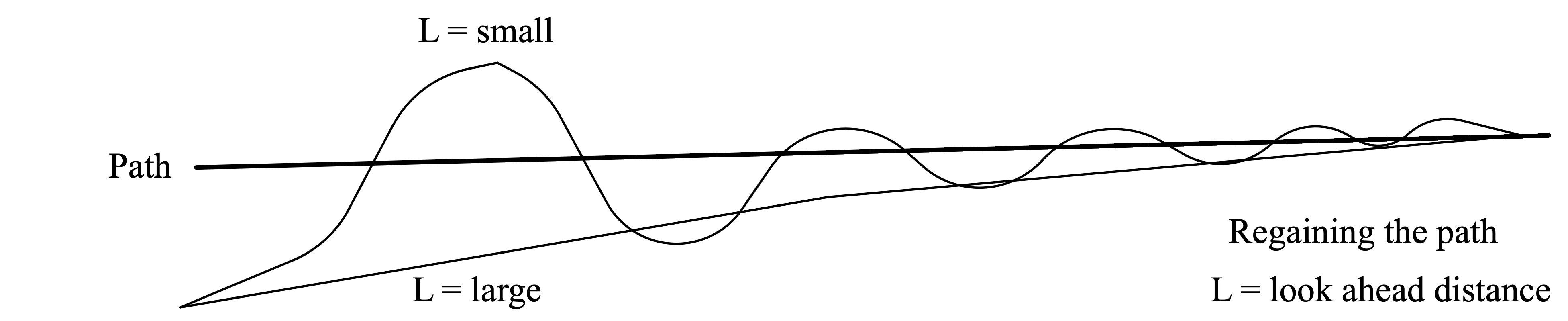}
    \caption{The effect of look-ahead distance magnitude on the behaviour of path convergence \cite{Degrave2019}}
    \label{fig:lookahead}
\end{figure}

Firstly, the algorithm estimates the position of the look-ahead point by projecting it along the desired path based on the vehicle's current position. Subsequently, the algorithm computes the curvature of the path at the look-ahead point using mathematical techniques such as interpolation or curve fitting. Finally, utilizing the vehicle's kinematic model, the algorithm calculates the required steering angle to navigate towards the look-ahead point. The look-ahead distance is a control parameter in the context of the algorithm and affects the outcome in two contexts:
1) Regaining a path, i.e., the vehicle is at a "large" distance from the path and must attain the path.  
2) Maintaining the path, i.e., the vehicle is on the path and wants to remain on the path.  
The effects of changing the parameter in the first problem are easy to imagine using the analogy to human driving. The longer look-ahead distances converge to the path more gradually and with less oscillation, as illustrated in Figure \ref{fig:lookahead}. It, therefore, depends on the state of the driving environment at any point in time to optimally set the look-ahead distance to manoeuvre the track safely.

\begin{figure}[htbp]
    \centering
    \includegraphics[width=0.5\linewidth]{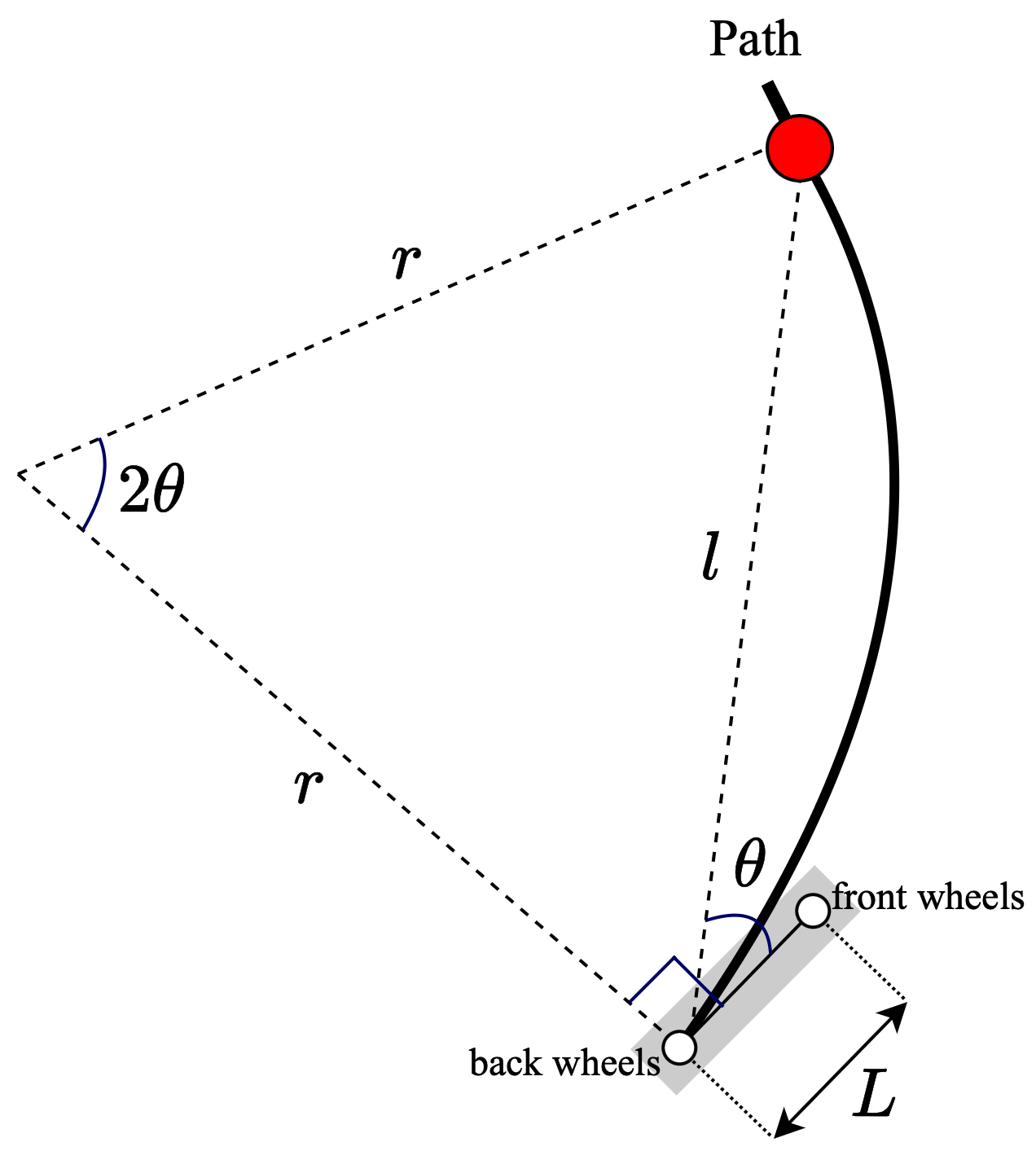}
    \caption{Geometric relationship between the circle of rotation, wheelbase of the vehicle (\(L\)), lookahead distance (\(l\)), and the heading difference (\(\theta\)).}
    \label{fig:pp_geometric}
\end{figure}

The Pure Pursuit model defines a reference point on the road as a vector of two observed scalar values: the heading difference (angle) to the direction of the reference point and the lookahead distance to the reference point. Formally, this can be represented as:
\begin{align*}
\label{eq:z}
\textbf{z} &= \begin{bmatrix}
           l \\
           \theta_{target}
     \end{bmatrix}
\end{align*}
where $\textbf{z}$ is the reference point, and $l$ and $\theta_{target}$ are the lookahead distance and heading difference accordingly. We assume that the vehicle follows the principle of a rigid body moving around a circle and, therefore, compute the required steering angle using this circle's instantaneous centre of rotation. Figure \ref{fig:pp_geometric} illustrates this concept. Consequently, from the arc of rotation depicted in the figure, we can use the law of sine to calculate the curvature, \(k\), of the path to the selected point as:

\begin{gather*} 
\frac{l}{\sin{2\theta}} = \frac{r}{\sin(\frac{\pi}{2} - \theta)}\\
\frac{l}{\sin\theta \cos\theta} = \frac{r}{\cos\theta}\\
\frac{l}{\sin\theta} = 2r\\
\end{gather*}
\begin{equation} \label{eq:2}
    k=\frac{1}{r} = \frac{2\sin\theta}{l}
\end{equation}

Here, \(r\) is the radius of rotation for the vehicle. Furthermore, we know from the kinematic bicycle model that the radius of rotation has an inverse relationship with the steering vehicle as:
\[r = L / \tan(\delta)\]
Here, \(\delta\) is the steering angle and \(L\) is the wheelbase of the vehicle (horizontal distance between rear and front wheel centers). Therefore, we can substitute \(r\) from the equation of curvature (Equation \ref{eq:2}) to get the required steering angle as:
\begin{equation} \label{eq:pure_pursuit}
    \delta = \arctan(\frac{2L\sin\theta}{l})
\end{equation}

\subsubsection{Physics Encoded Residual Feed-Forward Network \label{sec:pinn}}
Based on the approach introduced in section \ref{sec:approach}, we formulate our physics block as the mathematical operations calculating the target steering angle within the Pure Pursuit algorithm. As previously discussed in section \ref{section:pure_pursuit}, the selected reference point and look-ahead distance are key parameters that dictate the steering behaviour and are based on the geometry of the driving path ahead. Consequently, we formulate our learning block as a set of fully connected layers to output the optimal look-ahead distance $l$ and heading difference $\theta_{target}$ to represent the selected reference point $\textbf{z}$ on the road to follow (see Equation \ref{eq:z}). It is important to note that these variables are unknown in the input space (sensor information) and are thereby `discovered' by the learning blocks.

\begin{figure}[htbp]
    \centering
    \includegraphics[width=1\linewidth]{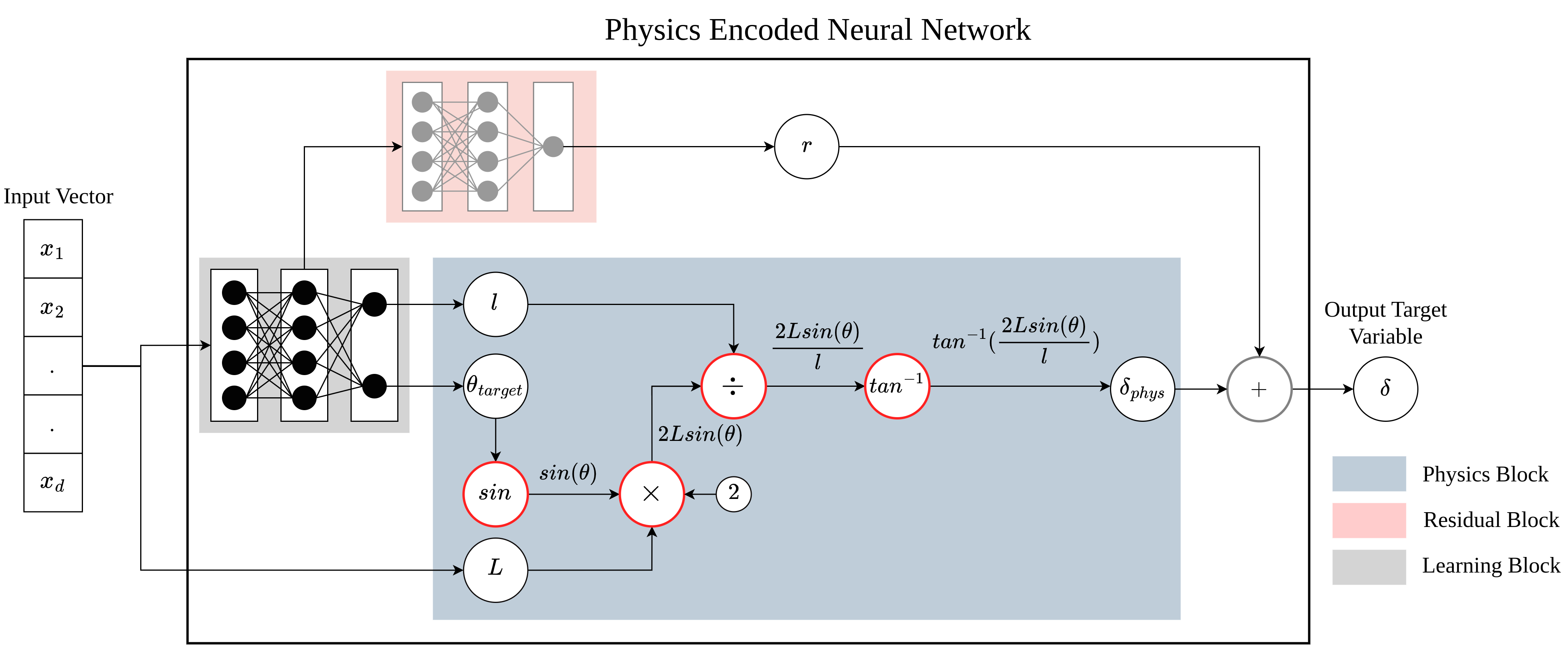}
    \caption{Physics Encoded Neural Network architecture for the steering model. The learning block predicts two unknown variables from the input space: $l$ (lookahead distance) and $\theta_{target}$ (heading difference to selected reference point). These variables are passed as input to the physics block to predict the steering angle $\delta$. The intermediate layer output from the learning block is passed to the residual block to predict the residual value to be added to $\delta$ for the final steering value }
    \label{fig:pinn}
\end{figure}

\begin{algorithm}
\caption{Two-Phased Behavior Cloning Using Physics-Encoded Neural Network}
\label{algo:1}
\begin{algorithmic}[1]

\State \textbf{Input:} Demonstration Data 1 ($\mathcal{D}_1$), Demonstration Data 2 ($\mathcal{D}_2$), Loss Threshold ($\ell_{\text{thresh}}$)
\State \textbf{Output:} Integrated Model ($M$)
\State Initialize Integrated Model  $M$ comprising three blocks $\mathcal{L}$, $\mathcal{R}$, $\mathcal{P}$
\State Initialize learning block $\mathcal{L}$ (learnable)
\State Initialize physics block $\mathcal{P}$ (non-learnable)
\State Initialize residual block $\mathcal{R}$ (learnable)
\State Initialize heuristic function $H$ for label generation
\State
\State \textbf{Phase I: Warm Start - Train Block $\mathcal{L}$ with Heuristic Labels}
\Repeat
    \State Sample mini-batch of state-action pairs from $\mathcal{D}_1$
    \State Generate labels using the heuristic function $H$
    \State Forward pass through block $\mathcal{L}$
    \State Compute loss ($\ell_{1}$) for predicted lookahead distance and target heading differences
    \State Backward pass and update the weights in block $\mathcal{L}$
\Until{convergence}
\State
\State \textbf{Phase II: Train Complete Model $M$}
\State Initialize block layers in $\mathcal{L}$ with weights from Phase I
\Repeat
    \State Sample mini-batch of state-action pairs from $\mathcal{D}_2$
    \State Forward pass through models $\mathcal{R}$ and $\mathcal{L}$ with static block $\mathcal{P}$
    \State Compute loss ($\ell_2$) for predicted steering
    \State Backward pass and update weights in blocks $\mathcal{R}$ and $\mathcal{L}$
\Until{$\ell_2 < \ell_{\text{thresh}}$}
\State
\State \Return Trained model $M$
\end{algorithmic}
\end{algorithm}

These variables are then fed to the physics block, a computational graph expressing the steering angle as a geometric combination of the lookahead distance and heading difference from the fully connected layers and the vehicle's wheelbase (available in the input space).  These include the operations in Equation \ref{eq:pure_pursuit} to calculate the target steering based on the input variables.

To formally define each block, let \textbf{\(\textbf{x}\)} be the vector of input parameters (explained in Table \ref{tab:variables_parameters}). Let us assume \(\mathcal{L}(\textbf{x},\phi)\) represents the function modelled by the learning block predicting the lookahead distance $l$ and target heading difference $\theta_{target}$, with known input variables $x$ and a set of learnable weights $\phi_{learn}$.  
\begin{gather*}
    \mathcal{L}: \mathbb{R}^d \times \mathbb{R}^m \rightarrow \mathbb{R}^2 \\
    \mathcal{L}(\textbf{x}, \phi_{learn}) = \begin{bmatrix}
           l \\
           \theta_{target}
           \end{bmatrix}
\end{gather*}
where $d$ is the number of state features in the input vector (from available state parameters shown in Table \ref{tab:variables_parameters}), and $m$ is the total number of weights in the fully connected layers.

Let us also assume our physics block $\mathcal{P}(\textbf{x},\mathcal{L})$ to predict the steering value $\delta_{phys}$ using the known input state and the intermediate unknown variables predicted by $\mathcal{L}$, using operators from Equation \ref{eq:pure_pursuit}.
\[\mathcal{P}(\textbf{x},\mathcal{L}) = \arctan(\frac{2L\sin\theta}{\mathcal{L}(x)})\]

We also define a residual block $\mathcal{R}(V;\phi_{res})$ comprising a set of fully connected layers with learnable weights $\phi_{res}$, which inputs a feature vector from an intermediate layer from the learning block $\mathcal{L}$ as input. The residual block predicts the residual target value $r$, which is added to the physics-predicted steering value $\hat{\delta}_{phys}$ to formulate the final steering value from the model. 
\[\hat{\delta} = \mathcal{P}(\textbf{x}, \mathcal{L}) + \textbf{r}\]
Based on the Behavior Cloning problem discussed in section \ref{sec:approach}, we can assume a set of demonstration data $T$ from an expert driving agent as i.i.d state-action pairs:
\[T = \{(\textbf{x}_1, \delta_1), (\textbf{x}_2, \delta_2) ... (\textbf{x}_n, \delta_n) \}\]
Here, $\textbf{x}_i$ and $\delta_i$ are observed input state and steering values from the expert driving agent. We can then define the predicted steering value (target value) based on the policy function $\pi$ as:
\[\hat{\delta} = \pi(\textbf{x}, \mathcal{L}, \mathcal{P}) \text{  where } \hat{a} \in \mathcal{A}\]
The loss function is based on Mean Squared Error (MSE), minimizing the error between the observed ($\delta$ ) and predicted ($\hat{\delta}$) steering angles. This is defined as:
\begin{equation*}
    L = \frac{1}{n}\sum_{i=1}^{n}(\delta_i-\hat{\delta}_i)^2
\end{equation*}
We use the \textit{Mini-Batch Gradient Descent} algorithm and \textit{Adam Optimizer} to optimize the loss function, where $n$ is any arbitrary batch size for each sample in the gradient descent algorithm.  The gradients for the learnable weights in $\phi_{learn}$ are calculated by the chain of function derivatives that involve the geometric and kinematic operators from Equation \ref{eq:2}. This allows our feed-forward network to learn the optimal lookahead while being forced to comply with the geometric and kinematic aspects of the steering problem dictated by the pure pursuit algorithm. This is done by constraining the calculation of gradients, which are calculated as an explicit part of the equations based on the pure pursuit algorithm. On the other hand, the residual block provides an alternative pathway for the gradients to flow back to the intermediate layers of the learning block. This allows the loss function to converge and optimize the weights in the residual and, more importantly, the learning block.

\subsubsection{Training on Demonstration Data}\label{sec:training_method}

We train our physics-encoded neural network model on the demonstration data $T$ in two sequential phases. The first phase involves a \textit{warm-start} phase where the learning block is trained separately to predict appropriate lookahead distance ($l$) and target heading difference ($\theta$) values. We generate two batches of data from our simulation (details in section \ref{sec:training_data}). The first batch is generated from the agent driving on empty segments of tracks, and the second batch is generated from the agent driving amongst other traffic vehicles. The sequence of steps in the training process is defined in Algorithm \ref{algo:1}.

\paragraph{Phase I: Warm Start \label{sec:warm_start}}

In this phase, the learning block is trained separately on the first dataset ($\mathcal{D}_1$) collected from empty road scenarios. We define a heuristic function to provide the learning block with proposed values of lookahead distance ($l$) and target heading difference ($\theta_{target}$) for each data point. The heuristic function selects the reference point to be in the direction of the furthest point on the road boundary from the vehicle, using the range-finding sensor vector \(R\) (refer to Table \ref{tab:variables_parameters}).

\begin{figure}[htbp]
    \centering
    \includegraphics[width=0.75\linewidth]{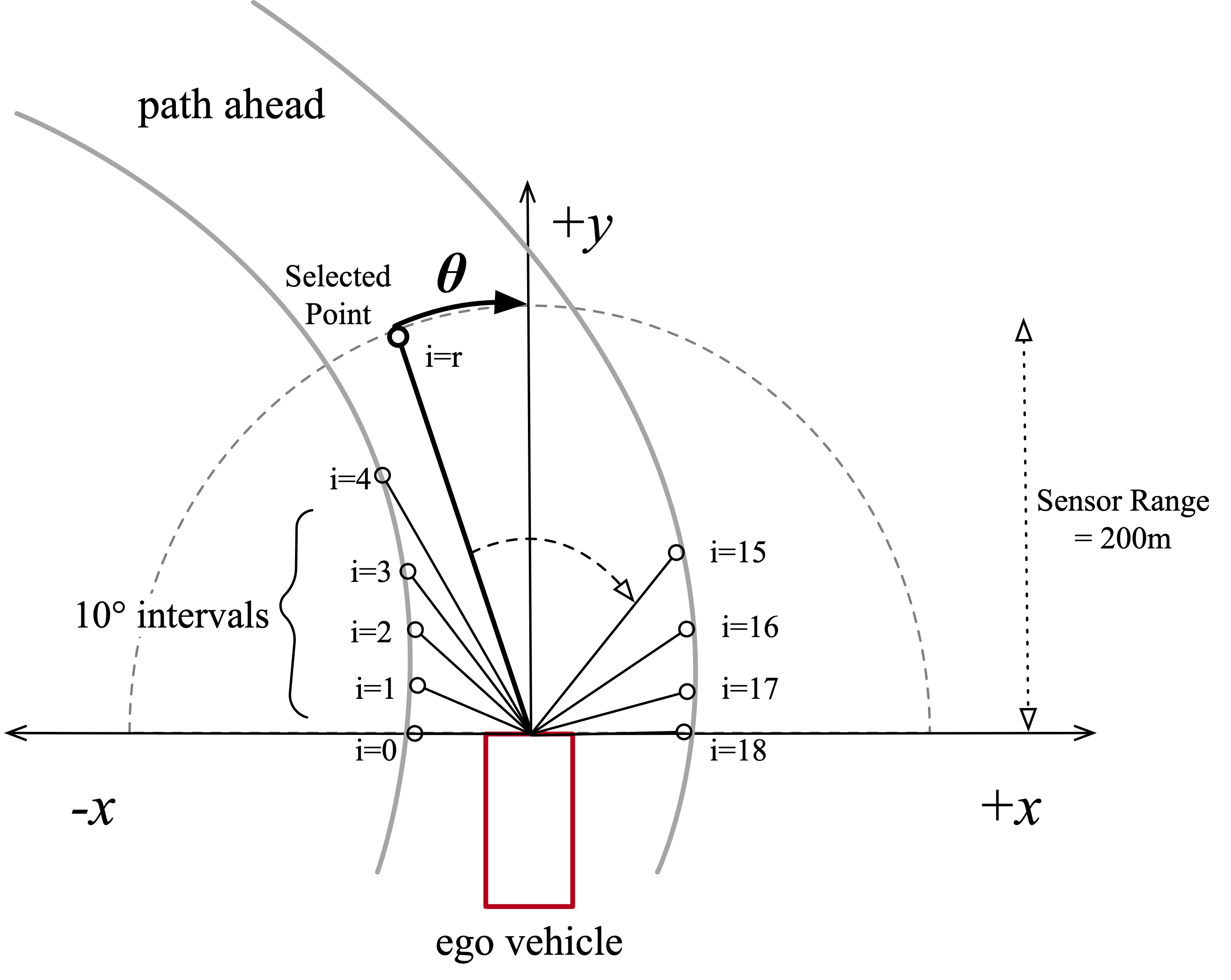}
    \caption{Proposed method to find the reference direction for vehicle, based on the maximum range finder sensor values}
    \label{fig:range_finder}
\end{figure}

Figure \ref{fig:range_finder} illustrates this method. The range finder sensors can be considered to run clockwise from the vehicle's horizontal plane, from the left side of the vehicle, forming a semi-circle axis in front of the vehicle. The sensors capture values every 10 degrees on this axis, clipped to a maximum value of 200m. These values depict the presence of a road boundary at the corresponding angle value. Formally, for \(R_t\) as the vector of range finder sensors from a data point $t$, we can define:
\[i = argmax(R_t)\]
where $i$ is the index of the maximum distance value in vector $R_t$. If there are multiple maximum values, the index of the first of these values is picked. Using this index $i$, we can calculate $\theta_t$ (in degrees) and $l_t$, for any given data point $t$, as follows:
\begin{gather*}
\theta_{t} = 10i - 90\\
l_t = R_t^i
\end{gather*}

We then use the output from the heuristic function to provide target labels for the learning block. The loss function is based on MAE (Mean Absolute Error), minimizing the error between the heuristic output and predicted values from the learning block. This is optimized using a simple Mini-Batch Gradient Descent with \textit{Adam Optimizer} until an arbitrary point of convergence where the mean loss function value across mini-batches falls below a threshold. 

\paragraph{Phase II: Integrated Training}

We refer to the complete model $M$ as the connected neural network with the learning ($\mathcal{L}$), learning ($\mathcal{R}$) and learning ($\mathcal{P}$) blocks, as shown in Figure \ref{fig:pinn}. The learned weights from Phase I are used to initialize the learning block. We then train $M$ on the second dataset ($\mathcal{D}_2$) comprising traffic scenarios, with the observed steering values as target values from which the model can learn. The training is based on the explanation in section \ref{sec:pinn}.

\FloatBarrier

\subsection{Case Study 2: Climate Digital Twins for CO$_2$ Emissions and Gap-Filling in Net Ecosystem Exchange Modelling}\label{sec:climate_digital_twins}

The application of digital twin technology to climate systems has emerged as a promising avenue for enhanced monitoring and forecasting of atmospheric CO$_2$ dynamics. Digital twins, which provide a virtual representation of real-world processes, are particularly valuable in climate science where continuous, high-quality observational data are critical yet often incomplete. Flux towers continuously monitor key atmospheric scalars such as Net Ecosystem Exchange (NEE), Latent Heat, and Sensible Heat \cite{zhuclementetal2022} using the eddy covariance (EC) method. However, these measurements frequently exhibit gaps due to instrumentation failures, power shortages, or environmental interferences \cite{reichsteinetal2005, moffatet2007}.

In the context of a digital twin for climate applications, robust gap-filling methods are essential not only for data completion but also for accurately capturing the underlying physical processes governing NEE. Traditional gap-filling techniques—such as Non-Linear Regression, Look-Up Tables, and Marginal Distribution Sampling—have shown reasonable performance for short-duration gaps, yet they struggle with longer intervals of missing data \cite{moffatet2007}. More recent approaches, such as the Random Forest Robust (RFR) method, have enhanced performance over longer gaps by improving the statistical robustness of the estimates, although challenges remain for nighttime measurements \cite{zhuclementetal2022}.

Integrating the underlying physics of NEE into the gap-filling process can address these challenges. As demonstrated in previous work, NEE dynamics can be modelled as a Stochastic Differential Equation (SDE) that decomposes the signal into a deterministic drift component—governed by temperature-dependent ecosystem respiration following an Arrhenius-type law \cite{lasslopetal2010, lloydtaylor1994}—and a stochastic diffusion term representing measurement noise, assumed to be Gaussian \cite{White2008686}. Embedding this SDE within a digital twin framework allows the inherent physics of ecosystem CO$_2$ exchange to directly inform the data reconstruction process, leading to more reliable and interpretable predictions.

Our proposed Physics Encoded Neural Network (PERNN) framework naturally extends to this climate modelling use case. In this setting, the physics blocks encapsulate the SDE-based model of NEE, capturing both the deterministic Arrhenius-type response of ecosystem respiration and the inherent stochastic fluctuations. Complementary learning blocks infer latent variables and compensate for unobserved dynamics during periods of data absence. Moreover, residual blocks not only aid gradient flow during training but also facilitate the integration of these physics-informed estimates with empirical corrections, ensuring that the digital twin maintains fidelity to both the established physical laws and the observational data.

This hybrid architecture enhances gap-filling performance while simultaneously improving the interpretability of NEE trends, making it a robust tool for climate digital twins. By directly incorporating physical models into the neural network, our approach bridges the gap between purely data-driven methods and traditional physics-based modeling, thereby providing a scalable and generalizable framework for environmental monitoring and forecasting \cite{Tao2019, Jones2020}.

\subsubsection{Flux Tower Data}\label{sec:flux_data}

The observational data employed in this study were collected from a flux tower located in East Anglia, UK \cite{cummingdata2012022} from 2012 to 2019. The tower is equipped with an eddy covariance (EC) system designed to measure key atmospheric scalars including Net Ecosystem Exchange (NEE), sensible heat flux density (H), momentum flux (Tau), relative humidity (RH), vapor pressure deficit (VPD), global radiation ($\text{R}_\text{g}$), friction velocity (Ustar), soil temperatures (Tsoil1 and Tsoil2), and air temperature at 2\,m ($\text{T}_\text{air}$). The dataset spans eight years, from 2012 through the end of 2019, with measurements recorded at 30-minute intervals.

These high-frequency measurements provide a rich temporal resolution of the ecosystem's CO$_2$ exchange dynamics. However, the dataset also presents common challenges such as missing values caused by power shortages, sensor malfunctions, or adverse weather conditions. These data gaps can range from short intervals to periods extending over several months, emphasizing the necessity for robust gap-filling methods that can reliably reconstruct missing measurements while preserving the underlying physical relationships.

A detailed description of the measured variables is provided in Table~\ref{tab:climate_variables}. In this study, particular attention is given to the NEE variable, which is central to understanding the carbon dynamics of the ecosystem. Given that latent heat (L) is measured by the same instrument as NEE, any gaps in NEE often correspond to gaps in latent heat measurements; hence, latent heat is excluded from the analysis to maintain consistency. Additionally, time-based attributes such as season, hour, day of week, month, and day of year are incorporated to account for diurnal and seasonal variations.

\begin{table}[htbp]
  \caption{List of variables from the flux data \cite{cummingdata2012022}}
  \label{tab:climate_variables}
  \centering
  \begin{tabularx}{\textwidth}{>{\raggedright\arraybackslash}X >{\centering\arraybackslash}X >{\raggedright\arraybackslash}X}
    \toprule
    \textbf{Variable} & \textbf{Units} & \textbf{Description} \\
    \midrule
    NEE    & $\upmu\text{mol C m}^{-2}\text{s}^{-1}$ & Net ecosystem $\ce{CO_2}$ exchange flux density before data gap-filling. \\[3pt]
    H      & $\text{W m}^{-2}$                        & Sensible heat flux density. \\[3pt]
    Tau    & $\text{kg m}^{-1}\text{s}^{-2}$           & Momentum flux. \\[3pt]
    RH     & \%                                       & Relative humidity at 2\,m. \\[3pt]
    VPD    & $\text{HPa}$                             & Vapor pressure deficit. \\[3pt]
    $\text{R}_\text{g}$ & $\text{W m}^{-2}$              & Global radiation. \\[3pt]
    Ustar  & $\text{m s}^{-1}$                        & Friction velocity. \\[3pt]
    Tsoil1 & $^{\circ}\text{C}$                       & Soil temperature at a depth of 0.05\,m. \\[3pt]
    Tsoil2 & $^{\circ}\text{C}$                       & Soil temperature at a depth of 0.05\,m. \\[3pt]
    $\text{T}_\text{air}$ & $^{\circ}\text{C}$             & Air temperature at 2\,m. \\
    \bottomrule
  \end{tabularx}
\end{table}

\subsubsection{The Net Ecosystem Exchange Model}\label{sec:neemodels}

The Net Ecosystem Exchange (NEE) quantifies the net flux of CO$_2$ between an ecosystem and the atmosphere. It is defined as the balance between ecosystem respiration and photosynthetic uptake. Specifically, NEE can be expressed as:
\begin{equation}
    \text{NEE}_t = \text{R}_{\text{eco},t} - \text{GPP}_t,
    \label{nee_equation_rephrased}
\end{equation}
where $\text{R}_{\text{eco},t}$ represents the rate of CO$_2$ release due to biological respiration, and $\text{GPP}_t$ denotes the Gross Primary Production associated with photosynthesis. By convention, negative NEE values indicate net CO$_2$ uptake by the ecosystem \cite{lasslopetal2010, keenan2019}.

For periods when global radiation is insufficient (i.e., $\text{R}_\text{g} < 20\,\text{W.m}^{-2}$), photosynthesis is assumed to be negligible, resulting in $\text{GPP}_t \approx 0$. Under these conditions, the measured NEE predominantly reflects ecosystem respiration. The temperature dependence of ecosystem respiration is commonly modelled using an Arrhenius-type expression. Accordingly, the respiration term at night is formulated as:
\begin{equation}
    \text{R}_{\text{eco},t} = r_{night} \exp\!\Biggl( E_0 \Bigl( \frac{1}{T_{ref}-T_0} - \frac{1}{\text{T}_{\text{air},t}-T_0} \Bigr) \Biggr),
    \label{eq:neenighttimemodel_rephrased}
\end{equation}
where $r_{night}$ (in $\upmu\text{mol C m}^{-2}\text{s}^{-1}$) is the baseline respiration rate at the reference temperature $T_{ref}=15^\circ\text{C}$, $E_0$ (in $^\circ\text{C}$) is a constant that characterizes temperature sensitivity, $\text{T}_{\text{air},t}$ is the ambient air temperature at time $t$, and $T_0$ is a fixed offset (typically $-46.02^\circ\text{C}$) \cite{lasslopetal2010}. For consistency, the parameter $r_{night}$ is periodically updated—commonly every five days—using estimations derived from 15-day windows of historical observations.

This formulation of NEE as described above forms the foundation of our modelling approach.

\subsubsection{ODE-based Net Ecosystem Exchange Model}\label{sec:nee_ode}

In our approach, the dynamics of Net Ecosystem Exchange (NEE) are modeled using an ordinary differential equation (ODE) that captures the temperature-dependent behaviour of ecosystem respiration. Specifically, the evolution of NEE over time is expressed as
\begin{equation}
    \frac{d\,\text{NEE}_t}{dt} = \frac{d}{d\,\text{T}_{\text{air},t}} \text{R}_{\text{eco},t}(\text{T}_{\text{air},t}) \,\frac{d\,\text{T}_{\text{air},t}}{dt},
    \label{eq:nee_ode_full}
\end{equation}
where $\text{R}_{\text{eco},t}(\text{T}_{\text{air},t})$ denotes the ecosystem respiration rate, modeled via an Arrhenius-type function given in Equation \ref{eq:neenighttimemodel_rephrased}.

The derivative of the respiration function with respect to air temperature is given by
\begin{equation}
    \frac{d}{d\,\text{T}_{\text{air},t}} \text{R}_{\text{eco},t}(\text{T}_{\text{air},t}) = \frac{E_0}{(\text{T}_{\text{air},t} - T_0)^2}\,\text{R}_{\text{eco},t},
    \label{eq:dR_dT}
\end{equation}
and the rate of change of the ambient air temperature is modeled as
\begin{equation}
    \frac{d\,\text{T}_{\text{air},t}}{dt} = \pi\,\frac{\Delta \text{T}_{\text{air},t}}{t_{day}}\,\sin\!\Bigl(2\pi\,\frac{t-t_{T_{\max}}}{t_{day}}\Bigr).
    \label{eq:dT_dt}
\end{equation}

By incorporating Equation~\ref{eq:nee_ode_full} into our Physics Encoded Residual Neural Network (PERNN) framework, we directly embed the known temperature-driven dynamics of ecosystem respiration into the digital twin. This integration ensures that while the neural network components can infer latent variables and correct for discrepancies, the overall model remains grounded in the established physical relationships governing CO$_2$ exchange.

\subsubsection{Physics Encoded Residual Neural Network for NEE ODE}\label{sec:nee_pernn}

In our NEE modeling application, we extend the Physics Encoded Residual Neural Network (PERNN) framework to forecast the evolution of Net Ecosystem Exchange (NEE) by integrating the known deterministic dynamics of ecosystem respiration into a digital twin. The goal is to predict the rate of change of NEE, \(\frac{d\,\text{NEE}}{dt}\), and subsequently obtain the forecasted NEE at the next time step via
\begin{equation}
    \text{NEE}_{t+1} = \text{NEE}_t + \widehat{\frac{d\,\text{NEE}}{dt}} \cdot \Delta t,
    \label{eq:nee_forecast_final}
\end{equation}
where \(\Delta t\) is 30 minutes, consistent with the measurement interval of the flux data.

Let \(\mathbf{x} \in \mathbb{R}^d\) denote the vector of input features, where \(d\) corresponds to the number of variables defined in Table~\ref{tab:variables_parameters} (e.g., \(\text{T}_{\text{air}}\), \(\text{R}_\text{g}\), RH, VPD, etc.). These features encapsulate the meteorological and temporal context required for the NEE model.

The learning block, denoted by \(\mathcal{L}(\mathbf{x},\phi_{learn})\), is implemented as a set of fully connected layers. It processes the input state \(\mathbf{x}\) and infers the intermediate variables that are not directly observable:
\[
\mathcal{L}(\mathbf{x}, \phi_{learn}) = 
\begin{bmatrix}
    E_0 \\
    r_{b\_night} \\
    \left(\frac{dT}{dt}\right)_{pred}
\end{bmatrix}.
\]
Here, \(E_0\) represents the temperature sensitivity, \(r_{b\_night}\) is the base respiration at the reference temperature, and \(\left(\frac{dT}{dt}\right)_{pred}\) is the predicted rate of change of ambient air temperature. These variables are subsequently fed to the physics block.

The physics block, \(\mathcal{P}(\mathbf{x}, \mathcal{L})\), is a non-trainable computational graph that encapsulates the physical operators derived from the ODE governing NEE dynamics. Using the predicted temperature derivative \(\left(\frac{dT}{dt}\right)_{pred}\) from the learning block, the physics block computes the change in NEE based on Equation \ref{eq:nee_final_ode} as:
\begin{equation}
    \mathcal{P}(\mathbf{x}, \mathcal{L}) = \frac{d\,\text{NEE}_t}{dt} = \frac{E_0}{(\text{T}_{\text{air},t}-T_0)^2}\,\text{R}_{\text{eco},t}\,\left(\frac{dT}{dt}\right)_{pred}.
    \label{eq:nee_ode_pernn}
\end{equation}

To compensate for any discrepancies between the physics-based prediction and the actual dynamics observed in the data, a residual block \(\mathcal{R}(V;\phi_{res})\) is employed. Here, \(V\) represents an intermediate feature vector extracted from the learning block. The residual block predicts an additive correction \(r\) such that the final prediction of the rate of change in NEE is:
\begin{equation}
    \widehat{\frac{d\,\text{NEE}}{dt}} = \mathcal{P}(\mathbf{x}, \mathcal{L}) + r.
    \label{eq:nee_final_ode}
\end{equation}

\paragraph{Ground-Truth Computation}
Ground-truth values for the rates of change are computed from the flux tower measurements using a right-sided first-order finite difference approximation. Specifically, for NEE:
\begin{equation}
    \frac{d\,\text{NEE}_t}{dt} \approx \frac{\text{NEE}_{t+1} - \text{NEE}_t}{\Delta t},
    \label{eq:nee_ground_truth}
\end{equation}
and similarly for air temperature:
\begin{equation}
    \frac{d\,T_{air,t}}{dt} \approx \frac{T_{air,t+1} - T_{air,t}}{\Delta t},
    \label{eq:dT_ground_truth}
\end{equation}
with \(\Delta t = 30\) minutes.

\paragraph{Parameter Estimation for \(E_0\) and \(r_{night}\)}
Since the flux tower data does not directly provide ground-truth values for the parameters \(E_0\) and \(r_{night}\), these are estimated using the REddyProc partitioning algorithm \cite{Wutzler2018}. REddyProc applies a non-linear regression approach based on the methodology of \cite{reichsteinetal2005} to partition the flux data, particularly during nighttime conditions when photosynthetic uptake is negligible. In this context, the Lloyd-and-Taylor model \cite{lloydtaylor1994} is fitted to the scatter of \(\text{NEE}\) and \(\text{T}_{\text{air},t}\) values, providing estimates within established ranges (e.g., \(E_0 \in [50, 400]\) and \(r_{night} > 0\)) as discussed in \cite{lasslopetal2010}.

\paragraph{Loss Function and Two-Phased Training Procedure}
The model is trained to minimize the Mean Squared Error (MSE) loss, with two distinct phases. We use the \textit{Mini-Batch Gradient Descent} algorithm and \textit{Adam Optimizer} to optimize the loss function, where $n$ is any arbitrary batch size for each sample in the gradient descent algorithm.

\begin{itemize}
    \item \textbf{Phase I: Warm Start:} In this phase, the learning block \(\mathcal{L}\) is pre-trained using heuristic labels generated by inverting Equations~\ref{eq:nee_ground_truth} and \ref{eq:dT_ground_truth}. The MSE loss is computed solely on the intermediate variables:
    \begin{equation}
        L_1 = \frac{1}{n}\sum_{i=1}^{n}\Biggl[\Bigl(\widehat{E_0}^{(i)} - E_0^{(i)}\Bigr)^2 + \Bigl(\widehat{r}_{b\_night}^{(i)} - r_{b\_night}^{(i)}\Bigr)^2 + \Bigl(\widehat{\left(\frac{dT}{dt}\right)}^{(i)} - \left(\frac{dT}{dt}\right)^{(i)}\Bigr)^2\Biggr],
        \label{eq:mse_phase1}
    \end{equation}
    where \(n\) is the number of samples in the mini-batch.
    
    \item \textbf{Phase II: Integrated Training:} In the second phase, the full model \(M\) — comprising the learning block \(\mathcal{L}\), the fixed physics block \(\mathcal{P}\), and the residual block \(\mathcal{R}\) — is trained end-to-end. The MSE loss in this phase is computed on both the intermediate variables and the forecasted \(\text{NEE}_{t+1}\):
    \begin{equation}
\begin{split}
L_2 &= \frac{1}{N}\sum_{i=1}^{N} \Biggl\{
\Bigl(\widehat{\text{NEE}}_{t+1}^{(i)} - \text{NEE}_{t+1}^{(i)}\Bigr)^2 \\
&\quad + \Bigl(\widehat{E_0}^{(i)} - E_0^{(i)}\Bigr)^2 \\
&\quad + \Bigl(\widehat{r}_{b\_night}^{(i)} - r_{b\_night}^{(i)}\Bigr)^2 \\
&\quad + \Bigl(\widehat{\left(\frac{dT}{dt}\right)}^{(i)} - \left(\frac{dT}{dt}\right)^{(i)}\Bigr)^2
\Biggr\}.
\end{split}
\end{equation}
    where \(\widehat{\text{NEE}}_{t+1}^{(i)}\) is the forecasted value computed via Equation~\ref{eq:nee_forecast_final}.
\end{itemize}

\begin{figure}[htbp]
    \centering
    \includegraphics[width=1\linewidth]{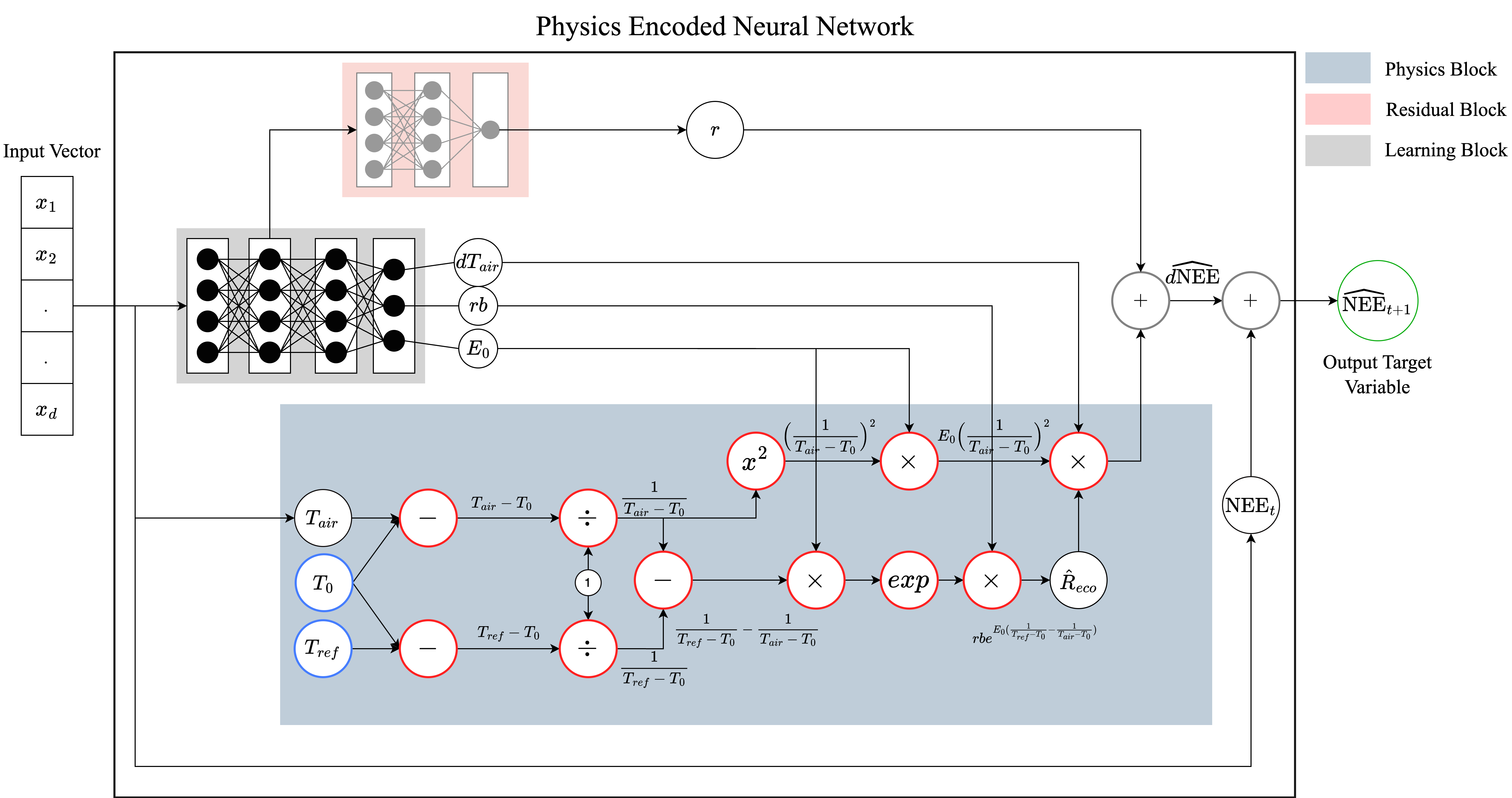}
    \caption{Physics Encoded Neural Network architecture for NEE forecasting. The learning block predicts three intermediate variables: $E_0$ (temperature sensitivity), $rb_{night}$ (base respiration) and $dT_{air}$ (change in air temperature). These variables are passed as input to the physics block based on the NEE ODE to predict $d\text{NEE}$ (change in NEE). The intermediate layer output from the learning block is passed to the residual block to predict the residual value to be added to $d\text{NEE}$ for the final prediction of the time derivative $\widehat{d\text{NEE}}$. This is then added to the current $\text{NEE}_t$ to output $\widehat{\text{NEE}}_{t+1}$.} 
    \label{fig:nee_pernn}
\end{figure}

Figure~\ref{fig:nee_pernn} illustrates the overall PERNN architecture for modeling the NEE ODE, and Algorithm~\ref{algo:nee_training} details the two-phased training procedure.

\begin{algorithm}
\caption{Two-Phased Training for PERNN for NEE}
\label{algo:nee_training}
\begin{algorithmic}[1]
\State \textbf{Input:} Training dataset \(\mathcal{D}\), Loss threshold \(\ell_{\text{thresh}}\)
\State \textbf{Output:} Trained PERNN model \(M\)
\State Initialize model \(M\) with:
    \begin{itemize}
        \item Learning block \(\mathcal{L}\) (learnable)
        \item Physics block \(\mathcal{P}\) (non-learnable)
        \item Residual block \(\mathcal{R}\) (learnable)
    \end{itemize}
\State
\State \textbf{Phase I: Pre-train Learning Block}
\Repeat
    \State Sample a mini-batch from \(\mathcal{D}\)
    \State Forward pass through \(\mathcal{L}\) to predict intermediate variables: \(\widehat{E_0}\), \(\widehat{r}_{b\_night}\), \(\widehat{\left(\frac{dT}{dt}\right)}\)
    \State Compute MSE loss \(L_1\) on these variables:
    \[
    L_1 = \text{MSE}\Bigl(\begin{bmatrix}\widehat{E_0} \\ \widehat{r}_{b\_night} \\ \widehat{\left(\frac{dT}{dt}\right)}\end{bmatrix},\begin{bmatrix}E_0 \\ r_{b\_night} \\ \left(\frac{dT}{dt}\right)\end{bmatrix}\Bigr)
    \]
    \State Backpropagate and update weights in \(\mathcal{L}\)
\Until{\(L_1\) converges}
\State
\State \textbf{Phase II: Train Integrated Model}
\Repeat
    \State Sample a mini-batch from \(\mathcal{D}\)
    \State Forward pass through \(\mathcal{L}\), then through \(\mathcal{P}\) (fixed), and finally through \(\mathcal{R}\) to obtain forecasted \(\widehat{\text{NEE}}_{t+1}\)
    \State Compute MSE loss \(L_2\) on both the intermediate variables and \(\widehat{\text{NEE}}_{t+1}\):
    \[
    L_2 = \text{MSE}\Biggl(\begin{bmatrix}\widehat{E_0} \\ \widehat{r}_{b\_night} \\ \widehat{\left(\frac{dT}{dt}\right)} \\ \widehat{\text{NEE}}_{t+1}\end{bmatrix},\begin{bmatrix}E_0 \\ r_{b\_night} \\ \left(\frac{dT}{dt}\right) \\ \text{NEE}_{t+1}\end{bmatrix}\Biggr)
    \]
    \State Backpropagate and update weights in \(\mathcal{L}\) and \(\mathcal{R}\)
\Until{\(L_2 < \ell_{\text{thresh}}\)}
\State
\State \Return Trained model \(M\)
\end{algorithmic}
\end{algorithm}

\section{Experiments \label{sec:experiments}}

In this section, we present a comprehensive set of experiments and analyses that build upon the theoretical foundations described in the preceding sections. For the simulation-based autonomous vehicle steering dynamics use case—where the available prior physics is simple and only partially known—we evaluate the performance of our approach in terms of generalizability, data efficiency, and model complexity. In the climate digital twin application, we assess our method's ability to accurately model Net Ecosystem Exchange (NEE) behavior over time and effectively perform gap-filling, comparing its performance against state-of-the-art techniques. These experiments serve to validate the advantages of our approach in both controlled simulation environments and real-world scenarios.

\subsection{Autonomous Vehicle Simulation \label{sec:driving_results}}

In this section, we present details on the experimentation and results on the problem of modelling steering dynamics in a simulated autonomous vehicle problem. We compare our Physics Encoded Residual Neural Network (PERNN) approach against two methods. The first is based on traditional fully connected neural networks (FCNN) to serve as a baseline for a purely data-driven method. The second method is based on physics-regularized neural networks which are currently one of the most popular generic physics-informed machine learning methods in literature. This is based on applying soft constraints to the learning process by appropriately penalizing the loss function of neural networks. Assuming an external physics model \(P\), input variables \(\textbf{x}\), ground truth target variable \(y\)  and any feed-forward neural network \(f\), the loss function of such a physics-informed network is:

\[L = L_{data} + L_{phy}\]
where \(L_{data}\) can be any function measuring the similarity between the predicted variable \(f(\textbf{x})\) and the target variable \(y\), while \(L_{phy}\) can be any  function measuring the similarity between the predicted variable \(f(\textbf{x})\) and the output from the physics model \(P(\textbf{x})\). \(L_{phy}\) can also represent the degree to which the neural network output satisfies the constraint on the system, which can be an equation such as a partial differential equation. The loss function can also be represented as a weighted sum of \(L_{data}\) and \(L_{phy}\), where the weight value can be represented as a hyperparameter or learnable parameter. In the context of the experimentation in this section, we do not employ a weighted sum to calculate the loss term. Since it can be argued that the extra loss term \(L_{phy}\) ``informs" the neural network of the physical properties, we will refer to this approach as Physics Informed Neural Network (PINN) throughout the experimentation section ahead.

The comparisons between the methods presented herewith are drawn in terms of regression metrics on the test dataset and driving performance inside the simulation on unseen tracks. We also analyse the unknown intermediate variables: lookahead distance and heading difference to reference points chosen by our trained PERNN-based agent in different driving scenarios.

\subsubsection{
Data Generation}\label{sec:training_data}

To generate data in TORCS as expert demonstrations for our experimentation, we use the implementation of an advanced heuristic-based agent provided by the simulation based on \textit{Ahura: A Heuristic-Based Racer for the Open Racing Car Simulator} \cite{Bonyadi2017}. The details of this implementation are irrelevant to the approach explained in this paper since, theoretically, this expert system can be any high-fidelity and accurate driving model. We generate state and actuator values at each time instance on each of the 36 race tracks in the simulation, corresponding to two driving scenarios: empty road and traffic bourne. We refer to these two collected datasets as $\mathcal{D}_1$ and $\mathcal{D}_2$.

\begin{figure}[htbp]
    \centering
    \includegraphics[width=0.75\linewidth]{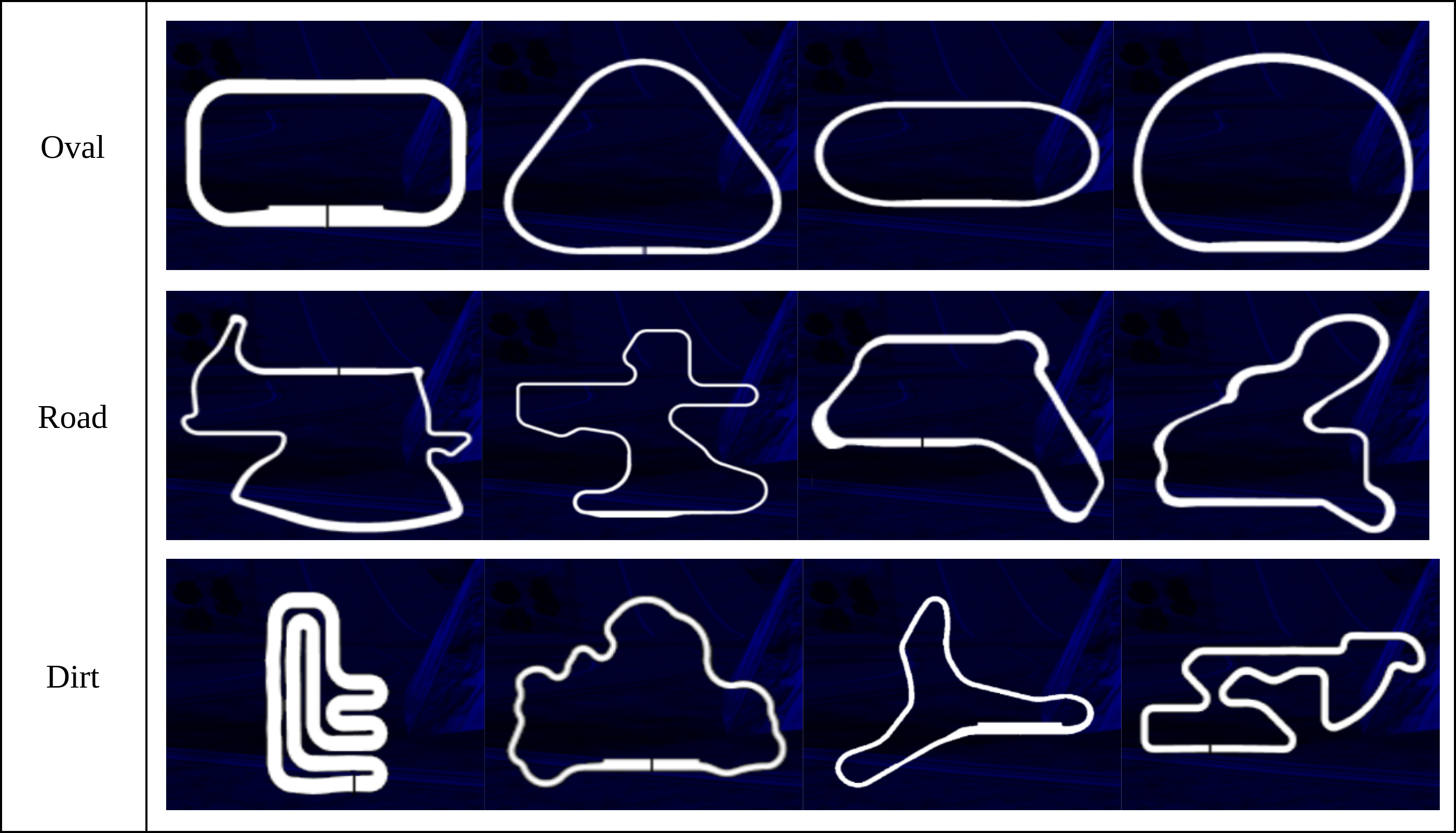}
    \caption{Sample track maps representing geometrical variations in each of the three track categories}
    \label{fig:track_geometry}
\end{figure}

The race tracks are divided into three categories: road, dirt and oval speedway tracks based on variations in track geometry. Most of the oval speedway tracks offer relatively simple geometry compared to the more complex geometry on the road tracks in terms of sharpness and variation in the directions of turns. The dirt tracks offer significantly tougher terrain owing to sharper turns and high variation in the altitude values leading to dips and troughs, represented as the $z$ value in Table \ref{tab:variables_parameters}.  Figure \ref{fig:track_geometry} shows a sample of maps representing geometrical variations in each of the three track categories. 

We reserve the data from a set of tracks as our test cases from both $\mathcal{D}_1$ and $\mathcal{D}_2$, where we analyze the behaviour of the trained models in the simulation. These include all dirt tracks and several road tracks. The goal is to test how well each approach performs in navigating new geometrical scenarios and their consequent capability for error recovery. 

\subsubsection{Experimental Setup}\label{sec:results}

We test four different agents in our experimentation. The first agent is based on the model trained using our Physics Encoded Residual Neural Network (PERNN) explained earlier in the paper. The second agent is based on the same framework as our method but without the residual block. This will help us illustrate the effectiveness of the residual-based architecture. We will refer to this model as Physics Encoded Neural Network (PENN). The third agent, as a benchmark, is based on a model trained using a conventional feed-forward neural network (FCNN) with a set of fully connected layers and ReLU activation functions. Finally, the fourth type of agent is trained using a physics-informed neural network (PINN) having the same base architecture as the FCNN. The output from all networks is a single steering value based on the state information provided to the model. 

For the PINN approach, the Pure Pursuit model serves as the physics model regularizing the loss function (discussed at the beginning of section \ref{sec:driving_results}). The Pure Pursuit model takes as input appropriate values of lookahead distance ($l$) and target heading difference ($\theta_{target}$), and since there is no concept of learning blocks predicting intermediate variables in the PINN approach, we use our heuristic function from section \ref{sec:warm_start} to generate these values for each data point in the training data. This means that the regularizing term \(L_{phy}\) is calculated for empty road scenarios where the logic behind the heuristic function holds. For the traffic-based data points, \(L_{phy}\) is set to zero to not influence the loss value for traffic-based scenarios.

For experimental coherence, we use a simple PID controller to control the speed and brake of the vehicle and a basic functional logic to control the gears based on the vehicle's speed. The controller's implementation details are irrelevant to the experiment results since they are kept constant for both driving agents. 

We compare three different FCNN models with our PERNN model regarding mean absolute error (MAE) on the test dataset, average distance travelled, and average jerk in predicted steering on the test tracks during live simulation. The three FCNN models are trained on the collective data from $\mathcal{D}_1$ and $\mathcal{D}_2$, comprising a single phase of training, as opposed to the two-phased setup for the PERNN model (see section \ref{sec:training_method}). The three FCNN models differ in architectural complexity and training data size (number of tracks comprised in training data) as shown in Table \ref{tab:model_evaluation_ext}. Since the size of the training data and the number of parameters in the model are related in terms of the bias-variance trade-off, each of the three FCNN models consumes a differing, dedicated size of training data for fair comparison.

We also compare the performance of the agents using two key driving metrics. The first metric is a macro-level statistic of the average distance travelled by the agents on the test tracks before they hit the road boundaries. The second, more micro-level metric is based on the \textit{smoothness} factor of the driving trajectory. This is shown through the jerk and entropy values associated with the predicted steering angle during the test runs. Jerk is a third-order derivative of the predicted steering angle. If we refer to the predicted steering angle as $\delta$, then the steering velocity is calculated as $\delta^\prime$, steering acceleration as $\delta^{\prime\prime}$ and finally the jerk value as $\delta^{\prime\prime\prime}$. 

To keep experimental consistency, the PENN architecture for the learning and physics blocks is kept the same as that in the PERNN model. In addition, the PINN-based model architecture is kept the same as the smallest FCNN model for a direct and fair comparison of the effectiveness of the method in comparison to our approach.

In summary, we investigate the effectiveness of the PERNN model against the FCNN and PINN counterparts based on the two areas:
\begin{enumerate}
    \item Data requirements include the number of unique tracks as part of the training data.
    \item The complexity of the architecture in terms of the number of parameters in the neural network to capture driving behaviour from the expert demonstrations.
\end{enumerate}

\subsubsection{Results}

Table \ref{tab:model_evaluation_ext} presents the results of this analysis. It can be seen that the PERNN consumes much less training data (only 6 out of 36 tracks used in training data) compared to the FCNN models in general and is composed of significantly fewer model parameters.  Despite this, the PERNN model converges better than the small and medium FCNN models on the test dataset on MAE (Mean Absolute Error). It is important to note that the PENN model comprising no residual blocks struggles to converge on the test dataset owing to restricted gradient flow. The agent based on this model also fails to generalize the learned driving behaviour in the live simulation with a significantly low average distance travelled on the test tracks.

\begin{figure*}[htbp]
  \centering
    \begin{subfigure}[b]{0.8\textwidth}  
    \centering
    \includegraphics[width=\textwidth]{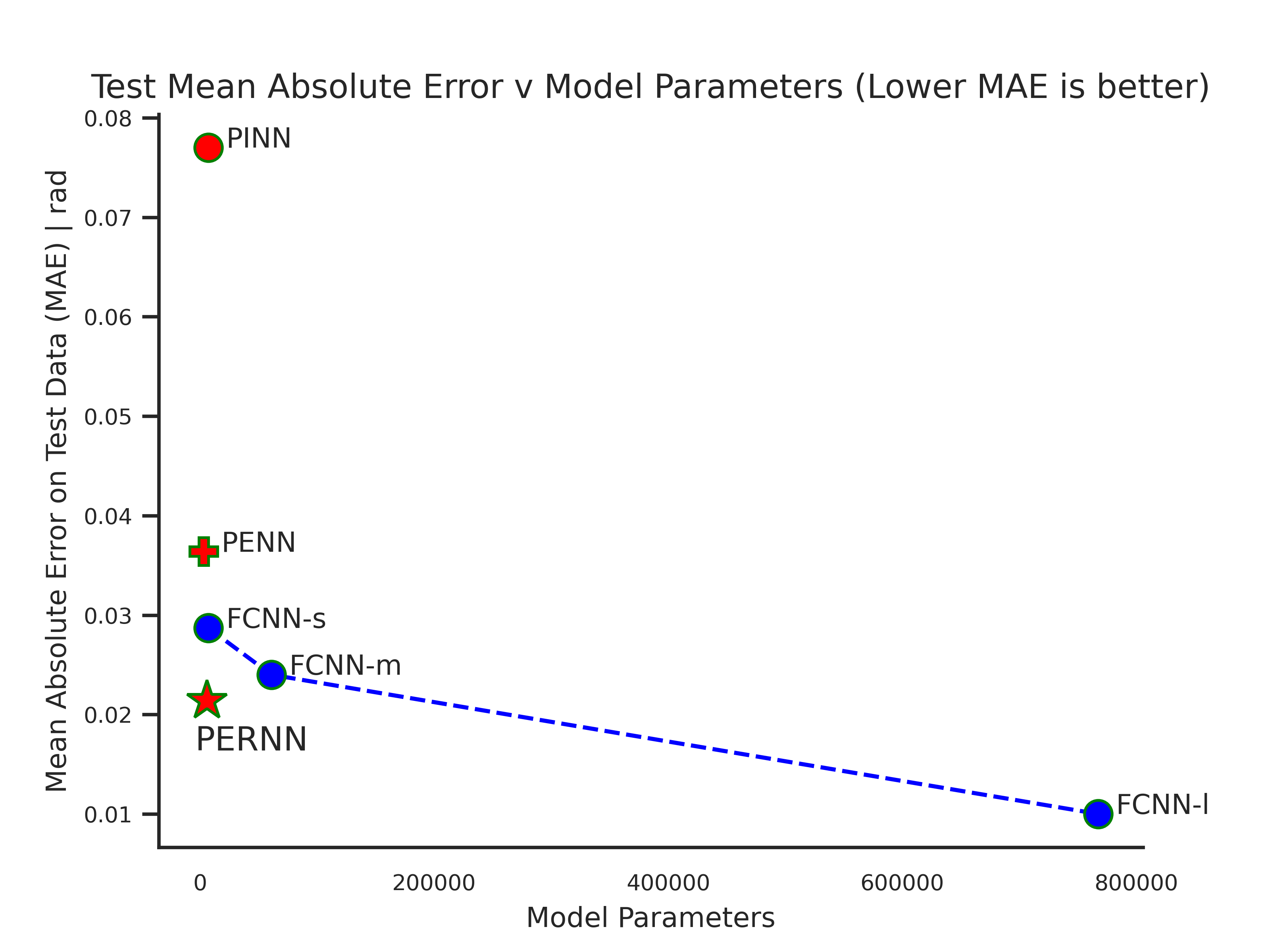}  
    \caption{Comparison of MAE score to model parameters for each model in the experimentation. Our PERNN model shows comparable MAE scores on the test dataset to the FCNN-large model but with approximately 260 times fewer model parameters. This is due to the physics blocks providing prior steering knowledge to the model as an inductive bias}
    \label{fig:mae_v_params}
  \end{subfigure}

  \begin{subfigure}[b]{0.8\textwidth}  
    \centering
    \includegraphics[width=\textwidth]{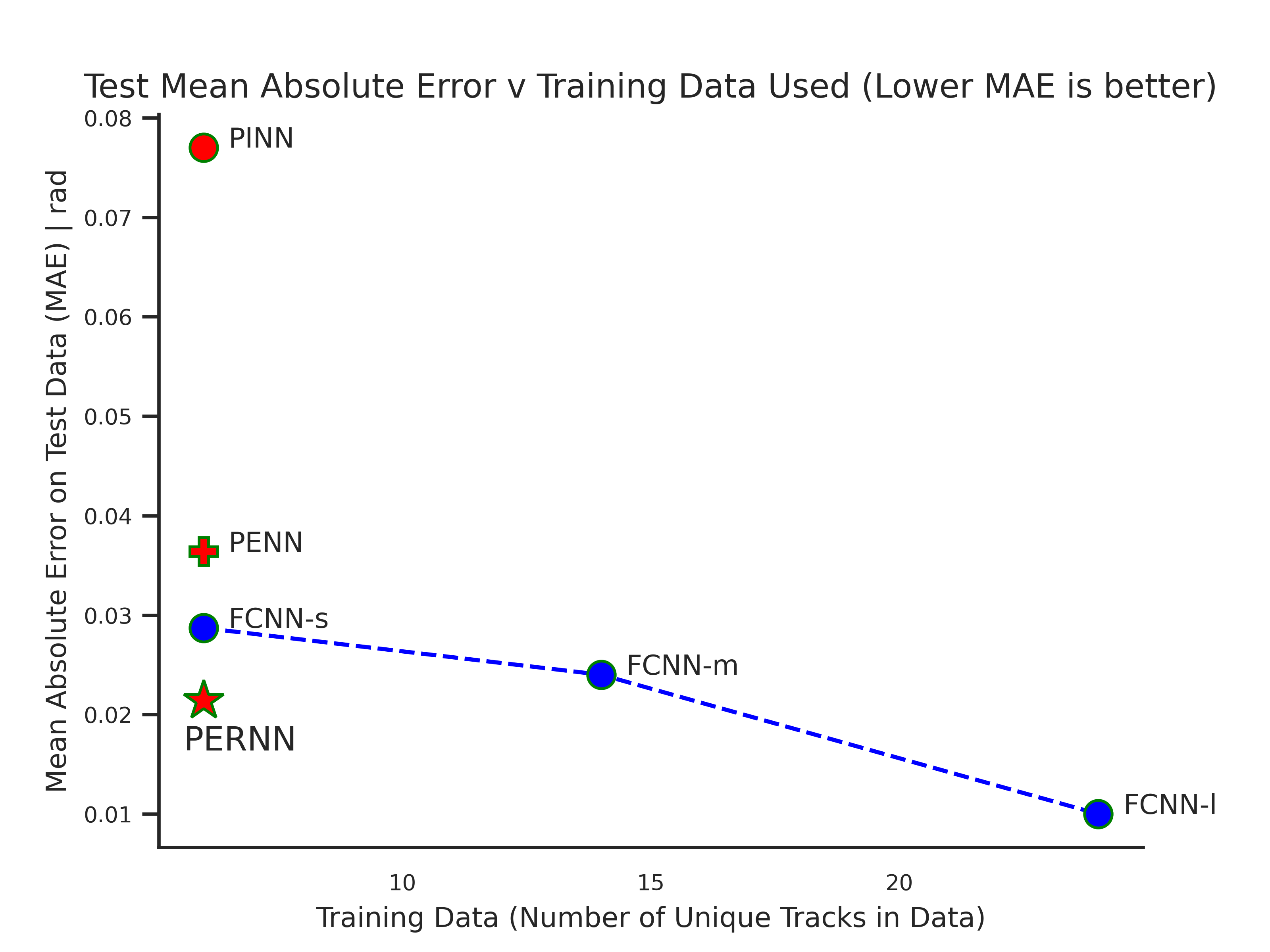}  
    \caption{Comparison of MAE score to training data requirements for each model in the experimentation. Our PERNN model shows comparable MAE scores on the test dataset to the FCNN-large model while using five times less training data in the form of unique tracks in the generated observation data. This is due to the physics blocks providing prior steering knowledge to the model as an inductive bias.}
    \label{fig:mae_v_tracks}
  \end{subfigure}
  
  \caption{Comparison of model accuracy (Mean Absolute Error), number model of parameters and training data requirements.}
  \label{fig:experiments_mae}
\end{figure*}

\begin{figure*}[htbp]
  \centering
    \begin{subfigure}[b]{0.8\textwidth}  
    \centering
    \includegraphics[width=\textwidth]{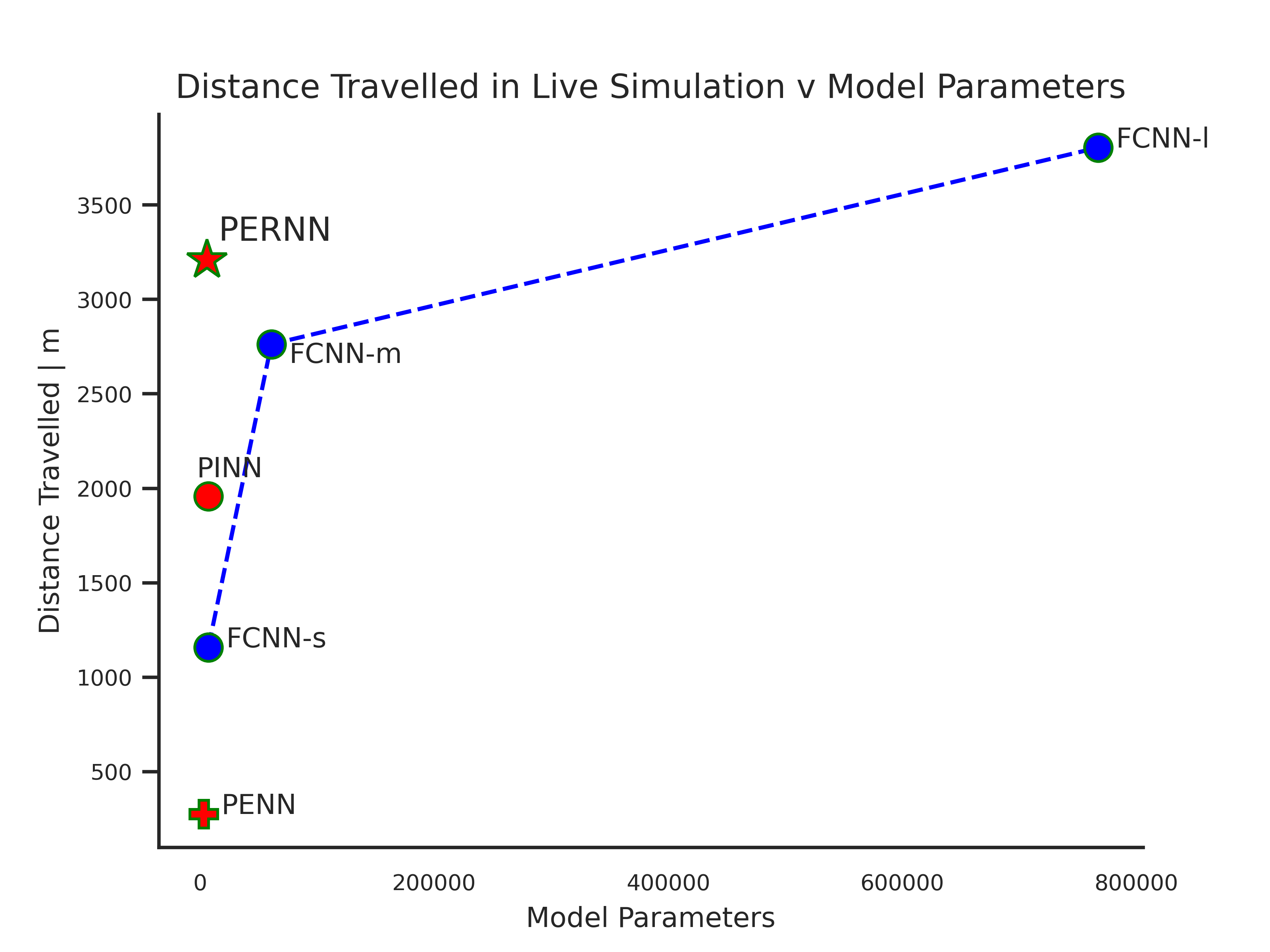}  
    \caption{Comparison of distance successfully manoeuvred by the agent (before the collision with road boundary) on test tracks against model parameters for each model in the experimentation. The PERNN model outperforms all other approaches except FCNN-large with lesser model complexity.}
    \label{fig:dists_v_params}
  \end{subfigure}

  \begin{subfigure}[b]{0.8\textwidth}  
    \centering
    \includegraphics[width=\textwidth]{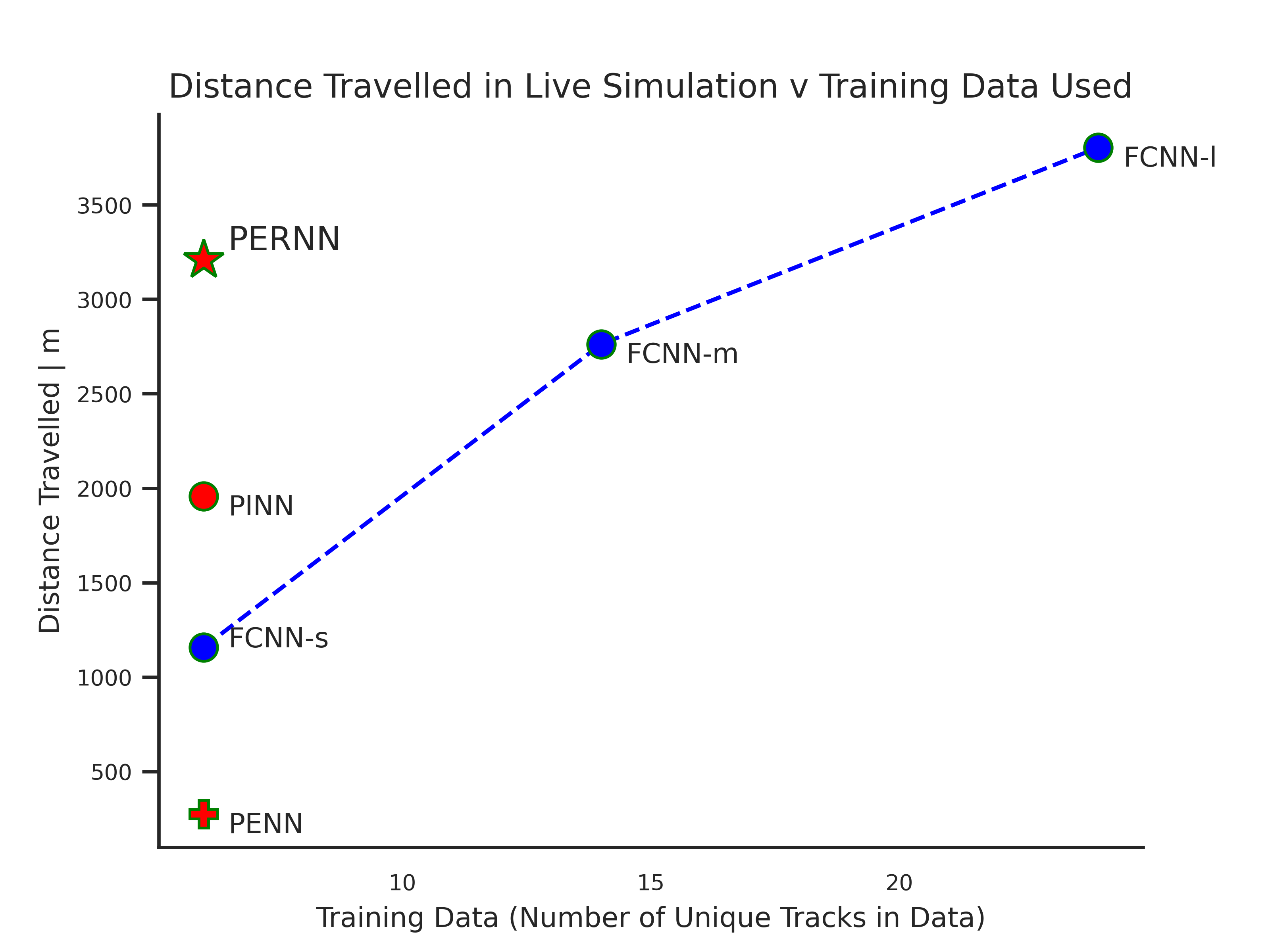}  
    \caption{Comparison of distance successfully manoeuvred by the agent (before the collision with road boundary) on test tracks against unique tracks used in training data for each model in the experimentation. The PERNN model outperforms all other approaches except FCNN-large with lesser training data.}
    \label{fig:dists_v_tracks}
  \end{subfigure}
  
  \caption{Comparison of distance travelled by agent in simulation on test tracks, number model of parameters and training data requirements.}
  \label{fig:experiments_dists}
\end{figure*}

Figure \ref{fig:experiments_mae} elaborates on the results. Figure \ref{fig:mae_v_params} compares the MAE scores to the number of parameters for each model in the experimentation. The PERNN model shows comparable MAE scores on the test dataset to the FCNN-large model but with approximately 260 times fewer model parameters. Similarly, Figure \ref{fig:mae_v_tracks} compares MAE scores to training data requirements for each model in the experimentation. Our PERNN model shows comparable MAE scores on the test dataset to the FCNN-large model while using five times less training data in the form of unique tracks in the generated observation data. Both of these observations are a testament to the effectiveness of the physics blocks in the PERNN architecture in providing prior steering knowledge to the model as an inductive bias. Similarly, Figure \ref{fig:experiments_dists} compares the average distances travelled without collision during the live simulation on test tracks by each agent. PERNN-based agent achieves higher distances compared to FCNN-small, FCNN-medium models and the PINN model while using less training data and comprising fewer parameters.

\begin{figure}[htbp]
    \centering
    \includegraphics[width=1\textwidth]{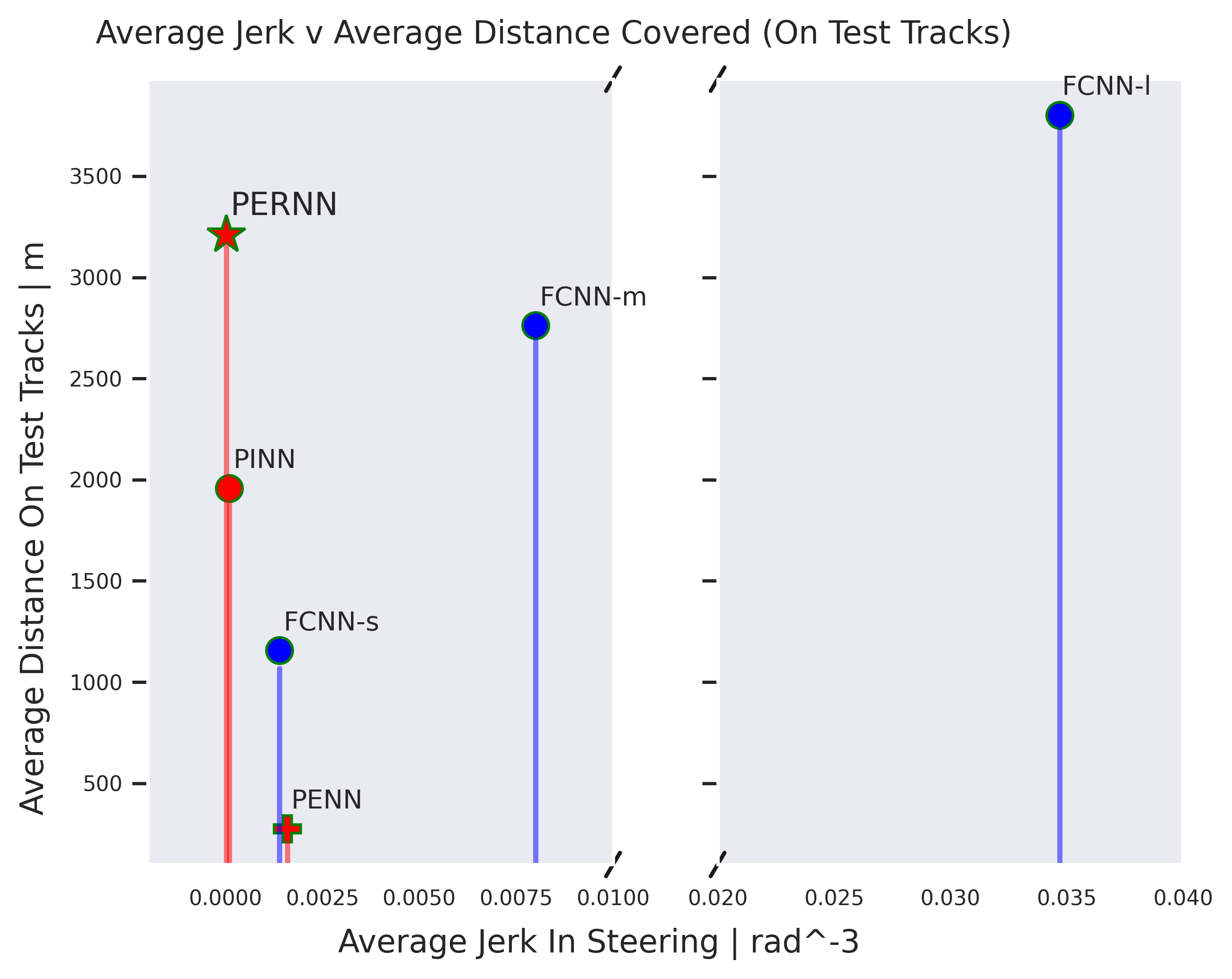}
    \caption{
Comparison of the distance successfully manoeuvred by the agent (before the collision with road boundary) on test tracks against the jerk produced by the steering (lower the jerk, smoother the driving trajectory). The experiment is carried out for each model. The PERNN model achieves a comparable successful distance travelled to the FCNN-large model but with significantly smaller jerk values, indicating a smoother driving trajectory.}
    \label{fig:dist_v_jerk}
\end{figure}

Figure \ref{fig:dist_v_jerk} shows the comparison of the distance successfully manoeuvred by the agent (before the collision with the road boundary) against the jerk produced by the steering (lower the jerk, smoother the driving trajectory). The analysis is carried out for the driving agent based on each model in the experimentation. The PERNN model outperforms FCNN-small and FCNN-medium models regarding both higher average distance travelled and significantly less jerk in predicted steering. The large version of FCNN exhibits better MAE values on the test set compared to the PERNN model and eventually surpasses it in navigating longer patches of tracks during the live simulation. It however suffers from higher values of steering jerk due to the model attempting to recover from \textit{cascading errors}, a known issue in these methods under Behavioral Cloning settings. Since the PERNN model inherently encodes the basic error recovery function of following the reference point from any arbitrary state, it offers smaller jerk values and a smoother driving trajectory without explicit variations in training data.

\begin{figure}[htbp]
    \centering
    \includegraphics[width=1\textwidth]{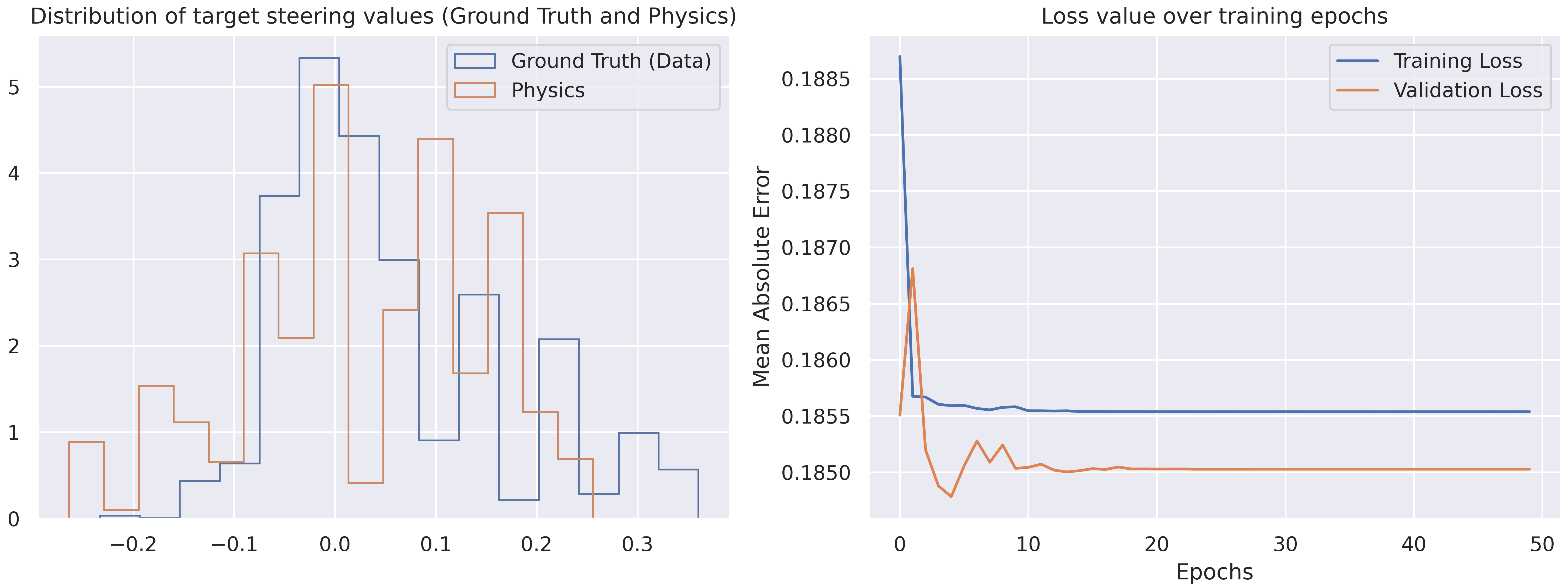}
    \caption{
Distribution of target steering values from the observation (ground-truth) data and the physics model in the PINN approach (shown to the left). The decoupled distributions lead to the loss function struggling to converge on the training and validation dataset (shown to the right).}
    \label{fig:pinn_issues}
\end{figure}

It is important to understand the inherent limitation of the PINN approach which is overcome by our PERNN method, allowing better accuracy. Figure \ref{fig:pinn_issues} illustrates the difference in the distribution of the ground-truth steering values and the steering values generated from the physics model (Pure Pursuit) for empty road scenarios. This forms decoupled targets for the model to learn, which is depicted in the graph to the right of the distribution showing the validation dataset loss values failing to converge anywhere close to the other methods. Despite this, it outperforms FCNN-small (with equal size of training data and number of model parameters) on live simulation on the test tracks in terms of both average distance travelled and average jerk values. This illustrates the benefit in terms of incorporating prior physics knowledge of driving in the model. PERNN takes a step ahead and provides the capability to learn from both the data and the physics using the concept knowledge blocks, facilitated by residual blocks and therefore does not suffer from the decoupled behaviour in physics and data.
\begin{table*}[!btp]
  \caption{Numerical results of the experiments comparing the performance of PERNN against FCNN-based agents}
  \label{tab:model_evaluation_ext}
  \centering
  
  \begin{tabular}{p{1cm}p{2cm}p{2cm}p{2cm}p{2cm}p{1.5cm}p{1cm}}
    \toprule
    \textbf{Model} & \textbf{Tracks in Train Set}&\textbf{Neural Configuration} & \textbf{Parameters} & \textbf{Test MAE} ($rad$)& \textbf{Avg. Distance}($m$) & \textbf{Avg. Jerk} ($rads^{-3}$) \\
    \midrule
    FCNN-small& 6& [64, 32, 16, 8, 4, 1] & 6977& 0.0287& 1157& 0.0014\\
    FCNN-medium& 14 & [256, 128, 64, 32, 16, 8, 4, 1] & 60801& 0.024& 2762& 0.008\\
    FCNN-large& 25 & [1024, 512, 256, 128, 64, 32, 16, 8, 4, 1] & 767617& \textbf{0.01}& \textbf{3803}& 0.0347\\
 PINN& 6& [64, 32, 16, 8, 4, 1] & 6977& 0.077& 1958&0.0001\\
    PERNN& \textbf{6}& Learning Block: [32, 16, 8, 2] + Residual Block: [8, 4, 1] + PhysBlock& 5611& 0.0214& \textbf{3209}& \textbf{0.00003}\\
 PENN& 6& Learning Block: [32, 16, 8, 2] + PhysBlock& 2794& 0.0364& 275&0.0016\\
  \end{tabular}
\end{table*}


\subsubsection{Analysis of Unobserved Intermediate Variables}

Another key benefit of the PERNN model is that it predicts the intermediates variables, the lookahead distance and heading difference values which are not explicitly part of the observation data from the driving environment. We analyse the distribution of these variables for a set of road scenarios to analyse the agent's decision-making based on the PERNN model predictions. This is specifically defined using the range finder sensors $R$ and $O$ (see Table \ref{tab:variables_parameters}), for filtering examples based on the relative distances from the track edge and traffic vehicles on either side of the vehicle in the driving space ahead. Index value $9$ in both $R$ and $O$ suggests distances to objects in the direction of the vehicle's heading (straight ahead).

\begin{figure}[H]
    \centering
    \includegraphics[width=0.8\linewidth]{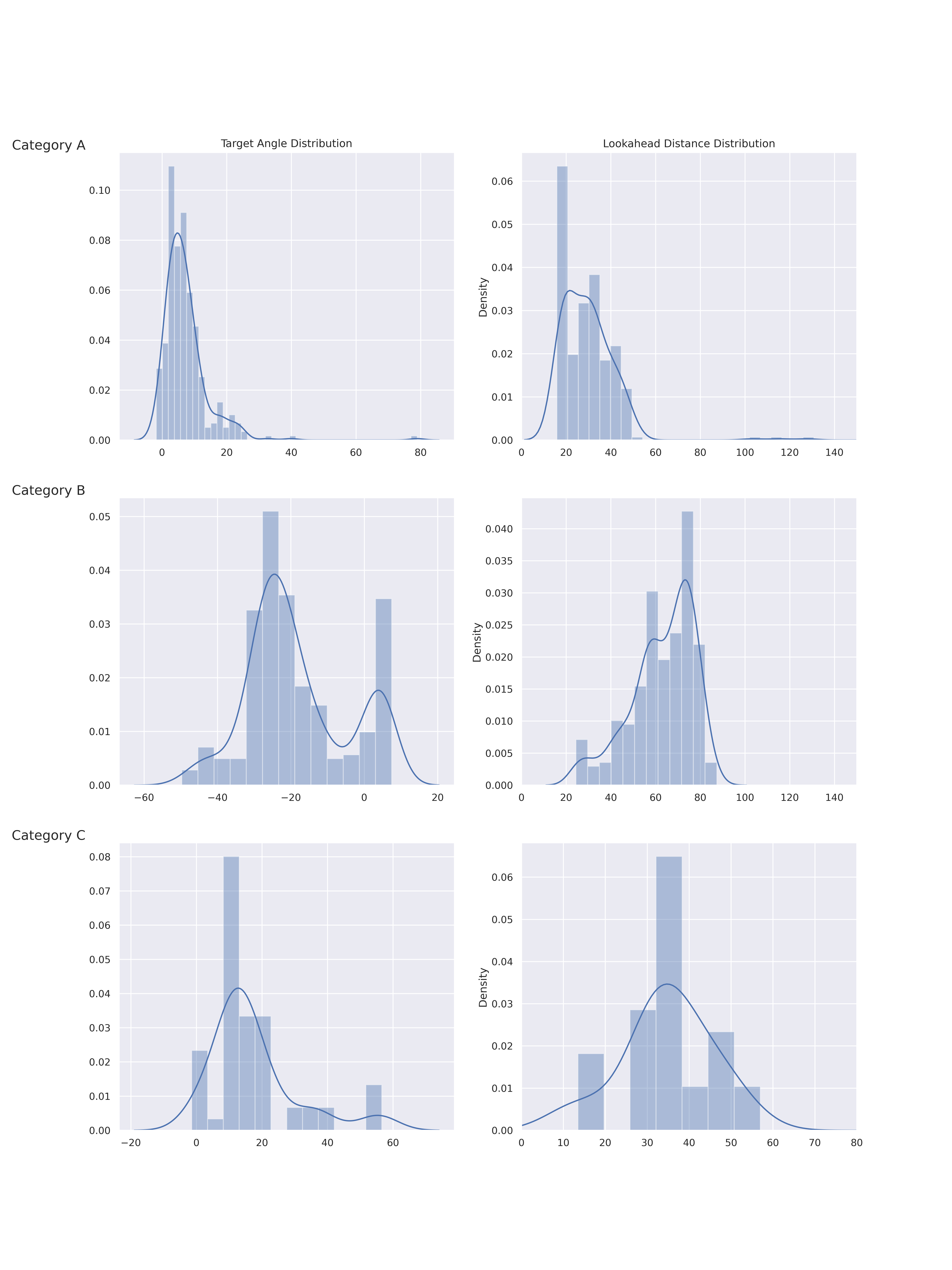}
    \caption{An analysis of the model rationale in predicting the intermediate variables in the environment. The plot shows distributions of the intermediate variables, target angle (heading difference) and lookahead distance to the selected point on the road by the PERNN-based agent. Each category is represented as a row of two distribution plots.}
    \label{fig:discovery}
\end{figure}

We divide the road scenarios in the test data into three simplified categories:
\begin{itemize}
    \item Category A: Scenarios with relatively more space towards the right of the driving space ahead. The data points in the set follow the principle that for distances defined in vectors $R$ and $O$:

\[R_{10} > R_8 + 20 \]
\[O_{10} > O_8 + 20 \]
This implies that the track boundary or vehicle at 10 degrees to the left of the agent vehicle is at least 20m further compared to the track boundary or vehicle at 10 degrees to the right of the vehicle's direction. 

    \item Category B: Scenarios with relatively more space towards the left of the driving space ahead. The data points in the set follow the principle that for distances defined in vectors $R$ and $O$:

\[R_{8} > R_{10} + 20 \]
\[O_{8} > O_{10} + 20 \]
This implies that the track boundary or vehicle at 10 degrees to the right of the agent vehicle is at least 20m further compared to the track boundary or vehicle at 10 degrees to the left of the vehicle's direction. 
    \item Category C, where there is limited space on either side, and a traffic vehicle is situated at the center in front. This makes scenarios where overtaking is improbable. The data points in the set follow the principle that for distances defined in vectors $R$ and $O$:

\[R_8 < 40 \quad \text{and} \quad R_{10} < 40\]
\[O_8 < 40 \quad \text{and} \quad O_{10} < 40\]
\[O_9 < 70\]
This implies that the track boundary or vehicle at 10 degrees to the right and left of the agent vehicle's direction of motion is less than $40m$ and that there is a vehicle in front at a distance of less than $70m$.
\end{itemize}
The values for the distance bounds are empirically found to depict the behaviour most vividly and can be set as any arbitrary value based on the concept of the three category divisions. Figure \ref{fig:discovery} illustrates the results of this analysis. The distribution of the predicted heading angle for the Category A dataset clearly shows a right skew, suggesting that the model mostly selected points that were to the right side of the driving space in the front. Similarly, for Category B, the left skew is vividly depicted, which is understandable based on the specifications of the scenarios. The heading difference in Category C is slightly more evenly distributed despite a right skew. It could be inferred that the reference agent preferred to stay more towards the right in this scenario, which behaviour the PERNN model captured. It can be seen that the lookahead distance distribution is generally less uniform for Category A and B compared to Category C. This is understandable since Category C comprises more restricted scenarios based on available driving space for the agent. 

The encouraging results of this analysis suggest the PERNN model offers significantly more explainability compared to conventional neural networks through the discovery of unknown, human-understandable variables in the environment.

\subsection{Net Ecosystem Exchange Modeling \label{sec:nee_experiments}}

In this section, we present details on the experimentation and results on the problem
of NEE modeling for gap-filling in flux-tower observational data. We first present a short analysis on the difference in NEE behavior reflected in the simplistic physics model of NEE based on Equation \ref{eq:neenighttimemodel_rephrased} and the flux-tower measurements. In addition, similar to section \ref{sec:driving_results}, we present and compare results for PERNN with PENN (Physics Encoded Neural Network) as a variant of our method without the residual block and PINN (Physics Informed Neural Network) that uses the NEE ODE as a regularization on the loss term only, and finally with a conventioanl FCNN (Fully Connected Neural Network). We also present comparison against Random Forest which serves as the current state-of-the-art method for NEE gap-filling \cite{zhuclementetal2022} and XgBoost, another decision-tree based algorithm similar to Random Forest for extended comparison.

\subsubsection{Analysis of NEE Physics Model}

\begin{figure}[htbp]
    \centering
    \begin{subfigure}[b]{0.48\linewidth}
        \includegraphics[width=\linewidth]{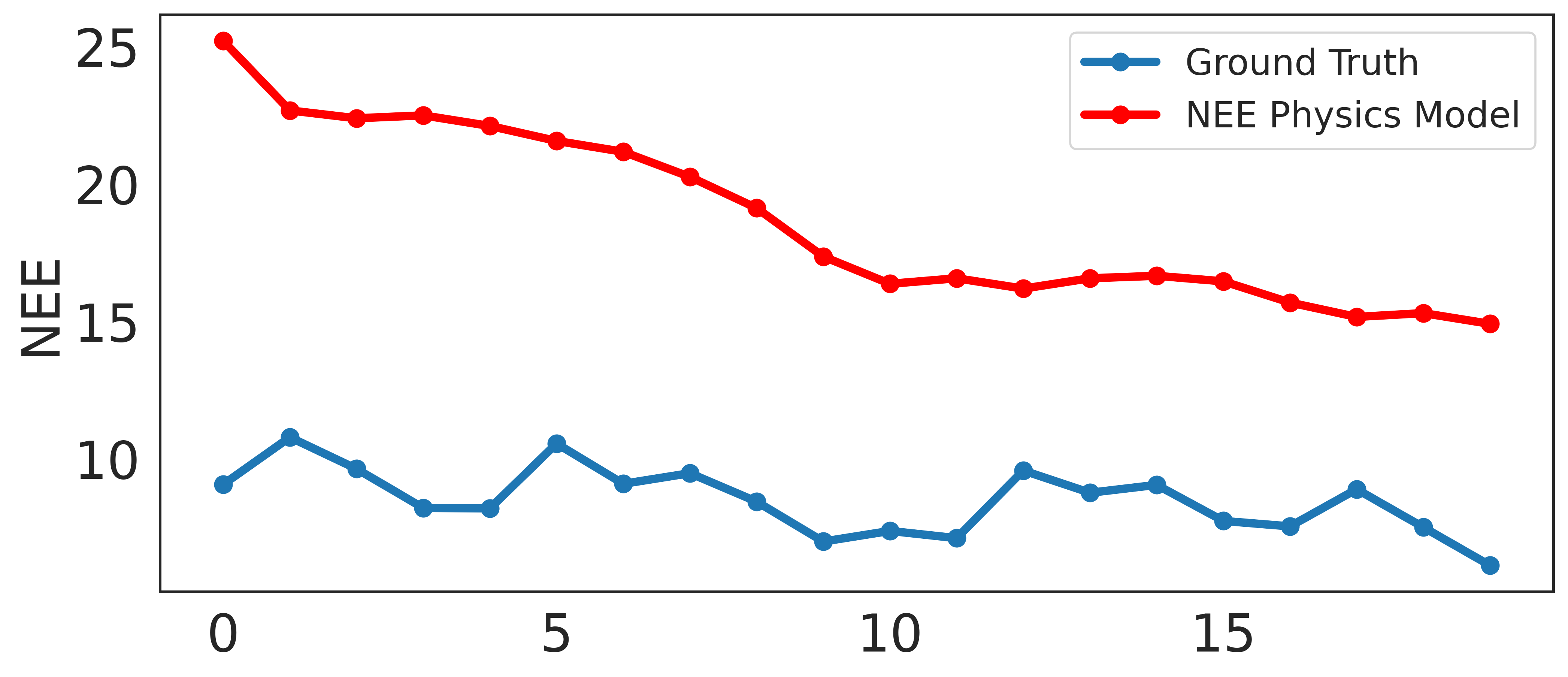}
    \end{subfigure}\hfill
    \begin{subfigure}[b]{0.48\linewidth}
        \includegraphics[width=\linewidth]{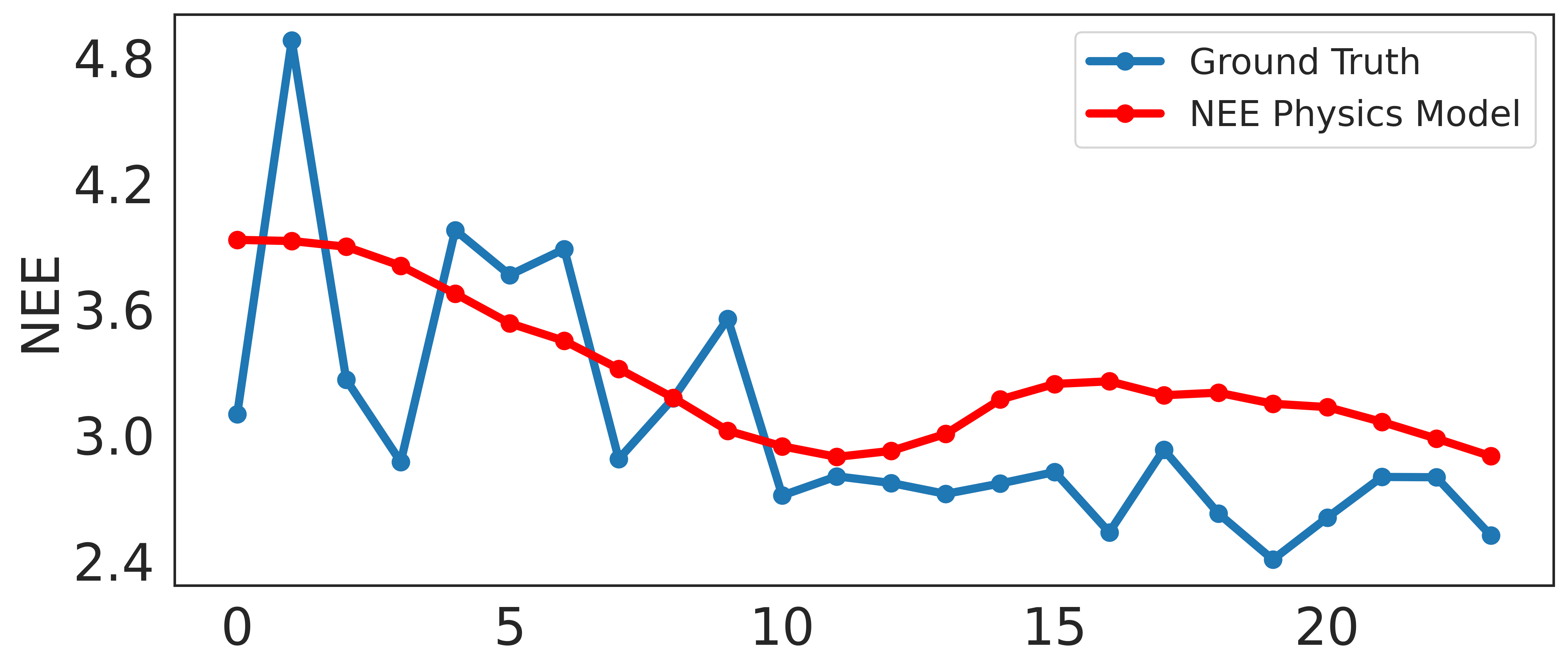}
    \end{subfigure}
    
    \vspace{2mm} 
    
    \begin{subfigure}[b]{0.48\linewidth}
        \includegraphics[width=\linewidth]{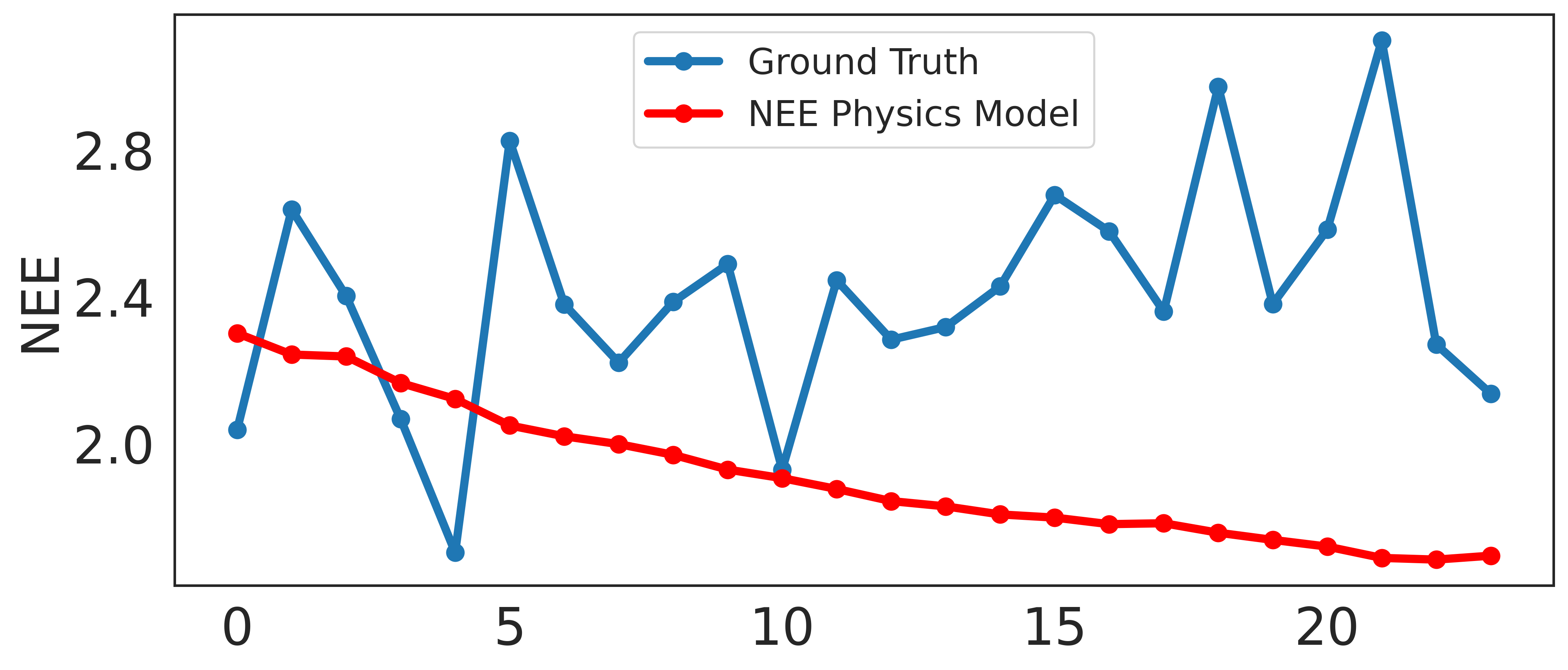}
    \end{subfigure}\hfill
    \begin{subfigure}[b]{0.48\linewidth}
        \includegraphics[width=\linewidth]{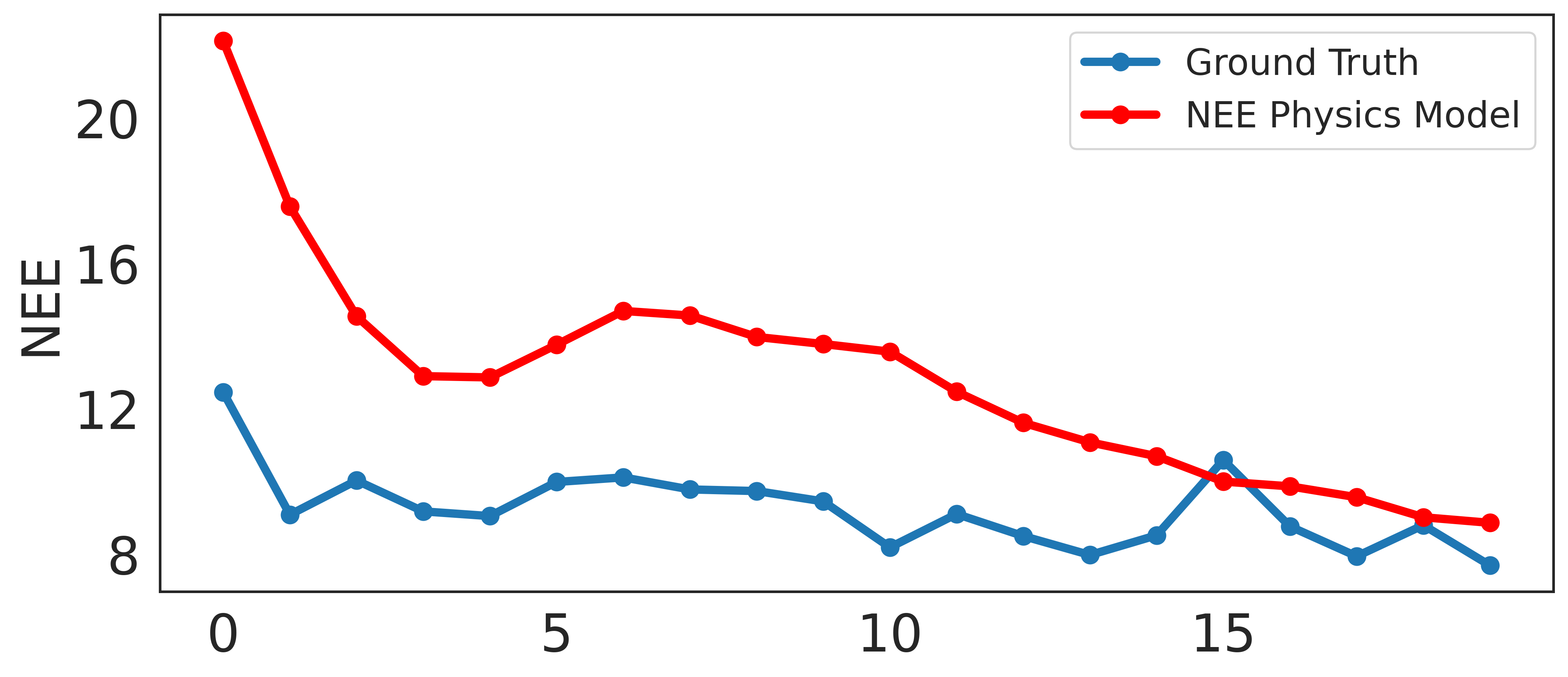}
    \end{subfigure}
    
    \caption{The comparison of daily NEE values from flux-data and the output from NEE physics model based on Equation \ref{eq:neenighttimemodel_rephrased} for four randomly sampled daily cases from the test-set. The blue line represents the ground-truth observations and the red line represents the values from the NEE physics model. The top row shows samples from dates 2018-08-18 and 2019-10-24, while the bottom row shows samples from dates 2019-03-04 and 2019-08-21 (left to right).}
    \label{fig:gt_v_phy}
\end{figure}

Since the NEE ODE (based Equation \ref{eq:nee_final_ode}) incorporated in our Physics Block is based on the equation modeled using an Arrhenius-type expression from Equation \ref{eq:neenighttimemodel_rephrased}, we compare the NEE dynamics based on this equation with the ground-truth NEE values from the flux-tower observational data. Figure \ref{fig:gt_v_phy} presents four plots from randomly sampled daily cases of NEE observational data. The blue lines represent these observed values, while the red lines represent the values from the NEE physics model calculated using corresponding observed parameter, $T_{air}$ and estimated parameters $E_0$ and $r_{b,night}$. It can be observed that the underlying NEE physics model (and consequently the NEE ODE) is an over-simplification of the actual NEE trends in the observational data, albeit following the general progression most of the time.

\subsubsection{Experimental Setup}

We split the flux dataset into training and test sets based on the year of recording. The training set comprised years 2012-2017 (inclusive) while the test set was reserved for years 2018 and 2019 for experimental consistency when measuring gap-filling accuracy over different time scales: daily, weekly, monthly, and quarterly.

\begin{figure}[htbp]
    \centering
    \begin{subfigure}[b]{\linewidth}
        \includegraphics[width=\linewidth]{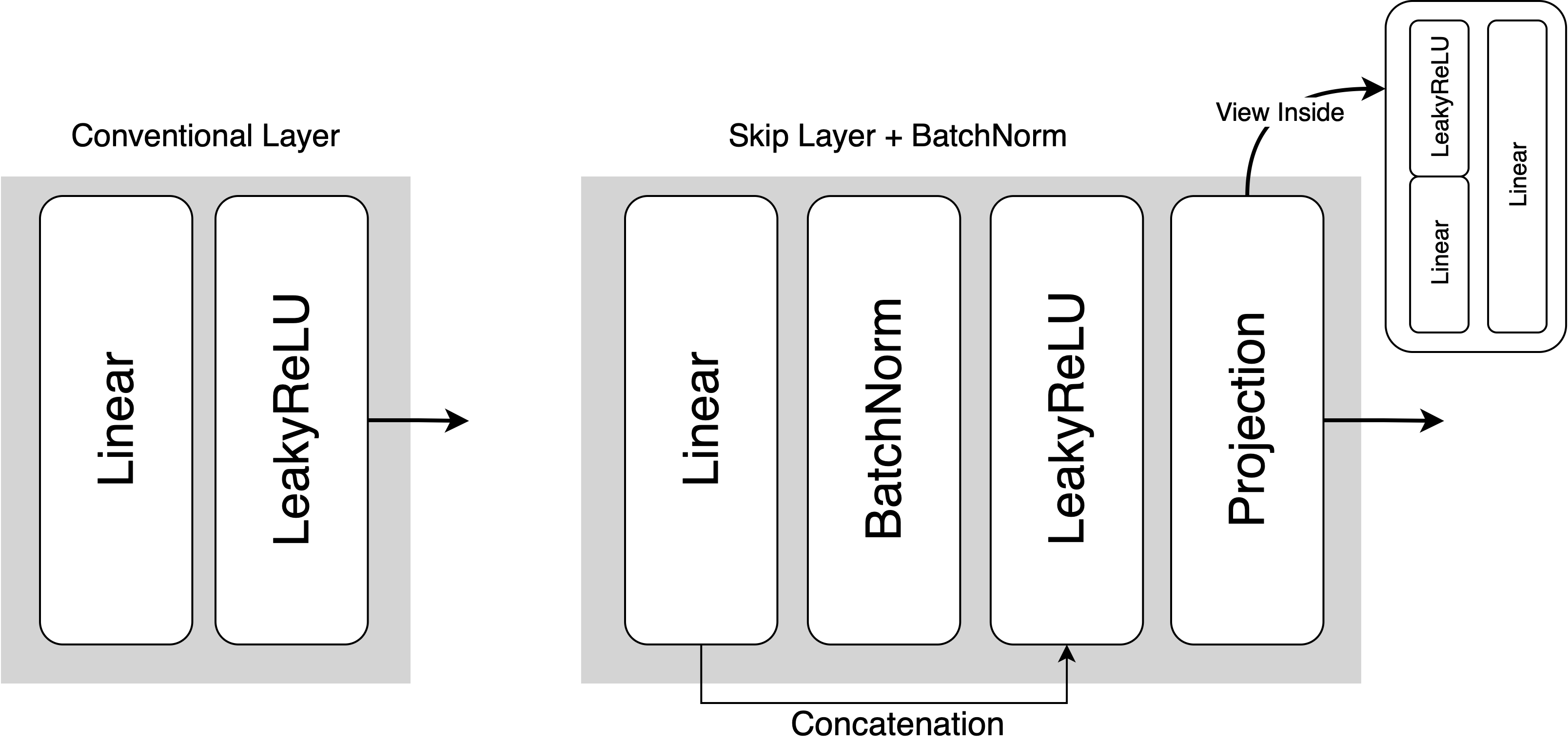}
    \end{subfigure}
    \caption{A comparison of conventional and skip connection with batch normalization layers}
    \label{fig:layer_types}
\end{figure}

For all neural network based methods in this study, we keep the core network architecture and hyperparameters same for experimental consistency. The models based on PERNN, PENN, PINN and FCNN all have a set of layers comprising a skip connections and batch normalization and a LeakyReLU activation function. Figure \ref{fig:layer_types} illustrates the difference between conventional fully connected linear layers with activation function, and our skip-connection based layers. These layers involve the concatenation of earlier linear layers with the output of the activation layer (LeakyReLU) followed by a linear projection. We observed that these skip connection based layers allowed better convergence of the model, which can be seen by the results in Table \ref{tab:nee_results} and explained in detail in section \ref{sec:nee_results}. To show the impact of these skip-connection based layers, we also include experiments on the versions of PERNN and PENN with conventional layers, labelled PERNN-NoSkip and PENN-NoSkip respectively.

\begin{figure}[htbp]
    \centering
    \begin{subfigure}[b]{\linewidth}
        \includegraphics[width=\linewidth]{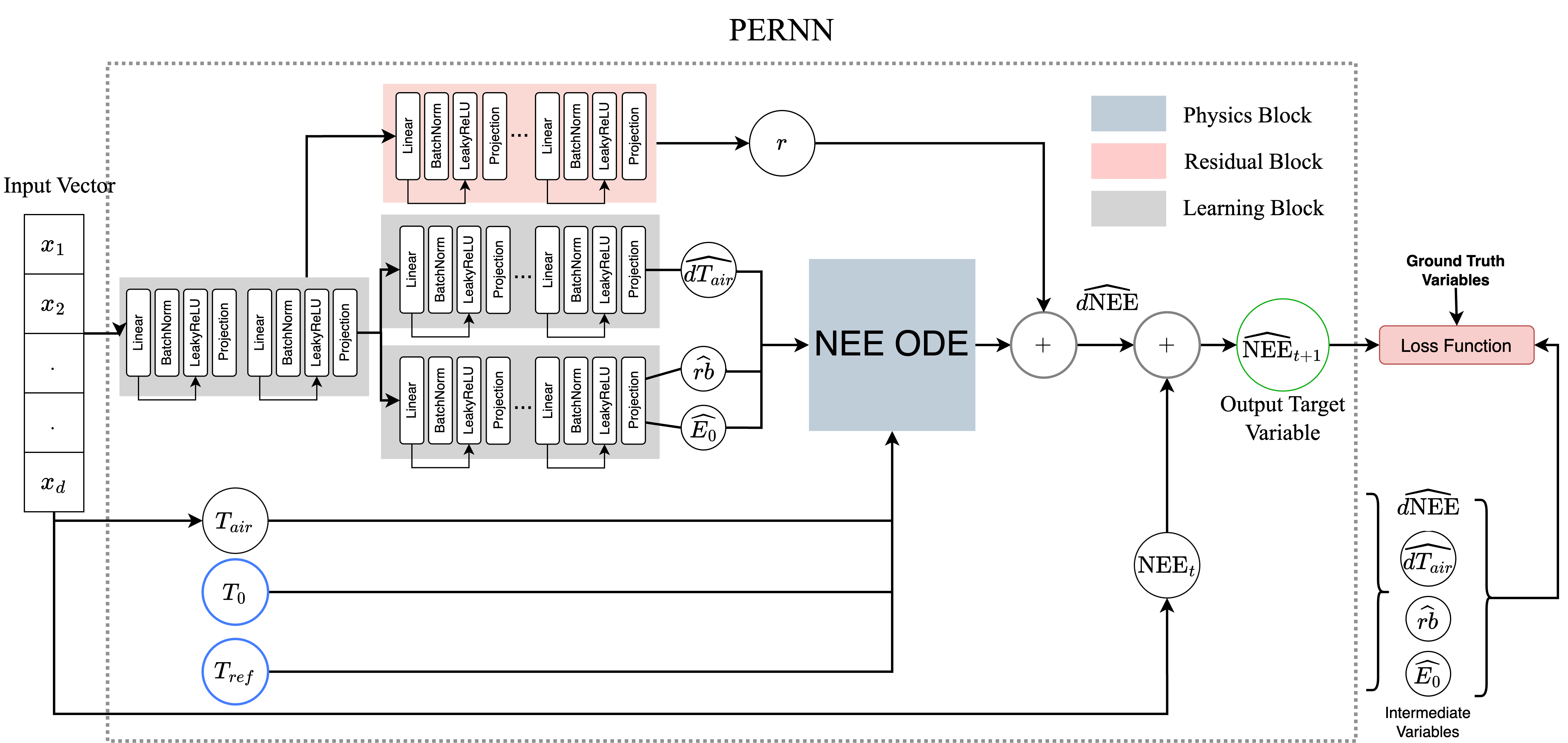}
    \end{subfigure}
    \vfill 
    \begin{subfigure}[b]{\linewidth}
    \includegraphics[width=\linewidth]{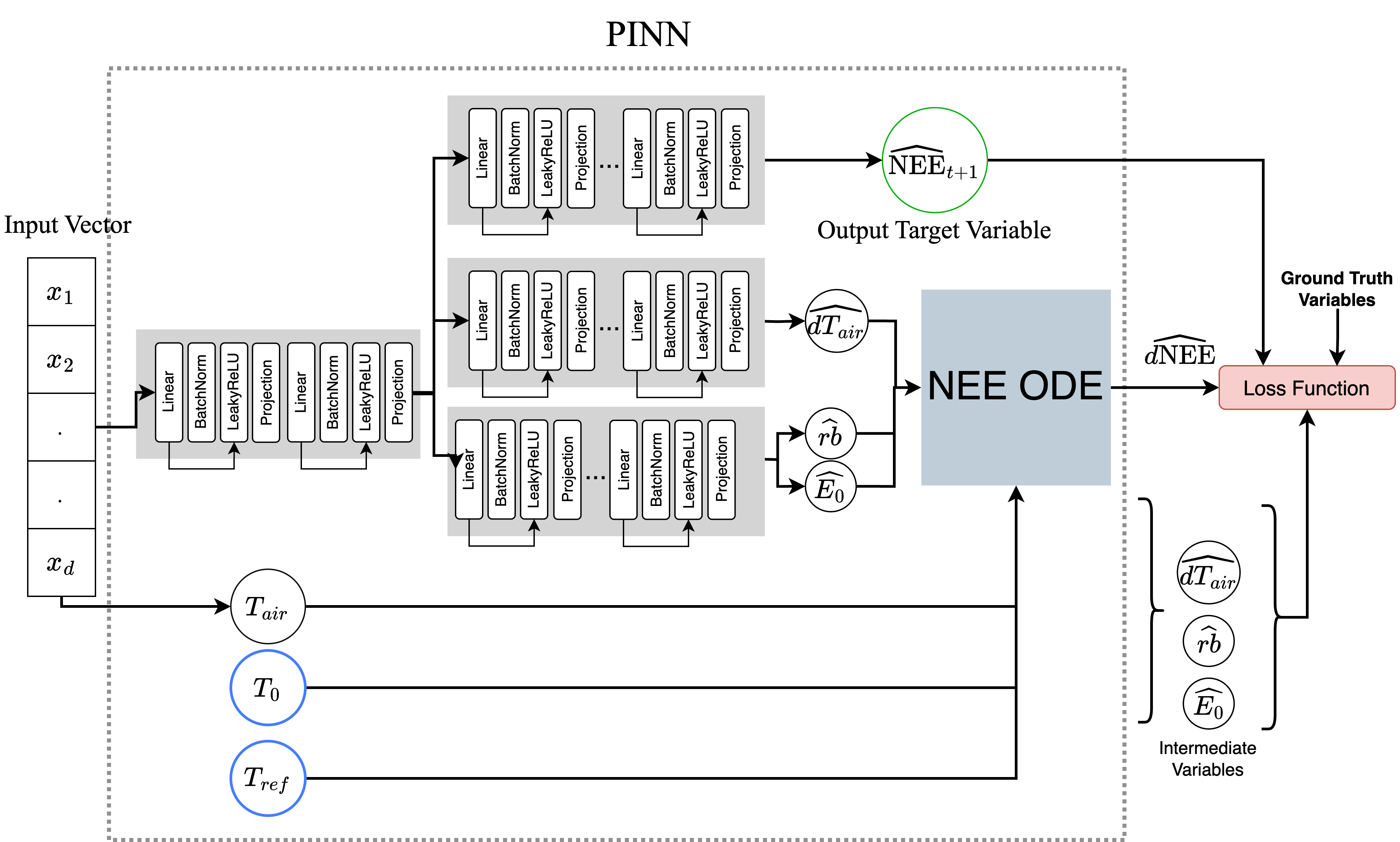}
    \end{subfigure}
    \caption{The architectures for the PERNN and PINN models. Top: PERNN architecture with two learning blocks, a residual block and physics block comprising NEE ODE as a direct part of computation graph. Bottom: PINN architecture with two branches outputting NEE ODE parameters which is used to regularize the loss function during training. A separate branch outputs the actual target NEE value.}
    \label{fig:pernn_v_pinn}
\end{figure}

In addition, Figure \ref{fig:pernn_v_pinn} shows the architectures of both PERNN and PINN models to illustrate the difference between the two approaches. In the PERNN model, the NEE ODE is a direct part of the computation graph inside the Physics Block, which outputs the final target variable ($\text{NEE}_{t+1}$). On the other hand, in the PINN model, the learnable layers directly predict the target NEE value ($\text{NEE}_{t+1}$) while the NEE ODE only offers external regularization to the loss function during training. In the PINN model, we present a branched architecture following similar structure to the learning blocks in PERNN. As seen in Figure \ref{fig:pernn_v_pinn}, two of these branches outputs the parameters for the NEE ODE, while a dedicated branch outputs the target NEE value. During testing in inference mode, only this dedicated branch is active and predicts the target NEE value. This experiment allows us to present the impact of our knowledge blocks in the neural architecture with the Physics Block explicitly housing the NEE ODE as a direct part of the computational graph.

\FloatBarrier

\subsubsection{Results \label{sec:nee_results}}

We evaluated the methods using three distribution-based metrics: Mean Maximum Discrepancy (MMD), Wasserstein Distance (Wsstn), and Kullback-Leibler Divergence (KL), to assess how well each technique captures the distribution of the target NEE. Additionally, we measured performance using mean absolute error (MAE) and R2 score to evaluate the fit to target variable. We also evaluate the accuracy of the intermediate variables from the learning blocks $E_0$, $r_{b,night}$ and $dT_{air}$, and the predicted time derivative of NEE, $d\text{NEE}$.

Table \ref{tab:nee_results} presents the results on the test set over these metrics for the different methods compared in this study. It can be seen that PERNN surpasses the state-of-the-art Random Forest method by a notable 38.8\% improvement in Wassertein Distance, 11\% improvement in KL Divergence score while also achieving lower MAE score. PENN also shows comparable performance to the Random Forest albeit falling behind PERNN in all metrics, clearly indicating the advantage in convergence led by the incorporation of residual blocks in the physics encoded neural architecture. It is interesting to note that PINN performs considerably lower on all metrics compared to PERNN and PENN, which could potentially indicate the advantage of a direct integration of the NEE ODE in the computational graph of the neural network architecture on emprical grounds. Both PERNN and PENN also appear to outperform the FCNN and XGB methods on all metrics. It can also be seen that the PERNN-NoSkip and PENN-NoSkip perform considerably worse than their counterparts, which validates the benefit of the skip-connection based layers introduced in the learning and residual blocks of both PERNN and PENN. 

\begin{figure*}[h!]
    \centering
    \begin{subfigure}[b]{0.24\textwidth}
        \includegraphics[width=\linewidth]{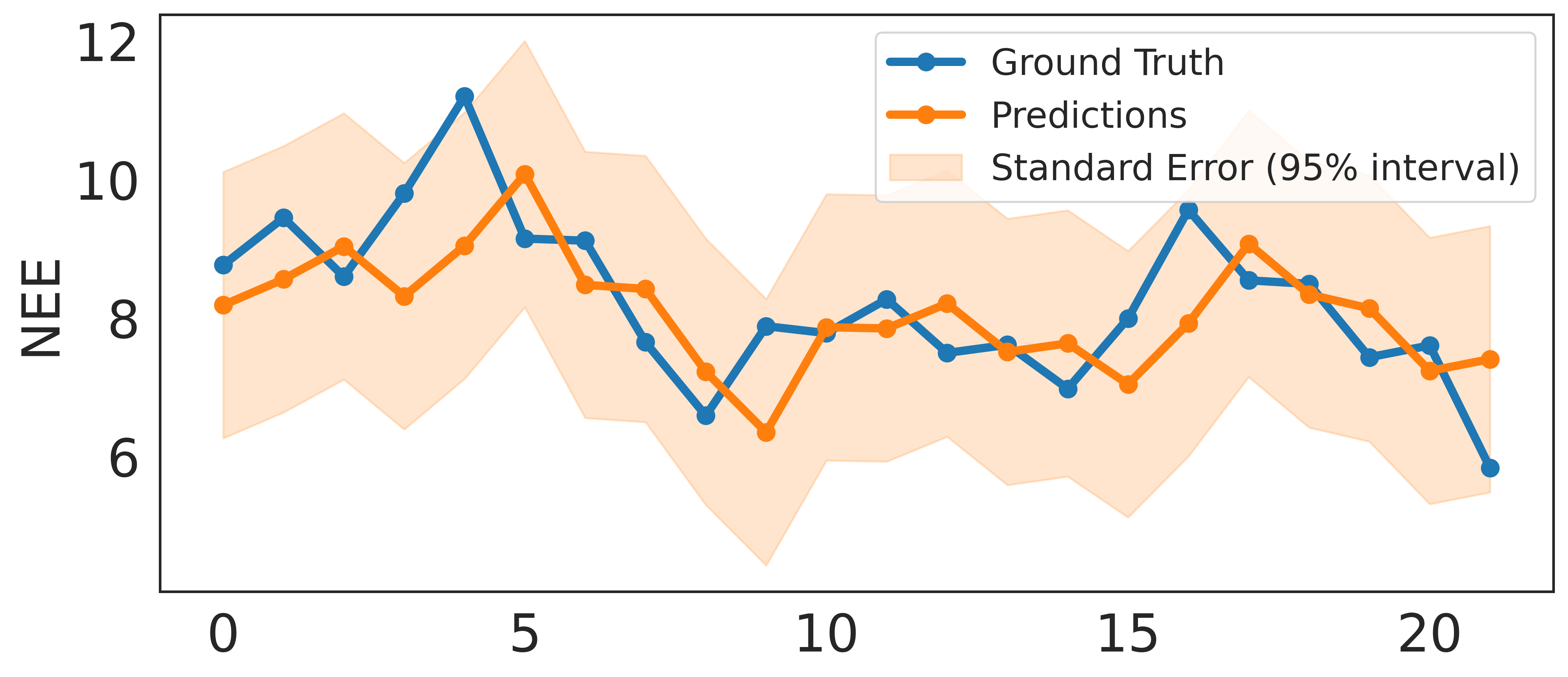}
    \end{subfigure}
    \begin{subfigure}[b]{0.24\textwidth}
        \includegraphics[width=\linewidth]{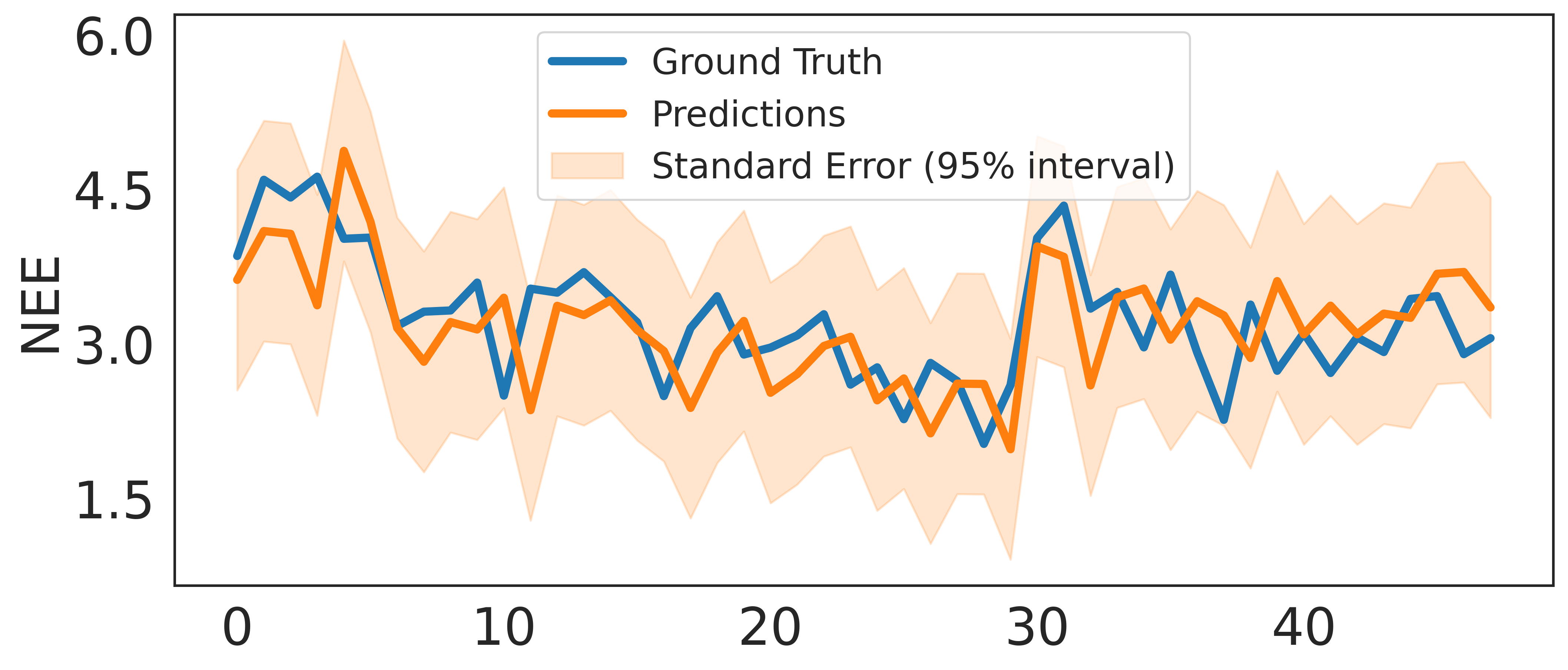}
    \end{subfigure}
    \begin{subfigure}[b]{0.24\textwidth}
        \includegraphics[width=\linewidth]{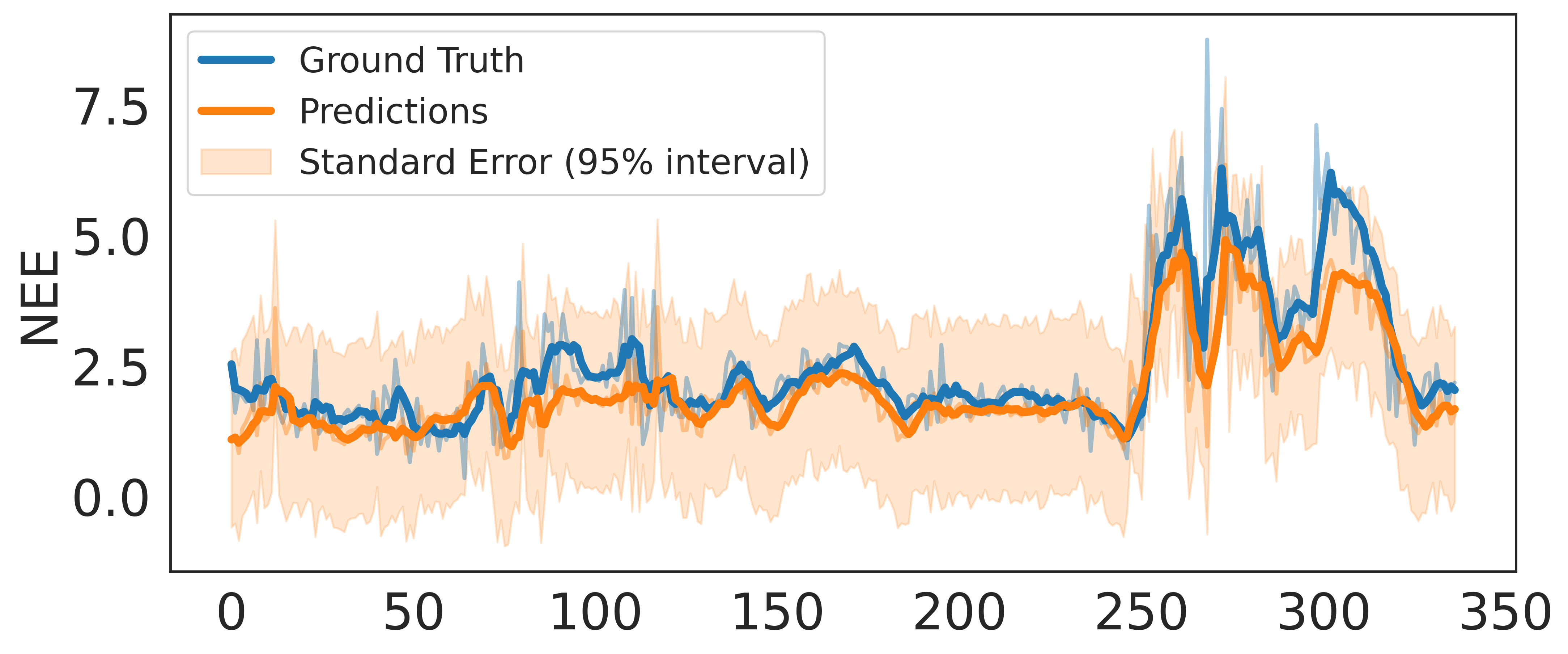}
    \end{subfigure}
    \begin{subfigure}[b]{0.24\textwidth}
        \includegraphics[width=\linewidth]{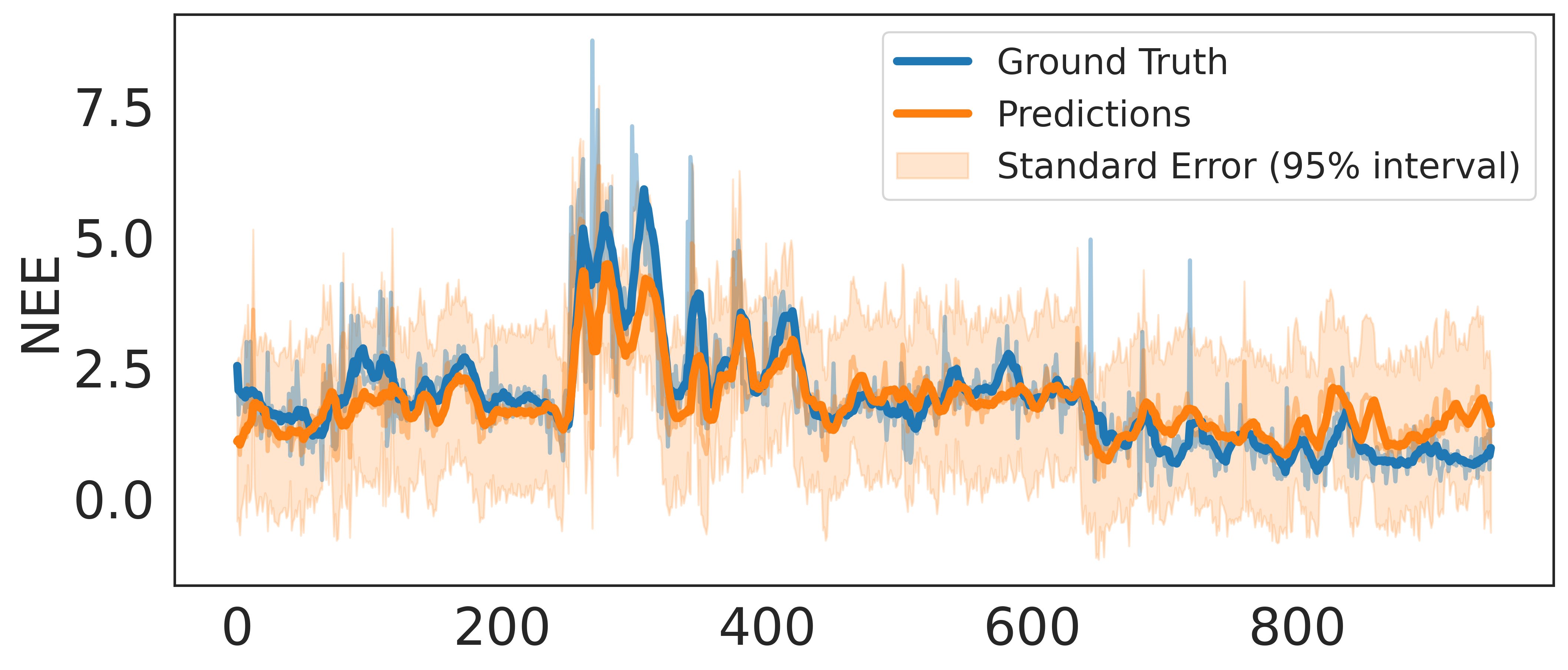}
    \end{subfigure}

    \begin{subfigure}[b]{0.24\textwidth}
        \includegraphics[width=\linewidth]{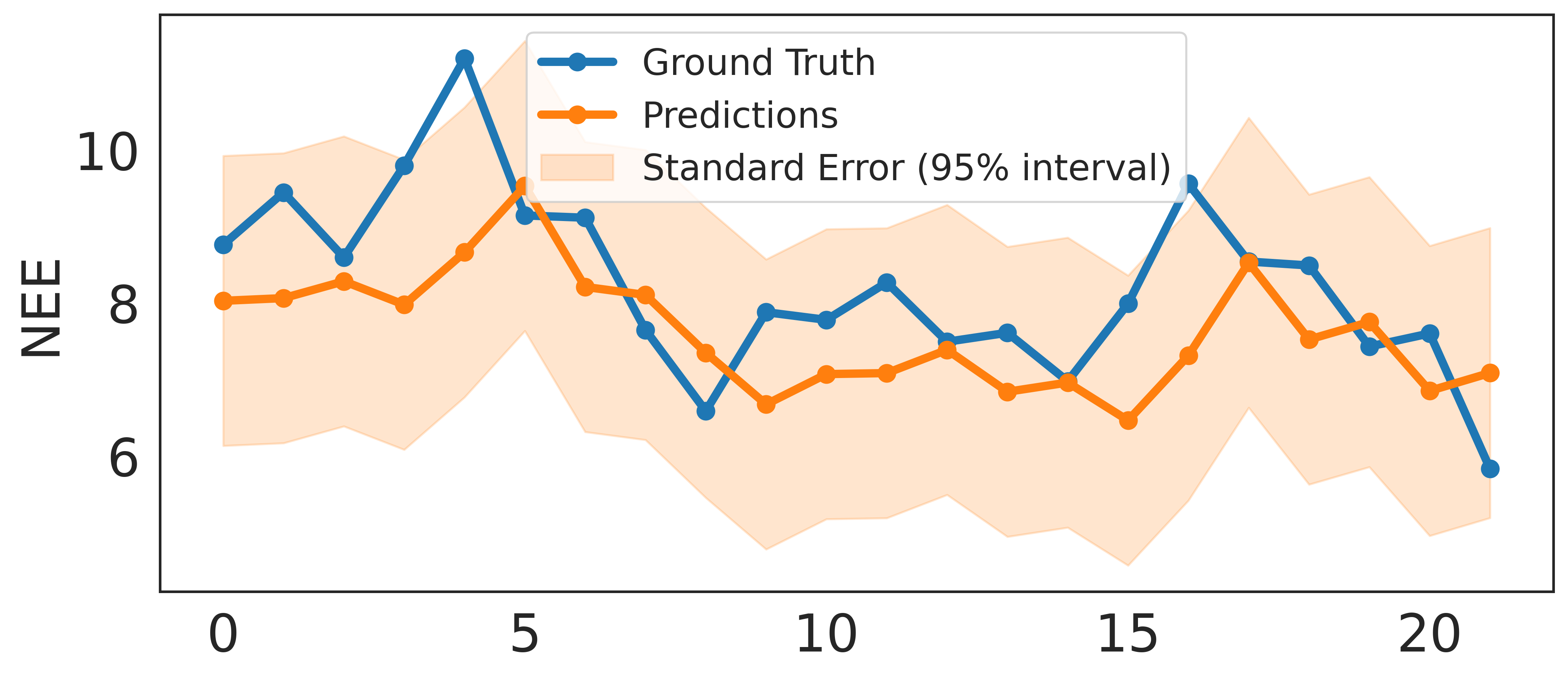}
    \end{subfigure}
    \begin{subfigure}[b]{0.24\textwidth}
        \includegraphics[width=\linewidth]{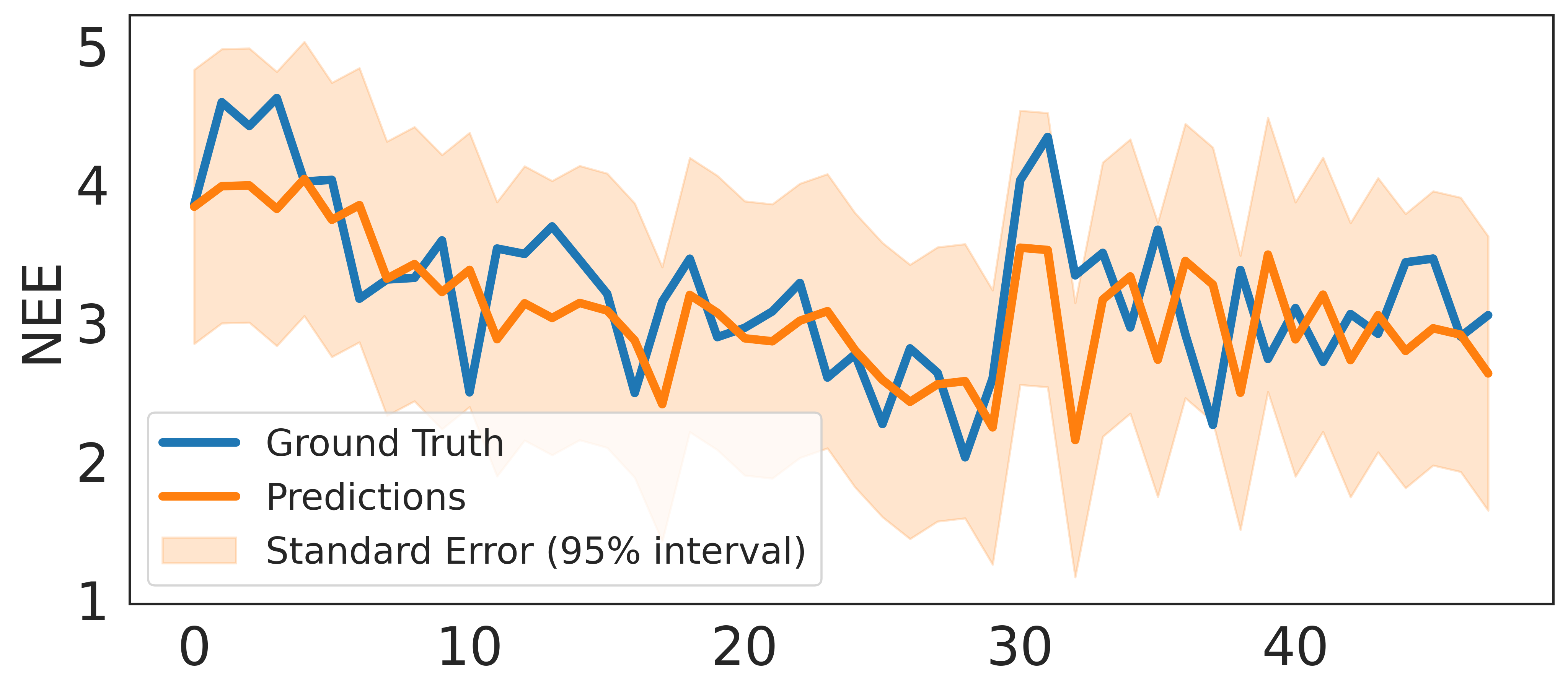}
    \end{subfigure}
    \begin{subfigure}[b]{0.24\textwidth}
        \includegraphics[width=\linewidth]{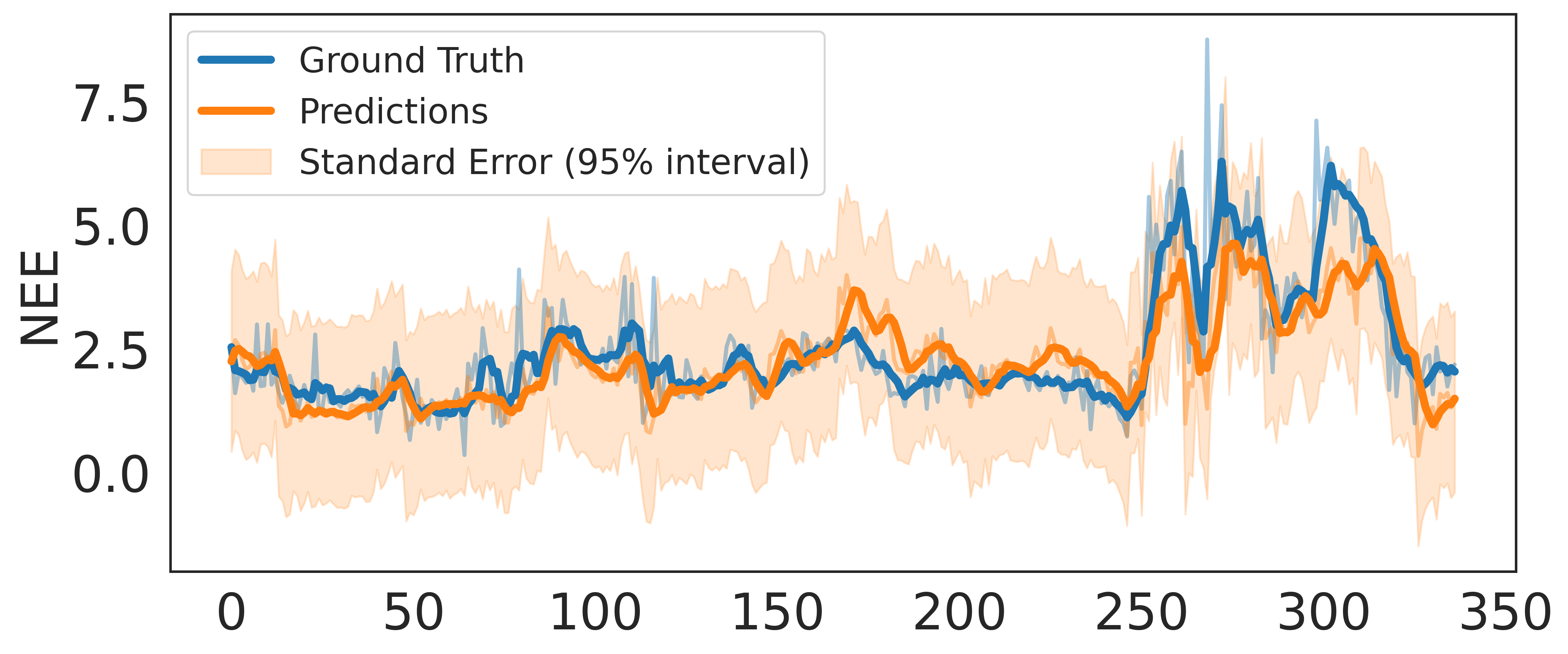}
    \end{subfigure}
    \begin{subfigure}[b]{0.24\textwidth}
        \includegraphics[width=\linewidth]{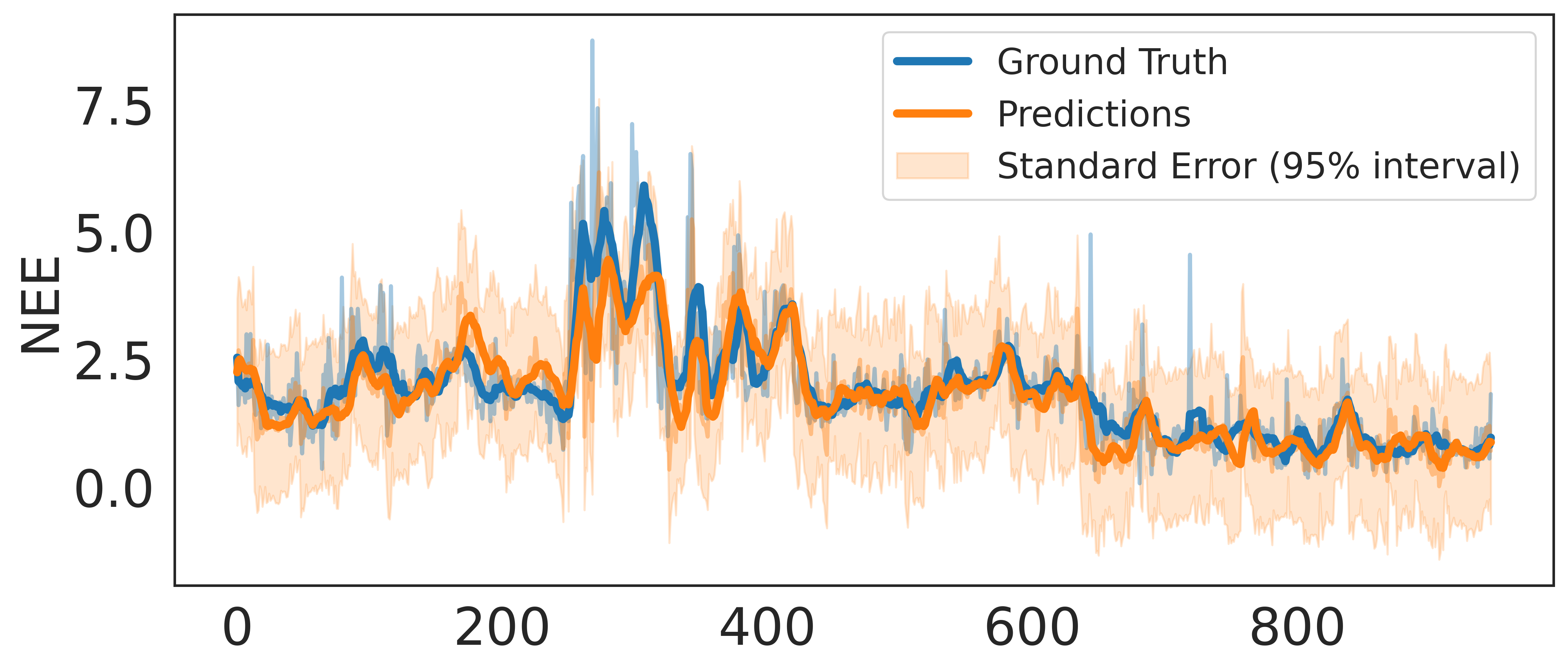}
    \end{subfigure}

    \begin{subfigure}[b]{0.24\textwidth}
        \includegraphics[width=\linewidth]{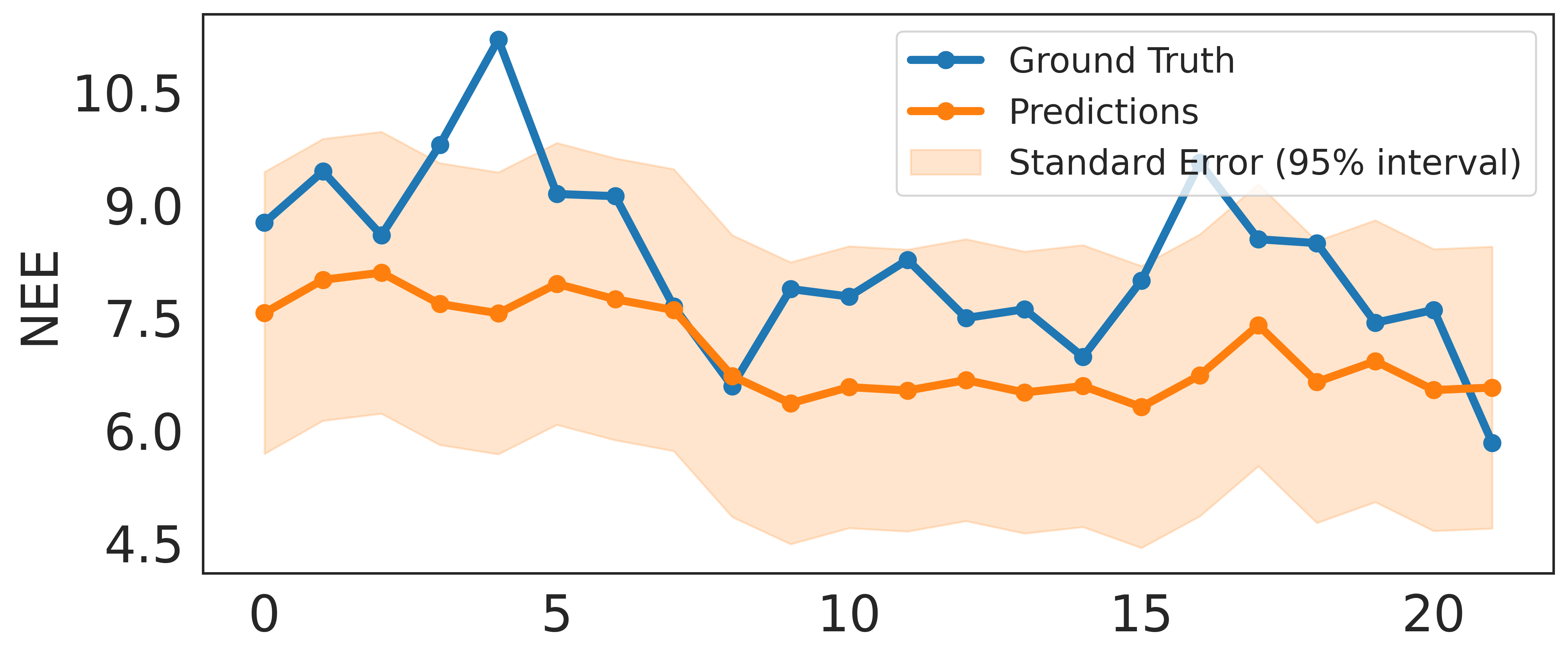}
    \end{subfigure}
    \begin{subfigure}[b]{0.24\textwidth}
        \includegraphics[width=\linewidth]{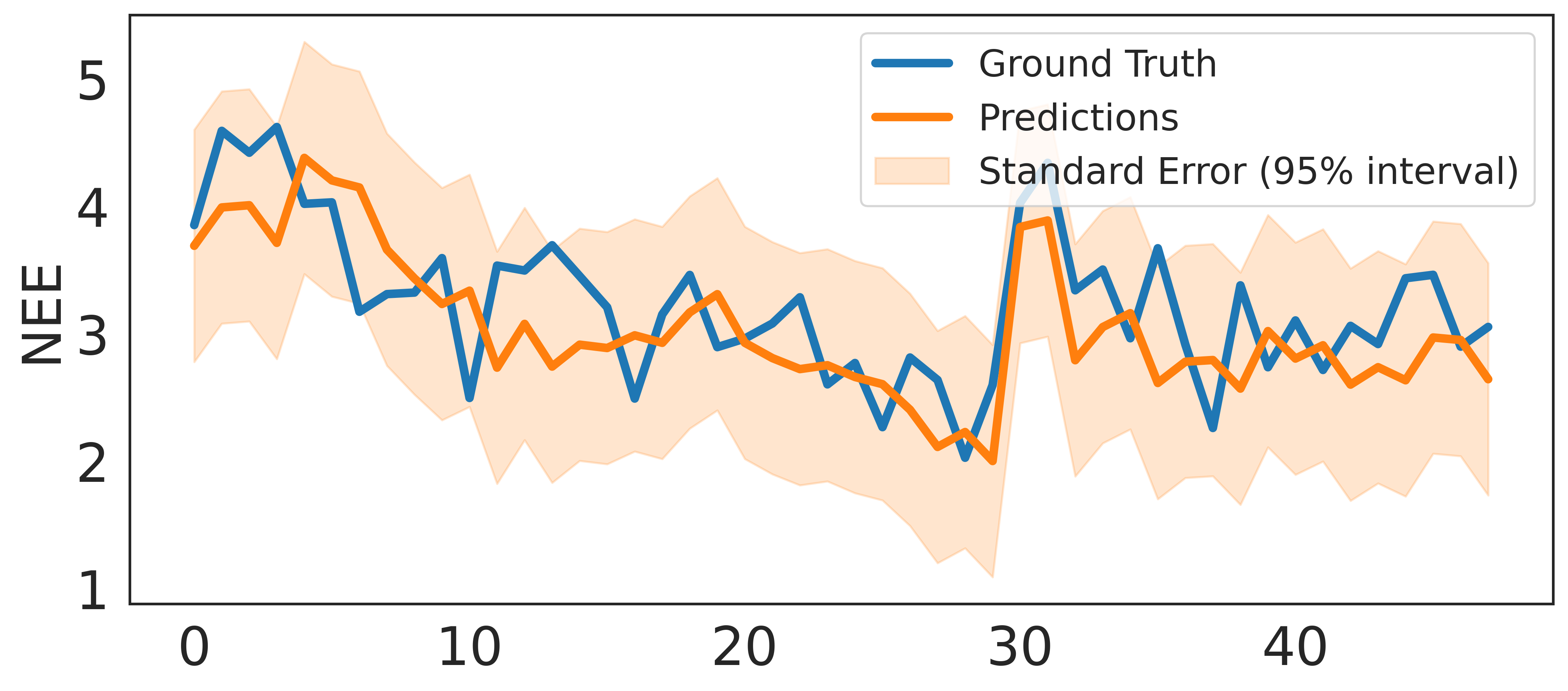}
    \end{subfigure}
    \begin{subfigure}[b]{0.24\textwidth}
        \includegraphics[width=\linewidth]{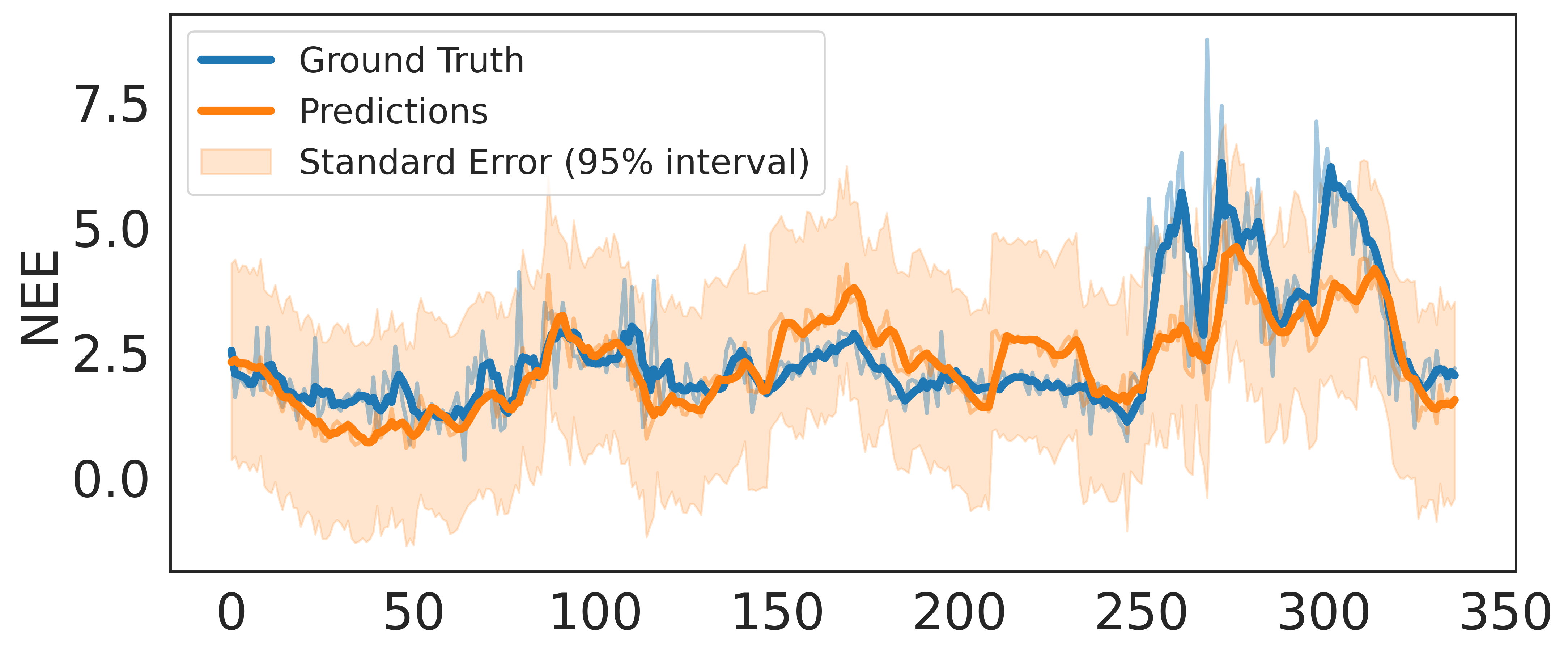}
    \end{subfigure}
    \begin{subfigure}[b]{0.24\textwidth}
        \includegraphics[width=\linewidth]{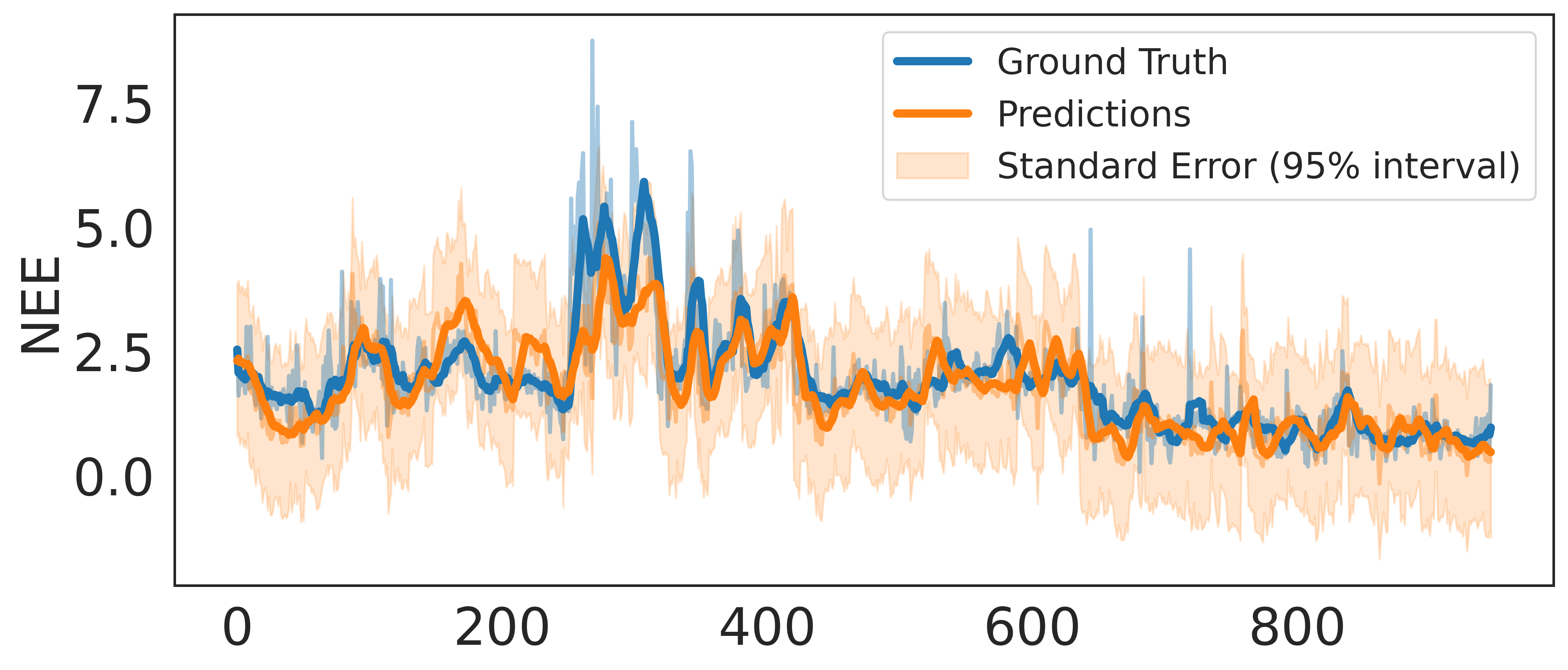}
    \end{subfigure}

    \begin{subfigure}[b]{0.24\textwidth}
        \includegraphics[width=\linewidth]{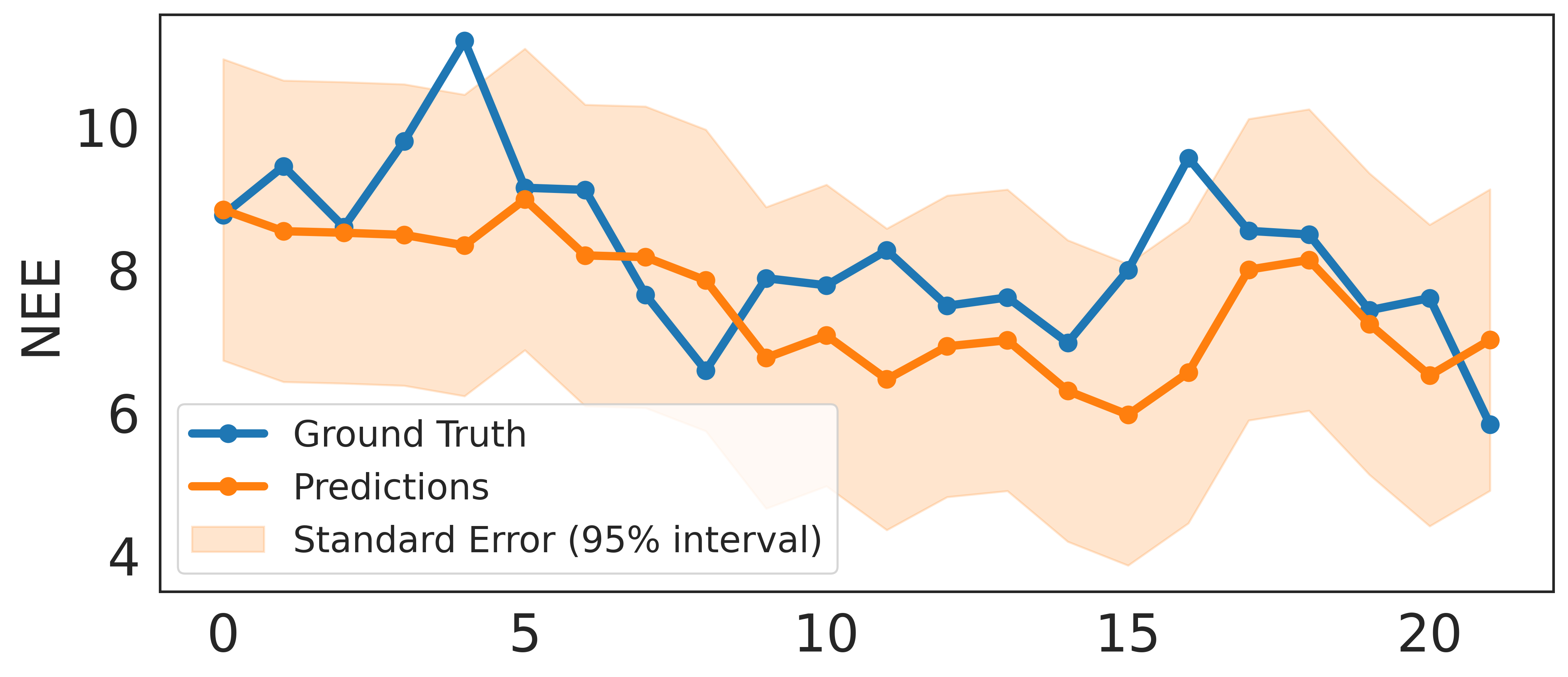}
    \end{subfigure}
    \begin{subfigure}[b]{0.24\textwidth}
        \includegraphics[width=\linewidth]{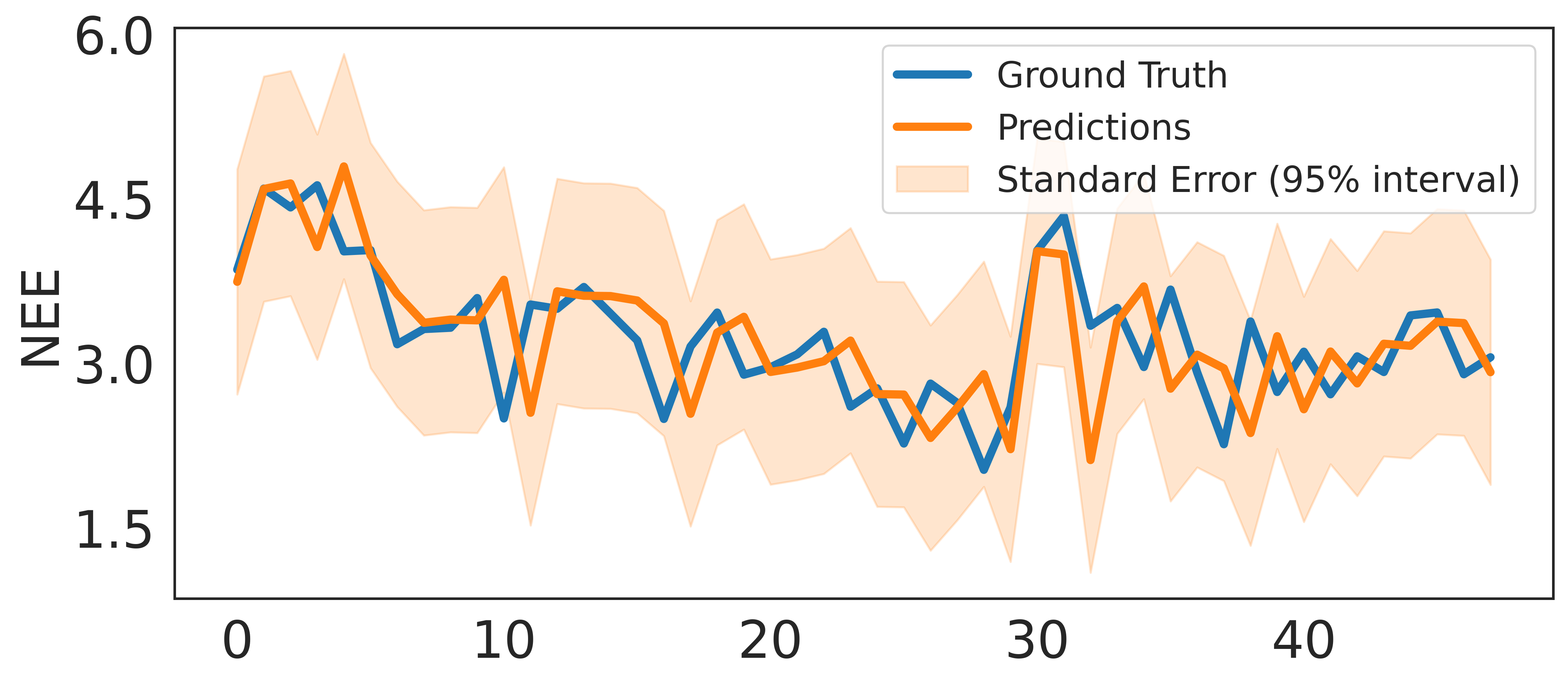}
    \end{subfigure}
    \begin{subfigure}[b]{0.24\textwidth}
        \includegraphics[width=\linewidth]{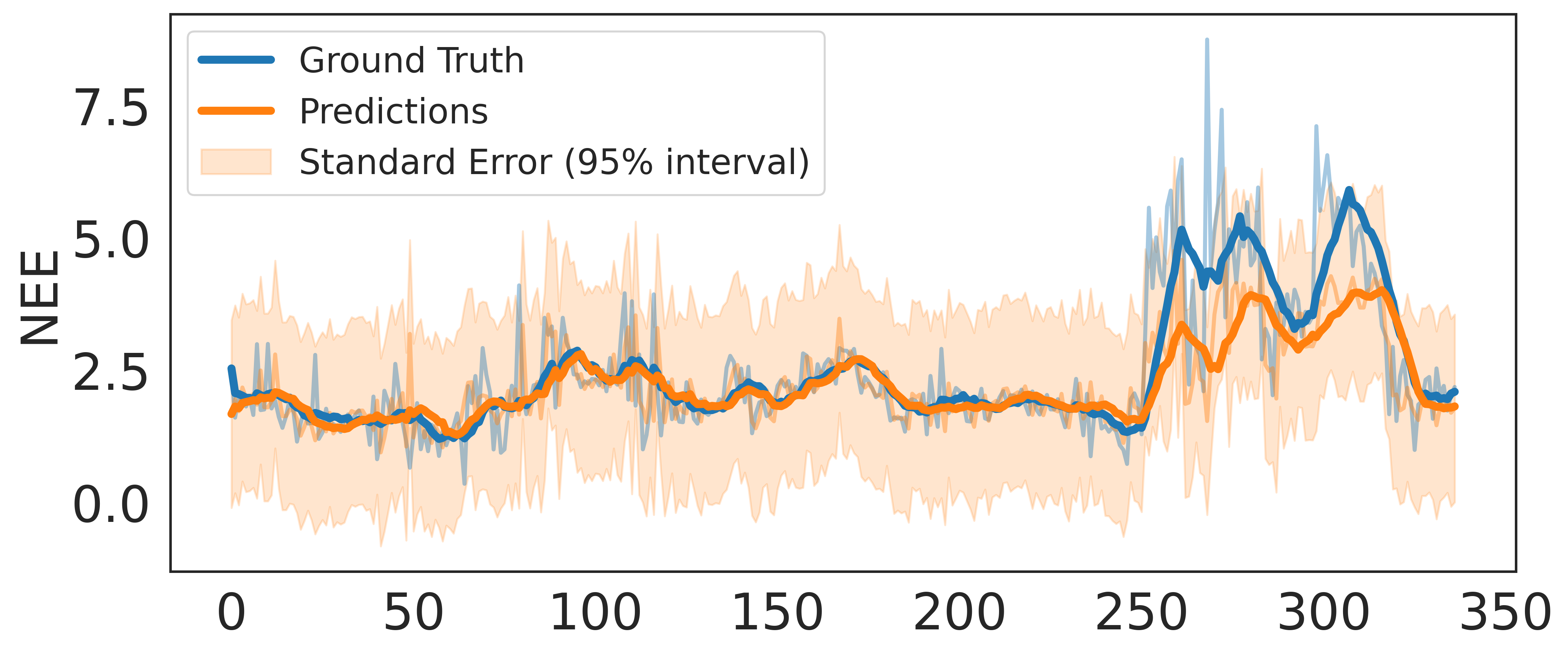}
    \end{subfigure}
    \begin{subfigure}[b]{0.24\textwidth}
        \includegraphics[width=\linewidth]{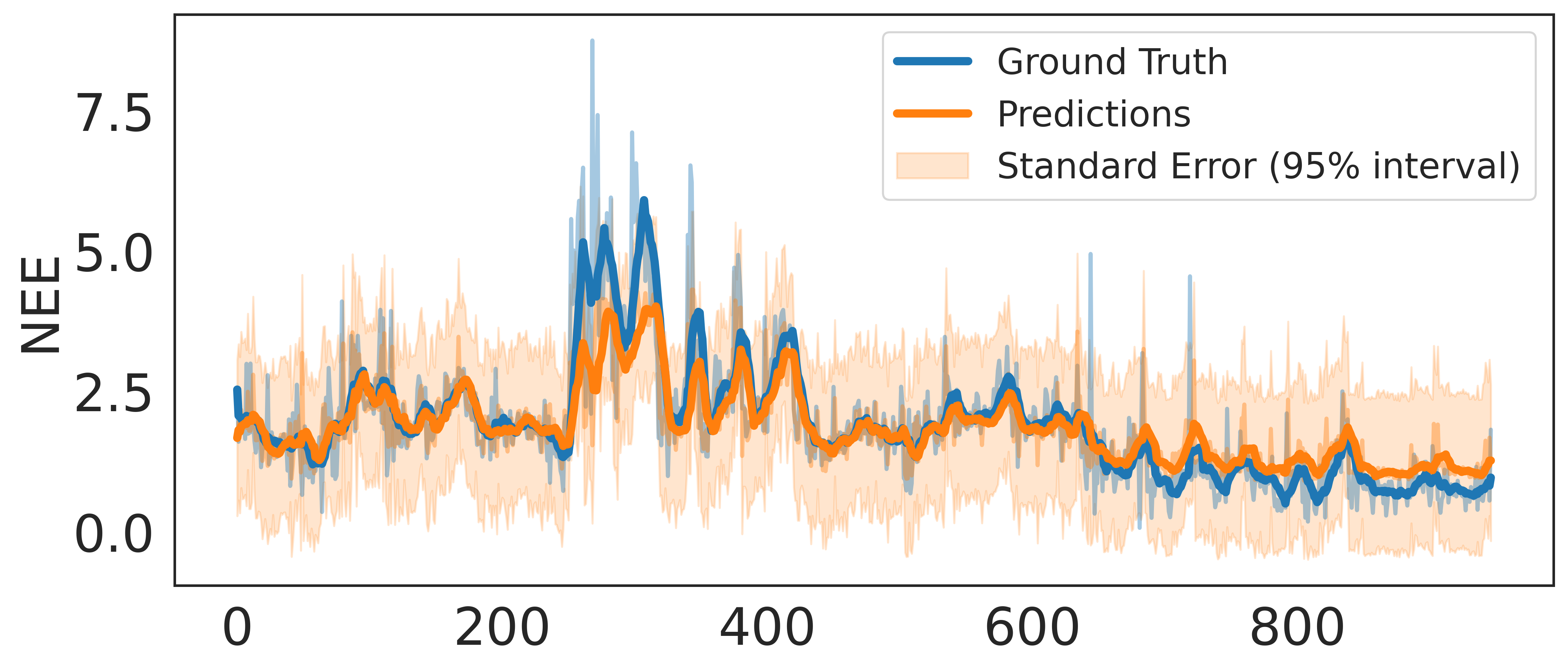}
    \end{subfigure}

    \begin{subfigure}[b]{0.24\textwidth}
        \includegraphics[width=\linewidth]{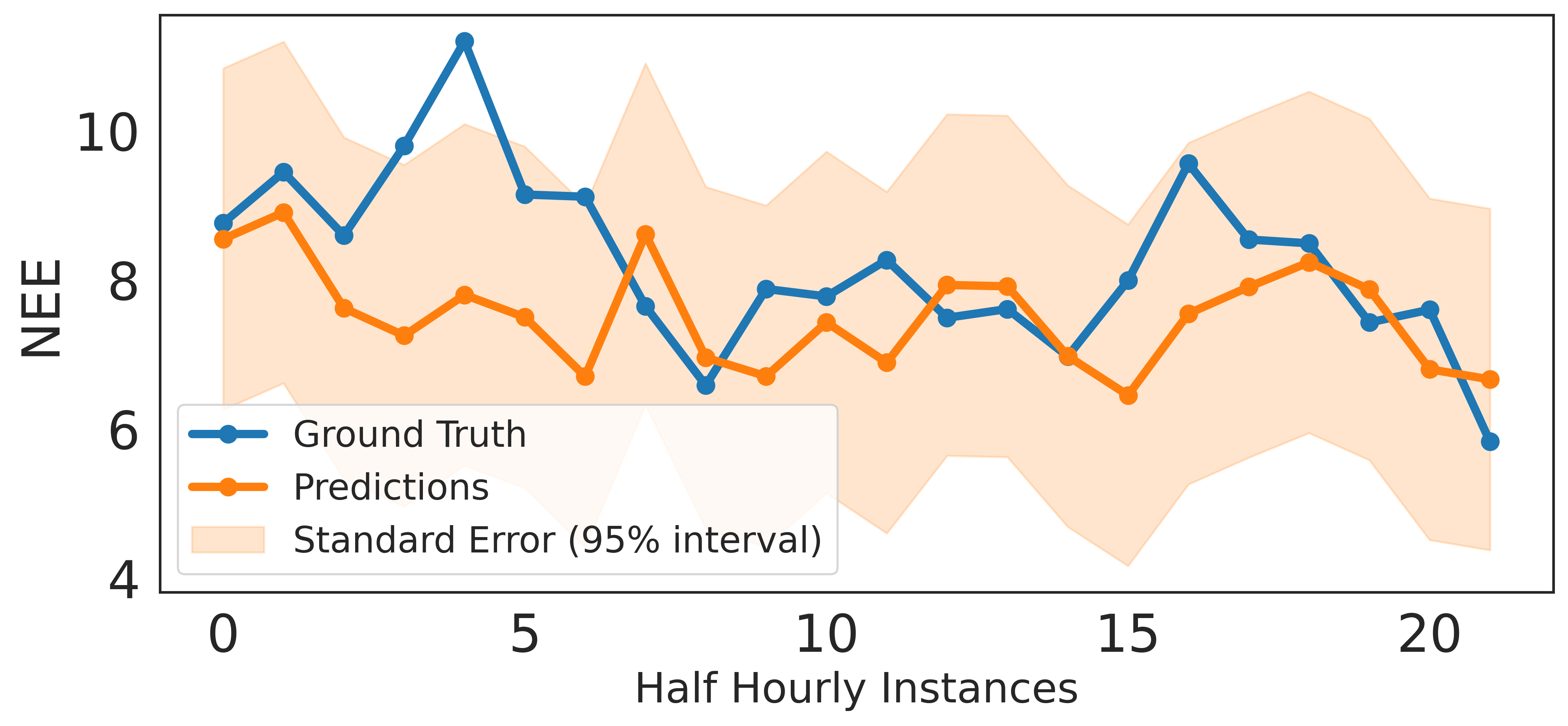}
    \end{subfigure}
    \begin{subfigure}[b]{0.24\textwidth}
        \includegraphics[width=\linewidth]{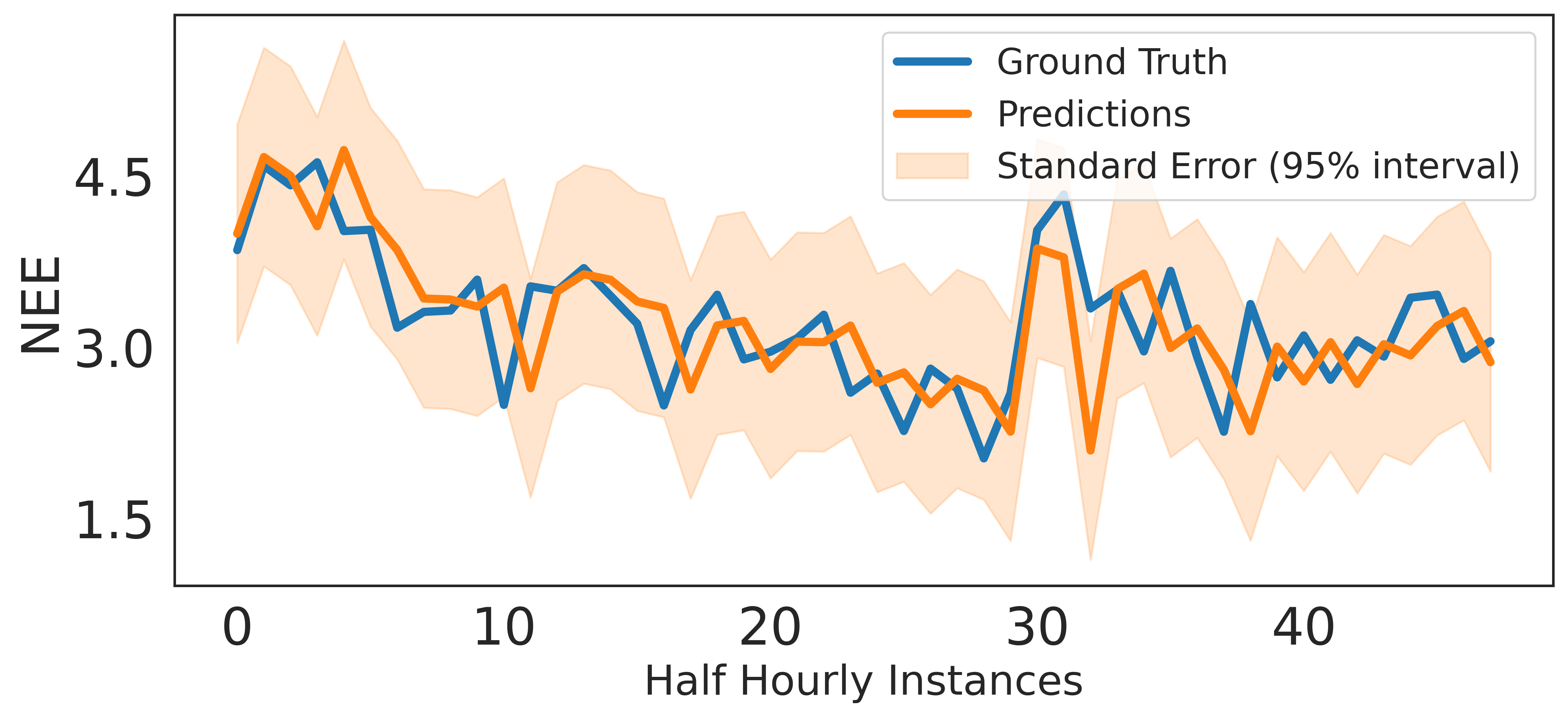}
    \end{subfigure}
    \begin{subfigure}[b]{0.24\textwidth}
        \includegraphics[width=\linewidth]{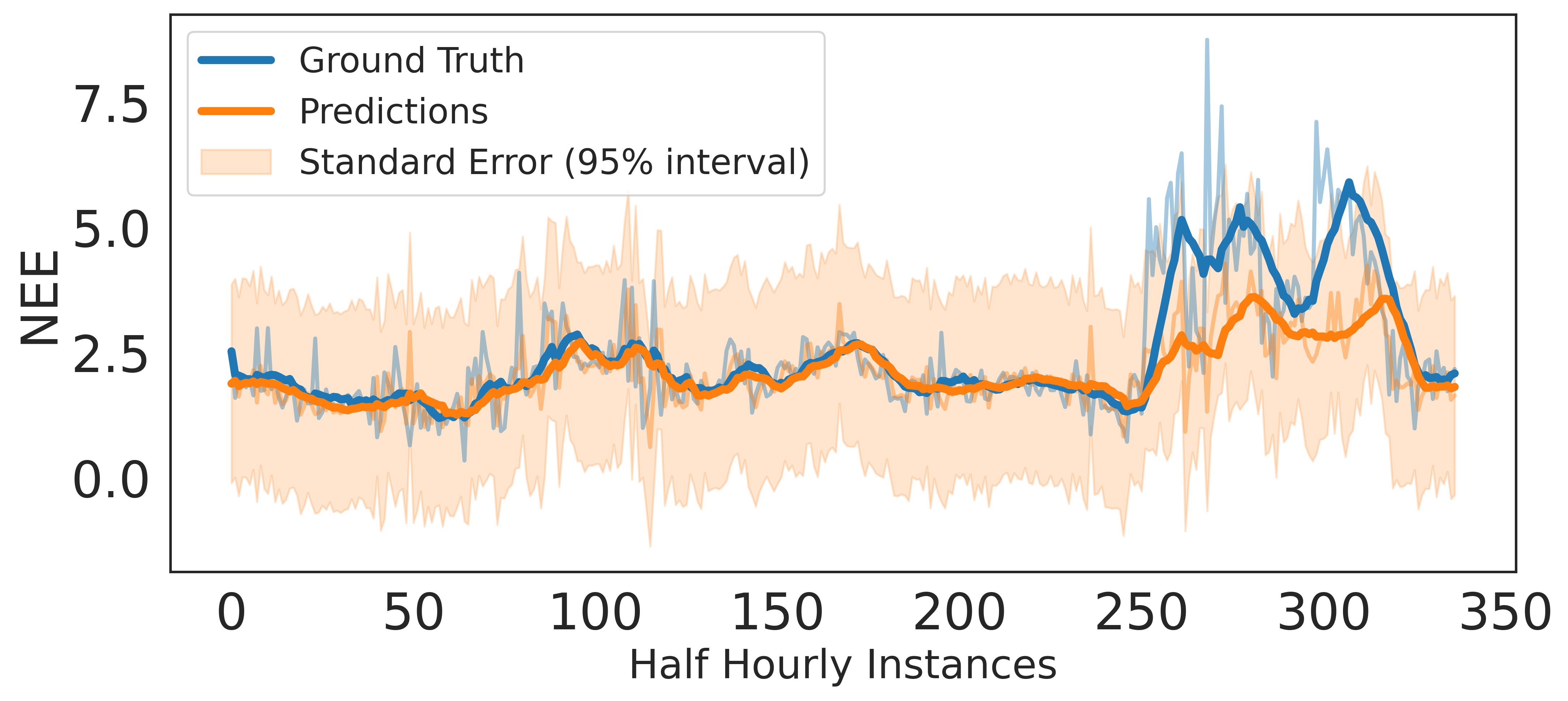}
    \end{subfigure}
    \begin{subfigure}[b]{0.24\textwidth}
        \includegraphics[width=\linewidth]{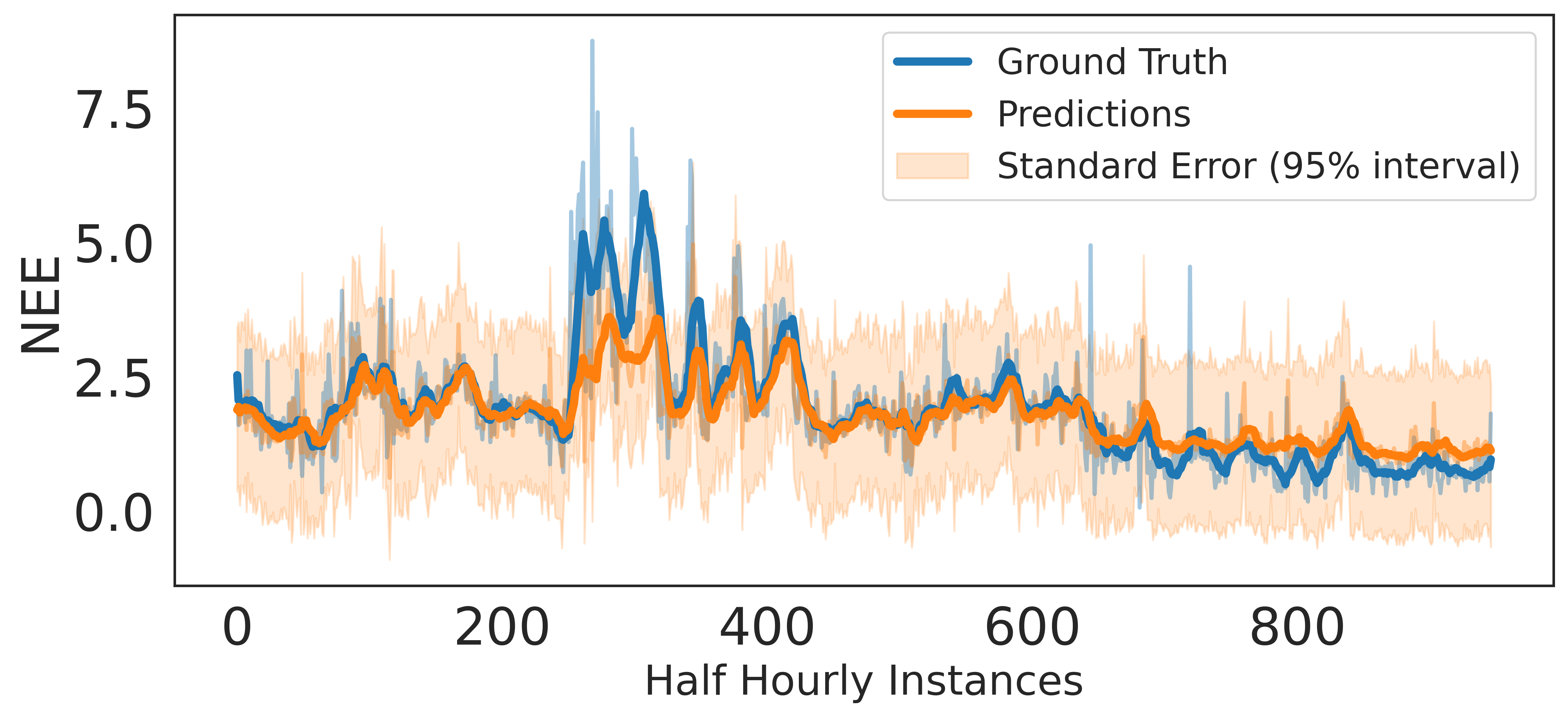}
    \end{subfigure}

    \caption{NEE predictions on test data across different time scales (daily, weekly, monthly, quarterly) for each approach in the experimentation : we create artificial gaps of the corresponding time scales that we fill with the corresponding model. The first column represents results for a single night in test set (2019-09-22), the second column represents results for a single week (2018-11-06 till 2018-11-11), the third columns represents results for a single month (2019-01-01 till 2019-01-29) and fourth column represents results for a single quarter (2019-01-01 till 2019-12-30). Row 1 shows results based on the PERNN model, Row 2 shows results based on the PENN model, Row 3 shows results from the PINN model, Row 4 shows results from the RF model, and Row 5 shows results from the XGB model respectively. The sequences illustrated in the graphs are randomly sampled from the test dataset and are kept consistent for each approach for fair validation.}
    \label{fig:nee_results}
\end{figure*}

Figure \ref{fig:nee_results} shows the visualizations of ground-truth and predicted values for $\text{NEE}_{t+1}$ used for gap-filling across four time scales: daily, weekly, monthly and quarterly. Randomly sampled sequences, consistent across methods for fair validation, show that PERNN consistently captures NEE trends better than PENN, PINN, RF and XgBoost methods. It can also be seen that PENN, albeit falling short of PERNN, appears to capture these trends better than the rest of the methods in the experimentation.

\begin{table*}[h!]
\caption{Results for NEE prediction on night time data and model experiments. The metrics MMD, Wasstn (Wasserstein Distance), KL (Kullback Leibler Divergence), and MAE (Mean Absolute Error) are expressed as the lower the better. R2 (score) is expressed as higher the better.}
\centering
\setlength{\tabcolsep}{3pt} 
\renewcommand{\arraystretch}{1.2} 
\begin{tabularx}{\textwidth}{@{\extracolsep{\fill}} l X X X X X | X X X X}
  \toprule
  \multicolumn{6}{c}{\textbf{$\text{NEE}$}} & \textbf{$E_0$} & \textbf{$r_b$} & \textbf{$dT_{air}$} & \textbf{$d\text{NEE}$} \\
  \midrule
  \textbf{Model} & \textbf{MMD} & \textbf{Wsstn} & \textbf{KL} & \textbf{MAE} & \textbf{R2} & \multicolumn{4}{c}{\textbf{MAE}}  \\
  \midrule
  \textbf{PERNN}      & \textbf{0.051} & \textbf{0.145} & \textbf{0.215} & \textbf{0.866} & \textbf{0.73}  & \textbf{28.8} & \textbf{0.939} & \textbf{0.046} & \textbf{0.141} \\
  PERNN-NoSkip & 0.138 & 0.396 & 0.260 & 1.087 & 0.67  & 29.9 & 1.29 & 0.036 & 0.11 \\
  PENN                & 0.086 & 0.190 & 0.212 & 0.986  & 0.69  & 30.7 & 1.66 & 0.08 & 0.063\\
  PENN-NoSkip         & 0.122 & 0.276 & 1.149 & 1.17  & 0.48  & 31.06 & 1.51 & 0.014 & 0.053 \\
  PINN         & 0.247 & 1.043 & 0.540 & 1.41  & 0.55  & 27.33 & 1.826 & 0.029 & 0.031 \\
  FCNN                & 0.211 & 0.872 & 0.726 & 1.29  & 0.62  & NA & NA & NA & NA \\
  RF                  & 0.055 & 0.237 & 0.242 & 0.901 & 0.721 & NA & NA & NA & NA \\
  XGB                 & 0.052 & 0.202 & 0.214 & 0.988 & 0.658 & NA & NA & NA & NA \\
  \bottomrule
\end{tabularx}
\label{tab:nee_results}
\end{table*}

\FloatBarrier

\section{Conclusion and Future Work}

This paper presents a novel physics-encoded neural network architecture that integrates physics models directly into neural networks through residual knowledge blocks. This approach provides a general framework for incorporating differentiable physics models into data-driven learning algorithms, thereby enhancing reliability and interpretability by enabling the discovery of unobserved environmental variables. Furthermore, the proposed method offers a rapid prototyping capability for digital twin applications in scenarios where both observational data and detailed physical models are limited.

We evaluated our framework in two distinct application domains. The first application involved developing a steering model for autonomous vehicles in the TORCS simulation environment. In this context, our method outperformed pure data-driven fully-connected neural networks and traditional physics-informed neural networks by delivering improved generalizability on unseen road scenarios with significantly reduced model complexity and data requirements. The integration of differentiable geometrical and kinematic operators with learned intermediate variables—facilitated by residual and skip-connection based layers—proved critical for robust performance.

The second application demonstrated the versatility of our framework in a real-world digital twin setting for climate modeling. Specifically, we modeled Net Ecosystem Exchange (NEE) using an ordinary differential equation (ODE) derived from an Arrhenius-type formulation of ecosystem respiration. By embedding this ODE directly into the network via a dedicated physics block and training the model to predict key intermediate variables (e.g., temperature sensitivity \(E_0\), base respiration \(r_{b\_night}\), and the rate of change of ambient air temperature \(\frac{dT}{dt}\)), our method achieved superior gap-filling performance on flux tower data. Experimental results indicate that our approach outperforms state-of-the-art methods, such as Random Forest Robust and XGBoost, in both error metrics and the ability to capture the distributional characteristics of NEE.

In summary, our contributions demonstrate that directly incorporating physical models into neural network architectures not only enhances interpretability and prediction accuracy but also reduces reliance on extensive datasets. Future work will explore extending this framework to a broader array of digital twin applications where data scarcity and incomplete physics are prevalent. Additionally, further studies on gradient flow and the development of more robust residual blocks will be pursued to tackle increasingly complex physics-based models.

\FloatBarrier

\backmatter

\section*{Declarations}

\begin{itemize}
\item \textbf{Funding: } The research presented in this manuscript is supported by BT (British Telecom) PLC, and two Engineering and Physical Sciences Research Council (EPSRC) grants: AI-driven Digital Twins for Net Zero (EP/Y00597X/1) and Clinical Care (EP/Y018281/1)
\item \textbf{Conflict of interest/Competing interests:} The IP for this work is owned by BT PLC, and as part of the agreement, the patent for this work has been filed before submitting the document to the journal.
\item \textbf{Ethics approval:} The experiments were performed on data not related to living beings. No human beings and animals were involved in conducting the experiments. Thus, approval from an ethical committee is not required.
\item \textbf{Consent to participate:} No living beings are involved in conducting the experiments for this research. Thus, consent to participate is not required.
\item \textbf{Consent for publication:} The authors involved in conducting this research give their consent for the publication of the article titled ``Physics Encoded Blocks in Residual Neural Network Architectures for Digital Twin Models".
\item \textbf{Availability of data and materials:} The simulation and agents used to generate data and test agents in this research are available online. The references and citations are provided in this manuscript.
\item \textbf{Code availability:} The code has been made available on GitHub. \url{https://github.com/saadzia10/Physics-Encoded-ResNet.git}
\item \textbf{Authors' contributions:} Muhammad Saad Zia is the primary author who contributed to the main ideation and experimentation in this manuscript. Corentin Houpert provided mathematical formulation for the Net Ecosystem Exchange ODE. Ashiq Anjum is the primary supervisor and reviewer on the work in this manuscript. Lu Liu is an academic reviewer and collaborator on the work in this manuscript. Anthony Conway and Anasol Peña-Rios are industrial collaborators on the work in this manuscript. 
\end{itemize}

\bibliography{references}

\end{document}